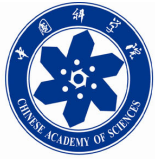

# 中国科学院大学
**University of Chinese Academy of Sciences**

# 博士学位论文

### 基于神经网络的词和文档语义向量表示方法研究

| | |
|---|---|
| 作者姓名 | 来斯惟 |
| 指导教师 | 赵军 研究员 |
| | 中国科学院自动化研究所 |
| 学位类别 | 工学博士 |
| 学科专业 | 模式识别与智能系统 |
| 培养单位 | 中国科学院自动化研究所 |

**2016 年 1 月**

# Word and Document Embeddings based on Neural Network Approaches

By

## Siwei Lai

A Dissertation Submitted to

The University of Chinese Academy of Sciences

In partial fulfillment of the requirement

For the degree of

Doctor of Engineering

Institute of Automation

Chinese Academy of Sciences

January, 2016

# 独创性声明

　　本人声明所递交的论文是我个人在导师指导下进行的研究工作及取得的研究成果。尽我所知，除了文中特别加以标注和致谢的地方外，论文中不包含其他人已经发表或撰写过的研究成果。与我一同工作的同志对本研究所做的任何贡献均已在论文中作了明确地说明并表示了谢意。

　　　　　　　　　　　　　签 名:________________ 日 期: ________________

# 关于论文使用授权的说明

　　本人完全了解中国科学院自动化研究所有关保留、使用学位论文的规定，即：中国科学院自动化研究所有权保留送交论文的复印件，允许论文被查阅和借阅；可以公布论文的全部或部分内容，可以采用影印、缩印或其他复制手段保存论文。
**(保密的论文在解密后应遵守此规定)**

　　　　　　　　　　　　签 名:________________ 导师签名:________________ 日 期: ________________

# 摘　要

　　数据表示是机器学习中的基础工作，数据表示的好坏直接影响到整个系统的性能。传统机器学习思路下，对数据的表示主要通过人工设计特征来完成，在很长一段时间里，文本、语音、图像领域中的各项任务均通过人工设计更好的特征来实现性能的提升。近年来，随着深度学习和表示学习的兴起，基于神经网络的数据表示技术在各个领域崭露头角。

　　在自然语言处理领域，最常用的语义表示方法是词袋子模型，该方法存在数据稀疏问题，并且不能保留词序信息。早期方法中提出的词性、句法结构等复杂特征，往往只能对特定的任务带来性能提升。本文从词和文档两个层次对文本的语义表示技术进行系统的总结分析，并提出了自己的表示技术，具体如下。

　　**一、词向量表示技术的理论及实验分析**。在这一部分中，本文对现有的词向量表示技术进行了系统的理论对比及实验分析。理论方面，本文阐述了现有各种模型之间的联系，从模型的结构与目标等方面对模型进行了比较，并证明了其中最重要的两个模型 Skip-gram 与 GloVe 之间的关系。实验方面，本文从模型、语料和训练参数三个角度分析了训练词向量的关键技术。本文选取了三大类一共八个指标对词向量进行评价，这三大类指标涵盖了现有的词向量用法。本工作为首个对词向量进行系统评价的工作，通过理论和实验的比较分析，文章提出了一些对生成词向量的参考建议。

　　**二、基于字词联合训练的中文表示及应用**。现有的中文表示技术往往沿用了英文的思路，直接从词的层面对文本表示进行构建。本文根据中文的特点，提出了基于字词联合训练的表示技术。该方法在字的上下文空间中融入了词，利用词的语义空间，更好地对汉字建模；同时利用字的平滑效果，更好地对词建模。文章在分词任务、词义相似度任务和文本分类任务上对字和词的表示进行了评价，实验表明字词联合训练得到的字词向量，相比单独训练字向量或词向量，有显著的提升。

　　**三、基于循环卷积网络的文档表示及应用**。在这一部分中，本文分析了现有的文档表示技术：基于循环网络的表示技术、基于递归网络的表示技术和基于卷积网络的表示技术。并且，针对现有的三种表示技术的不足，本文提出了

基于卷积循环网络的文档表示技术。该方法克服了此前递归网络的复杂度过高的问题，循环网络的语义偏置问题，以及卷积网络窗口较难选择的问题。文章在文本分类任务上对新提出的表示技术进行了对比分析，实验表明基于循环卷积网络的文本表示技术比现有的表示技术能取得更好的性能。

# Abstract

Data representation is a fundamental task in machine learning. The representation of data affects the performance of the whole machine learning system. In a long history, the representation of data is done by feature engineering, and researchers aim at designing better features for specific tasks. Recently, the rapid development of deep learning and representation learning has brought new inspiration to various domains.

In natural language processing, the most widely used feature representation is the Bag-of-Words model. This model has the data sparsity problem and cannot keep the word order information. Other features such as part-of-speech tagging or more complex syntax features can only fit for specific tasks in most cases. This thesis focuses on word representation and document representation. We compare the existing systems and present our new model.

First, for generating word embeddings, we make comprehensive comparisons among existing word embedding models. In terms of theory, we figure out the relationship between the two most important models, i.e., Skip-gram and GloVe. In our experiments, we analyze three key points in generating word embeddings, including the model construction, the training corpus and parameter design. We evaluate word embeddings with three types of tasks, and we argue that they cover the existing use of word embeddings. Through theory and practical experiments, we present some guidelines for how to generate a good word embedding.

Second, in Chinese character or word representation, we find that the existing models always use the word embedding models directly. We introduce the joint training of Chinese character and word. This method incorporates the context words into the representation space of a Chinese character, which leads to a better representation of Chinese characters and words. In the tasks of Chinese character segmentation and document classification, the joint training outperforms the existing methods that train characters or words with traditional word embedding algorithms.

Third, for document representation, we analyze the existing document representation models, including recursive neural networks, recurrent neural networks and con-volutional neural networks. We point out the drawbacks of these models and present our new model, the recurrent convolutional neural networks. In text classification task, the experimental results show that our model outperforms the existing models.

# 目　录

# 表　格

# 插　图

# 术语与符号

## 0.1 术语

- 分布假说（distributional hypothesis）：上下文相似的词，其语义也相似。该假说由 Harris 在 1954 年提出 [35]，并由 Firth 在 1957 年进一步明确和完善 [29]。
- 分布表示（distributional representation）：分布（distributional）描述的是上下文的概率分布，因此用上下文描述语义的表示方法（基于分布假说的方法）都可以称作分布表示。与之相对的是形式语义表示。
- 分布式表示（distributed representation）：分布式（distributed）描述的是把信息分布式地存储在向量的各个维度中，与之相对的是局部表示（local representation），如词的独热表示（one-hot representation），在高维向量中只有一个维度描述了词的语义。一般来说，通过矩阵降维或神经网络降维可以将语义分散存储到向量的各个维度中，因此，这类方法得到的低维向量一般都可以称作分布式表示。

## 0.2　符号

为了更一致地描述词和文档表示中各项技术，全文的符号系统统一如下：

- 粗体小写字母表示列向量，如 $\boldsymbol{h}$、$\boldsymbol{p}$、$\boldsymbol{q}$、$\boldsymbol{e}$ 等。其中 $\boldsymbol{e}(w)$ 特指词 $w$ 的词向量，$\boldsymbol{e}'(w)$ 特指词 $w$ 的辅助词向量（具体作用在模型中会有介绍）。

- 大写字母表示矩阵，常用的符号有 $H$、$W$、$U$、$A$、$B$ 等。需要注意的是，为了与常用的数学及神经网络符号统一，$O$ 仍然为复杂度渐近上限记号，$E$ 表示能量函数。

- 双线体大写字母表示集合，具体包括：$\mathbb{D}$ 表示数据集（包括训练词向量的语料、训练文本分类的数据集以及训练分词模型的数据集）；$\mathbb{R}$ 表示实数集，$\mathbb{R}^a$ 表示 $a$ 维实数向量集合，$\mathbb{R}^{a \times b}$ 表示 $a$ 行 $b$ 列的实数矩阵集合；$\mathbb{V}$ 表示词表（单词的集合）。

- 正体字表示数学函数，如 exp、max、tanh 等。

- $\phi$ 表示非线性激活函数，可能为 tanh、sigmoid（Logistic 函数）、ReLU [32] 等。

- 绝对值符号 $|x|$ 对于集合表示集合的大小，如 $|\mathbb{V}|$ 表示词表中词的总个数；对于向量表示向量的维度，如 $|\boldsymbol{e}|$ 表示词向量的维度。

- $[\boldsymbol{p}_1; \boldsymbol{p}_2; \ldots; \boldsymbol{p}_n]$ 表示向量 $\boldsymbol{p}_1, \boldsymbol{p}_2, \ldots, \boldsymbol{p}_n$ 的拼接。

- 词、词序列的描述方法：

  - $v_i$ 和 $w_i$ 均特指词，其中 $v_i$ 表示词表中的第 $i$ 个词（满足 $1 \le i \le |\mathbb{V}|$）；$w_i$ 表示文本序列（句子或文档）中的第 $i$ 个词。

  - 词序列有两种描述方法，列举式描述 $w_2, w_3, \ldots, w_8$ 与起止点描述 $w_{2:8}$ 等价。

  - $c$ 特指词 $w$ 的上文或上下文。在语言模型中，词 $w_i$ 对应的上文 $c$ 为 $w_{1:i-1}$，或 $n$ 元语言模型中的 $w_{i-(n-1):i-1}$。在其它词向量模型中，词 $w_i$ 对应的上下文 $c$ 为 $w_{i-(n-1)/2}, \ldots, w_{i-1}, w_{i+1}, \ldots, w_{i+(n-1)/2}$。

## 0.3 图例

为了更清晰且一致地描述神经网络的结构，本文使用以下图案体系:

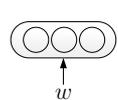

○ 表示神经网络节点（一个实数值）。

表示向量。其中节点个数仅作示意用，不是实际的向量维度。

→ 表示线性变换。箭头尾部向量与矩阵相乘，得到箭头头部向量。

表示词 $w$ 的词向量。$w$ 是词的独热表示，与词向量矩阵相乘之后，得到矩阵中的一行，也就是 $w$ 的词向量。词向量也可以简单地表示成 ($w$ ○○○)。两种方法含义一致，仅根据排版需要而选择。

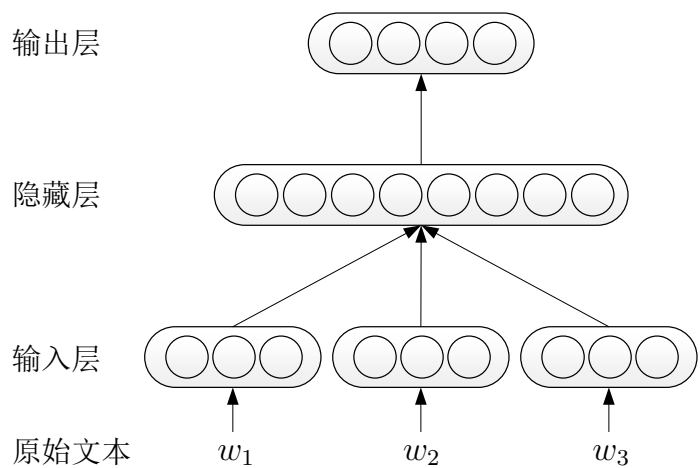

输出层

隐藏层

输入层

原始文本 $w_1$ $w_2$ $w_3$

图 0-1 神经网络模型结构示意图

图 0-1 展示了一个三层前馈网络的图。其中第一层由 3 个词向量拼接而成，经过线性变换，得到隐藏层。输入层到隐藏层可以看作三个向量与三个矩阵相乘，求和得到隐藏层；也可以理解为三个向量拼接成一个更长的向量，与一个大矩阵相乘，得到隐藏层。一般而言，神经网络的隐藏层都需要通过一个非线性的激活函数 $\phi$，因此在图中省去这部分的表示。最后隐藏层通过线性变换得到输出层，输出层一共有 4 个节点。

# 第一章 绪论

## 1.1 研究背景

数据表示是机器学习中的基础工作，数据表示的好坏直接影响到整个机器学习系统的性能 [5]。因此，人们投入了大量精力去研究如何针对具体任务，设计一种合适的数据表示方法，以提升机器学习系统的性能，这一环节也被称作特征工程。传统机器学习方法不能直接从数据中自动挖掘出有判别力的信息，而特征工程正是通过人类的智慧、知识和灵感来弥补机器学习方法的这一缺陷。在自然语言处理领域，最常用的文本表示方法是词袋子表示 [128]，该方法会面临数据稀疏问题，并且不能保留词序信息。研究人员针对这些缺陷，还提出了词法特征、句法特征等复杂特征。借助这些人工精心设计的特征，机器学习在自然语言领域逐步取代了以往基于规则的方法，成为自然语言处理中的主流方法。

特征工程在传统机器学习算法中，有着不可替代的地位，但是由于需要大量人力和专业知识，反而成为了机器学习系统性能提升的瓶颈。为了让机器学习算法有更好的扩展性，研究人员希望可以减少对特征工程的依赖。这样，当把机器学习算法推广到新的领域中时，就可以省去大量专家在新领域上的分析和探索，加快应用的进程，使得系统更为智能。从人工智能的角度看，算法直接从原始的感知数据中自动分辨出有效的信息，是机器走向智能的重要一步。

近年来，随着 Web2.0 的兴起，互联网上的数据急剧膨胀。根据国际数据公司（IDC）的统计和预测，2011 年全球网络数据量已经达到 1.8ZB，到 2020 年，全球数据总量预计将增长 50 倍。大量无标注数据的出现，也让研究人员开始考虑，如何利用算法从这些大规模无标注的数据中自动挖掘规律，得到有用的信息。2006 年 Hinton 提出的深度学习 [40]，为解决这一问题带来了新的思路。在之后的发展中，基于神经网络的表示学习技术开始在各个领域崭露头角。尤其在图像和语音领域的多个任务上，基于表示学习的方法在性能上均超过了传统方法。

但是，在自然语言处理领域，深度学习技术并没有产生类似图像和语音领域那样的突破。其中一个主要的原因是，在图像和语音领域，最基本的数据是

信号数据，我们可以通过一些距离度量，判断信号是否相似。而文本是符号数据，两个词只要字面不同，就难以刻画它们之间的联系，即使是"麦克风"和"话筒"这样的同义词，从字面上也难以看出这两者意思相同（语义鸿沟现象）。正因为这样，在判断两幅图片是否相似时，只需通过观察图片本身就能给出回答；而判断两个词是否相似时，还需要更多的背景知识才能做出回答。

我们希望计算机可以从大规模无标注的文本数据中自动学习得到文本表示，这种表示需要包含对应语言单元（词或文档）的语义信息，同时可以直接通过这种表示度量文本之间的语义相似度。

1954 年，Harris 提出分布假说（distributional hypothesis），即"上下文相似的词，其语义也相似"[35]，为词的分布表示提供了理论基础。基于分布假说，研究人员提出了多种词表示模型：如基于矩阵的 LSA 模型 [56]、基于聚类的 Brown clustering 模型 [12] 以及本文关注的神经网络词表示模型，本文第二章对这些模型进行了综述。在分布假说中，需要关注的对象有两个：词和上下文，其中最关键的是上下文的表示。在前两个模型中，上下文只能使用传统的词袋子表示，如果需要表示复杂的上下文，会遇到维数灾难问题。而神经网络模型可以使用组合方式对上下文进行建模，只需线性复杂度即可对复杂的 $n$ 元短语进行建模。神经网络模型生成的词表示通常被称为词向量（word embedding），是一个低维的实数向量表示，通过这种表示，可以直接对词之间的相似度进行刻画。相比传统的词袋子表示方法以及矩阵、聚类等衍生方法，词向量可以缓解维数灾难的问题。从广义上讲，传统的词袋子模型也是用向量描述文本，也应当被称作词的向量表示，但是这种向量是高维稀疏的。在本文中，"词向量"特指由神经网络模型得到的低维实数向量表示。

对于文本分类、信息检索等实际需求而言，仅使用词级别的语义表示不足以有效地完成这些任务，因此还需要通过模型，得到句子和文档级别的语义表示。但是，由于文档的多样性，直接使用分布假说构建文档的语义向量表示时，会遇到严重的数据稀疏问题；同时由于分布假说是针对词义的假说，这种通过上下文获取语义的方式对句子和文档是否有效，还有待讨论。为了获取句子和文档的语义表示，研究人员一般采用语义组合的方式。德国数学家弗雷格（Gottlob Frege）在 1892 年提出：一段话的语义由其各组成部分的语义以及它们之间的组合方法所确定 [30]。现有的句子或者文档表示也通常以该思路为基础，通过语义组合的方式获得。主流的神经网络语义组合方法包括递归神经网络、循环

神经网络和卷积神经网络，这些方法采用了不同的组合方式从词级别的语义组合到句子和文档级别。

## 1.2　论文结构

本文着眼于基于神经网络的词和文档表示技术，通过理论分析和实验比较，探索现有方法的联系和区别，比较其优劣，并提出自己的文本表示技术。

本文一共分为六章，后续章节安排如下：

- 第二章介绍了以分布假说为基础的分布表示的体系，系统地总结对比了基于矩阵、基于聚类和基于神经网络的分布表示方法，并详细介绍了基于神经网络的词向量表示技术。

- 第三章对现有的词向量表示技术进行了系统实验分析。具体从模型、语料和训练参数三个角度分析训练词向量的关键技术。本文选取了三大类一共八个指标对词向量进行评价，这三大类指标涵盖了现有的词向量用法。在此基础上，通过理论和实验的比较分析，提出了一系列针对生成词向量的参考建议。

- 第四章提出了基于字词联合训练的中文字词表示方法。现有的中文表示技术往往沿用了英文的思路，直接从词的层面对文本表示进行构建。本文根据中文的特点，提出了基于字词联合训练的中文表示技术。该方法在字的上下文空间中融入了词，利用词的语义空间，更好地对汉字建模。同时也利用字，加深了对词义的建模。实验表明，字词联合训练对字表示和词表示均有一定的提升。

- 第五章综述了现有的神经网络文档向量表示技术，并针对现有的三种表示技术的不足，提出了基于循环卷积网络的文档表示技术。该方法克服了此前递归网络复杂度过高的问题，循环网络的语义偏置问题，以及卷积网络窗口较难选择的问题。文章在文本分类任务上对新提出的表示技术进行了对比分析，实验表明，基于循环卷积网络的文本表示技术比现有的表示技术有更好的性能。

- 第六章对前面的工作进行了总结，并提出了进一步研究的展望。

# 第二章 现有词的分布表示技术

本章首先介绍以分布假说为基础的分布表示方法体系，然后详细介绍其中基于神经网络的分布表示方法（词向量），最后总结以上所有方法，并阐述它们之间的联系和区别。

## 2.1 分布表示

词是承载语义的最基本的单元 [1]，而传统的独热表示（one-hot representation）仅仅将词符号化，不包含任何语义信息。如何将语义融入到词表示中？Harris 在 1954 年提出的分布假说（distributional hypothesis）为这一设想提供了理论基础：上下文相似的词，其语义也相似 [35]。Firth 在 1957 年对分布假说进行了进一步阐述和明确：词的语义由其上下文决定（a word is characterized by the company it keeps）[29]。二十世纪 90 年代初期，统计方法在自然语言处理中逐渐成为主流，分布假说也再次被人关注。Dagan 和 Schütze 等人总结完善了利用上下文分布表示词义的方法，并将这种表示用于词义消歧等任务 [20, 21, 100, 101]，这类方法在当时被成为词空间模型（word space model）。在此后的发展中，这类方法逐渐演化成为基于矩阵的分布表示方法，期间的十多年时间里，这类方法得到的词表示都被直接称为分布表示（distributional representation）。1992 年，Brown 等人同样基于分布假说，构造了一个上下文聚类模型，开创了基于聚类的分布表示方法 [12]。2006 年之后，随着硬件性能的提升以及优化算法的突破，神经网络模型逐渐在各个领域中发挥出自己的优势，使用神经网络构造词表示的方法可以更灵活地对上下文进行建模，这类方法开始逐渐成为了词分布表示的主流方法。

到目前为止，基于分布假说的词表示方法，根据建模的不同，主要可以分为三类：基于矩阵的分布表示（2.1.1 小节）、基于聚类的分布表示（2.1.2 小节）和基于神经网络的分布表示（2.1.3 小节）。从广义上看，所有基于分布假说得到的表示均可称为分布表示（distributional representation），如上述的三种。而狭义的分布表示通常指基于矩阵的分布表示 [116]。本文以 Christopher Manning 的观点为基准[1]，文中出现的"分布表示"均指广义的分布表示，其它观点参考本

---

[1]Manning 在 2015 年深度学习暑期学校（蒙特利尔）中澄清，分布（distributional）意为使用上下文

章 2.3 小节的论述。

尽管这些不同的分布表示方法使用了不同的技术手段获取词表示，但由于这些方法均基于分布假说，它们的核心思想也都由两部分组成：一、选择一种方式描述上下文；二、选择一种模型刻画某个词（下文称"目标词"）与其上下文之间的关系。上文介绍的矩阵、聚类和神经网络三种方法，采用了不同的方式对上下文和目标词之间的关系进行建模。以下三个小节将简要介绍这三类模型。

### 2.1.1 基于矩阵的分布表示

基于矩阵的分布表示通常又称为分布语义模型（distributional semantic models）[3]，一些文献中也直接将其称作分布表示（distributional representation）[116]。这类方法需要构建一个"词-上下文"矩阵，从矩阵中获取词的表示。在"词-上下文"矩阵中，每行对应一个词，每列表示一种不同的上下文，矩阵中的每个元素对应相关词和上下文的共现次数。在这种表示下，矩阵中的一行，就成为了对应词的表示，这种表示描述了该词的上下文的分布。由于分布假说认为上下文相似的词，其语义也相似，因此在这种表示下，两个词的语义相似度可以直接转化为两个向量的空间距离。这类方法具体可以分为三个步骤：

一、选取上下文。最常见的有三种方法：第一种，将词所在的文档作为上下文，形成"词-文档"矩阵 [56]；第二种，将词附近上下文中的各个词（如上下文窗口中的 5 个词）作为上下文，形成"词-词"矩阵 [70, 90]；第三种，将词附近上下文各词组成的 $n$ 元词组（$n$-gram）作为上下文 [45]。在这三种方法中，"词-文档"矩阵非常稀疏，而"词-词"矩阵相对较为稠密，效果一般好于前者。"词-$n$ 元词组"相对"词-词"矩阵保留了词序信息，建模更精确，但由于比前者更稀疏，实际效果不一定能超越前者。

二、确定矩阵中各元素的值。"词-上下文"共现矩阵根据其定义，里面各元素的值应为词与对应的上下文的共现次数。然而直接使用原始共现次数作为矩阵的值在大多数情况下效果并不好 [3]，因此研究人员提出了多种加权和平滑方法，最常用的有 tf-idf、PMI 和直接取 log。

三、矩阵分解（可选）。在原始的"词-上下文"矩阵中，每个词表示为一个非常高维（维度是不同上下文的总个数）且非常稀疏的向量，使用降维技术可以

达语义。http://www.marekrei.com/blog/26-things-i-learned-in-the-deep-learning-summer-school/

将这一高维稀疏向量压缩成低维稠密向量。降维技术可以减少噪声带来的影响，但也可能损失一部分信息。最常用的分解技术包括奇异值分解（SVD）、非负矩阵分解（NMF）、典型关联分析（Canonical Correlation Analysis，CCA）[22, 23]、Hellinger PCA（HPCA）[59]。

基于矩阵的分布表示在这些步骤的基础上，衍生出了若干不同方法，如经典的 LSA [56] 就是使用"词-文档"矩阵，tf-idf 作为矩阵元素的值，并使用 SVD 分解，得到词的低维向量表示。在这类方法中，最新的为 GloVe 模型 [90]，下文简单介绍这一模型。

**Global Vector 模型（GloVe）**

总体上看，GloVe 模型是一种对"词-词"矩阵进行分解从而得到词表示的方法。矩阵第 $i$ 行第 $j$ 列的值为词 $v_i$ 与词 $v_j$ 在语料中的共现次数 $x_{ij}$ 的对数。在矩阵分解步骤，GloVe 模型借鉴了推荐系统中基于隐因子分解（Latent Factor Model）的方法 [4, 88]，在计算重构误差时，只考虑共现次数非零的矩阵元素，同时对矩阵中的行和列加入了偏移项。具体为最小化下式：

$$\sum_{i,j\in\mathbb{V},x_{ij}\neq 0} f\left(x_{ij}\right)\left(\log(x_{ij}) - \boldsymbol{p}_i^\mathrm{T}\boldsymbol{q}_j + \boldsymbol{b}_i^{(1)} + \boldsymbol{b}_j^{(2)}\right)^2 \tag{2.1}$$

其中 $\boldsymbol{p}_i$ 为词 $v_i$ 作为目标词时的词向量，$\boldsymbol{q}_j$ 为词 $v_j$ 作为上下文时的词向量，$\boldsymbol{b}^{(1)}$、$\boldsymbol{b}^{(2)}$ 为针对词表中各词的偏移向量，$f(x)$ 是一个加权函数，对低频的共现词对进行衰减，减少低频噪声带来的误差，定义为：

$$f(x) = \begin{cases} (x/x_{max})^\alpha & \text{如果 } x < x_{max} \\ 1 & \text{其它情况} \end{cases} \tag{2.2}$$

### 2.1.2　基于聚类的分布表示（分布聚类）

基于聚类的分布表示又称作分布聚类（distributional clustering）[91]，这类方法通过聚类手段构建词与其上下文之间的关系。其中最经典的方法是布朗聚类（Brown clustering）[12]。布朗聚类是一种层级聚类方法，聚类结果为每个词的多层类别体系。因此可以根据两个词的公共类别判断这两个词的语义相似度。

具体而言，布朗聚类需要最大化以下似然，其中 $c_i$ 为词 $w_i$ 对应的类别：

$$P(w_i|w_{i-1}) = P(w_i|c_i)P(c_i|c_{i-1})$$

布朗聚类只考虑了相邻词之间的关系，也就是说，每个词只使用它的上一个词，作为上下文信息。

除了布朗聚类以外，还有若干基于聚类的表示方法 [69, 91]。由于这类方法不是本文的重点，在此不再赘述。

### 2.1.3 基于神经网络的分布表示（词向量）

基于神经网络的分布表示一般称为词向量、词嵌入（word embedding）或分布式表示（distributed representation）[116]。神经网络词向量表示技术通过神经网络技术对上下文，以及上下文与目标词之间的关系进行建模。由于神经网络较为灵活，这类方法的最大优势在于可以表示复杂的上下文。在前面基于矩阵的分布表示方法中，最常用的上下文是词。如果使用包含词序信息的 $n$-gram 作为上下文，当 $n$ 增加时，$n$-gram 的总数会呈指数级增长，此时会遇到维数灾难问题。而神经网络在表示 $n$-gram 时，可以通过一些组合方式对 $n$ 个词进行组合，参数个数仅以线性速度增长。有了这一优势，神经网络模型可以对更复杂的上下文进行建模，在词向量中包含更丰富的语义信息。下一节将详细介绍不同神经网络模型是如何对上下文以及上下文与目标词之间的关系进行建模的。

## 2.2 神经网络词向量表示技术

神经网络词向量模型与其它分布表示方法一样，均基于分布假说，核心依然是上下文的表示以及上下文与目标词之间的关系的建模。构建上下文与目标词之间的关系，最自然的一种思路就是使用语言模型。从历史上看，早期的词向量只是神经网络语言模型的副产品。同时，神经网络语言模型对后期词向量的发展方向有着决定性的作用。因此，本节首先依照词向量模型的发展顺序介绍各模型的来龙去脉，其中 2.2.1 小节简述语言模型，2.2.2 到 2.2.4 小节介绍三种经典的神经网络语言模型，2.2.5 到 2.2.7 小节介绍专门用于生成词向量的神经网络模型。最后 2.2.8 小节从上下文的表示、上下文与目标词之间的关系归纳分析这些模型。

### 2.2.1 语言模型简介

语言模型可以对一段文本的概率进行估计，对信息检索、机器翻译、语音识别等任务有着重要的作用。

形式化讲，统计语言模型的作用是为一个长度为 $m$ 的字符串确定一个概率分布 $P(w_1, w_2, ..., w_m)$，表示其存在的可能性，其中 $w_1$ 到 $w_m$ 依次表示这段文本中的各个词。一般在实际求解过程中，通常采用下式计算其概率值：

$$P(w_1, w_2, ..., w_m) = P(w_1) \, P(w_2|w_1) \, P(w_3|w_1, w_2)$$
$$\ldots P(w_i \mid w_1, w_2, ..., w_{i-1}) \ldots P(w_m \mid w_1, w_2, ..., w_{m-1}) \quad (2.3)$$

在实践中，如果文本的长度较长，公式 2.3 右部 $P(w_i \mid w_1, w_2, \ldots, w_{i-1})$ 的估算会非常困难。因此，研究者们提出使用一个简化模型：$n$ 元模型（$n$-gram model）。在 $n$ 元模型中估算条件概率时，距离大于等于 $n$ 的上文词会被忽略，也就是对上述条件概率做了以下近似：

$$P(w_i \mid w_1, w_2, ..., w_{i-1}) \approx P(w_i \mid w_{i-(n-1)}, \ldots, w_{i-1}) \quad (2.4)$$

当 $n = 1$ 时又称一元模型（unigram model），公式 2.4 右部会退化成 $P(w_i)$，此时，整个句子的概率为：$P(w_1, w_2, ..., w_m) = P(w_1)P(w_2) \ldots P(w_m)$。从式中可以知道，一元语言模型中，文本的概率为其中各词概率的乘积。也就是说，模型假设了各个词之间都是相互独立的，文本中的词序信息完全丢失。因此，该模型虽然估算方便，但性能有限。

当 $n = 2$ 时又称二元模型（bigram model），将 $n$ 代入公式 2.4 中，右部为 $P(w_i|w_{i-1})$。常用的还有 $n = 3$ 时的三元模型（trigram model），使用 $P(w_i \mid w_{i-2}, w_{i-1})$ 作为近似。这些方法均可以保留一定的词序信息。

在 $n$ 元模型中，传统的方法一般采用频率计数的比例来估算 $n$ 元条件概率：

$$P(w_i \mid w_{i-(n-1)}, \ldots, w_{i-1}) = \frac{\text{count}(w_{i-(n-1)}, \ldots, w_{i-1}, w_i)}{\text{count}(w_{i-(n-1)}, \ldots, w_{i-1})} \quad (2.5)$$

其中，$\text{count}(w_{i-(n-1)}, \ldots, w_{i-1})$ 表示文本序列 $w_{i-(n-1)}, \ldots, w_{i-1}$ 在语料中出现的

次数。

为了更好地保留词序信息，构建更有效的语言模型，我们希望在 $n$ 元模型中选用更大的 $n$。但是，当 $n$ 较大时，长度为 $n$ 序列出现的次数就会非常少，在按照公式 2.5 估计 $n$ 元条件概率时，就会遇到数据稀疏问题，导致估算结果不准确。因此，一般在百万词级别的语料中，三元模型是比较常用的选择 [63]，同时也需要配合相应的平滑算法，进一步降低数据稀疏带来的影响 [33, 51]。

为了更好地解决 $n$ 元模型估算概率时遇到的数据稀疏问题，神经网络语言模型应运而生。

### 2.2.2 神经网络语言模型（NNLM）

Xu 等人在 2000 年首次尝试使用神经网络求解二元语言模型 [124]。2001年，Bengio 等人正式提出神经网络语言模型（Neural Network Language Model，NNLM）[6, 7]。该模型在学习语言模型的同时，也得到了词向量。

NNLM 同样也是对 $n$ 元语言模型进行建模，估算 $P(w_i \mid w_{i-(n-1)}, \ldots, w_{i-1})$ 的值。但与传统方法不同的是，NNLM 不通过计数的方法对 $n$ 元条件概率进行估计，而是直接通过一个神经网络结构，对其进行建模求解。图 2-1 展示了 NNLM 的基本结构。

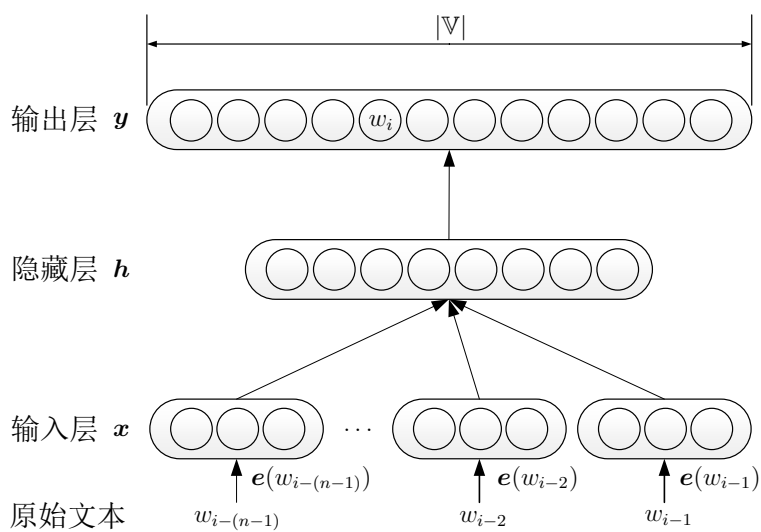

图 2-1　神经网络语言模型（NNLM）模型结构图

具体而言，对语料中一段长度为 $n$ 的序列 $w_{i-(n-1)}, \ldots, w_{i-1}, w_i$，$n$ 元语言

模型需要最大化以下似然：

$$P(w_i \mid w_{i-(n-1)}, \ldots, w_{i-1}) \tag{2.6}$$

其中，$w_i$ 为需要通过语言模型预测的词（目标词）。对于整个模型而言，输入为条件部分的整个词序列：$w_{i-(n-1)}, \ldots, w_{i-1}$，输出为目标词的分布。

神经网络语言模型采用普通的三层前馈神经网络结构，其中第一层为输入层。Bengio 提出使用各词的词向量作为输入以解决数据稀疏问题，因此输入层为词 $w_{i-(n-1)}, \ldots, w_{i-1}$ 的词向量的顺序拼接：

$$\boldsymbol{x} = [\boldsymbol{e}(w_{i-(n-1)}); \ldots; \boldsymbol{e}(w_{i-2}); \boldsymbol{e}(w_{i-1})] \tag{2.7}$$

当输入层完成对上文的表示 $\boldsymbol{x}$ 之后，模型将其送入剩下两层神经网络，依次得到隐藏层 $\boldsymbol{h}$ 和输出层 $\boldsymbol{y}$：

$$\boldsymbol{h} = \tanh(\boldsymbol{b}^{(1)} + H\boldsymbol{x}) \tag{2.8}$$

$$\boldsymbol{y} = \boldsymbol{b}^{(2)} + W\boldsymbol{x} + U\boldsymbol{h} \tag{2.9}$$

其中 $H \in \mathbb{R}^{|\boldsymbol{h}| \times (n-1)|\boldsymbol{e}|}$ 为输入层到隐藏层的权重矩阵，$U \in \mathbb{R}^{|\mathbb{V}| \times |\boldsymbol{h}|}$ 为隐藏层到输出层的权重矩阵，$|\mathbb{V}|$ 表示词表的大小，$|\boldsymbol{e}|$ 表示词向量的维度，$|\boldsymbol{h}|$ 为隐藏层的维度。$\boldsymbol{b}^{(1)}$、$\boldsymbol{b}^{(2)}$ 均为模型中的偏置项。矩阵 $W \in \mathbb{R}^{|\mathbb{V}| \times (n-1)|\boldsymbol{e}|}$ 表示从输入层到输出层的直连边权重矩阵。由于 $W$ 的存在，该模型可能会从非线性的神经网络退化成为线性分类器。Bengio 等人在文中指出，如果使用该直连边，可以减少一半的迭代次数；但如果没有直连边，可以生成性能更好的语言模型。因此在后续工作中，很少有使用输入层到输出层直连边的工作，下文也直接忽略这一项。如果不考虑 $W$ 矩阵，整个模型计算量最大的操作，就是从隐藏层到输出层的矩阵运算 $U\boldsymbol{h}$，后续的模型均有对这一操作的优化（2.2.3 节到 2.2.6 节）。

输出层一共有 $|\mathbb{V}|$ 个元素，依次对应下一个词为词表中某个词的可能性。这里将其中对应词 $w$ 的元素记作 $\boldsymbol{y}(w)$。由于神经网络的输出层并不直接保证各元素之和为 1，输出层的 $\boldsymbol{y}$ 并不是概率值。因此，在输出层 $\boldsymbol{y}$ 之后，需要加入 softmax 函数，将 $\boldsymbol{y}$ 转成对应的概率值：

$$P(w_i \mid w_{i-(n-1)}, \ldots, w_{i-1}) = \frac{\exp\left(\boldsymbol{y}(w_i)\right)}{\sum_{k=1}^{|\mathbb{V}|} \exp\left(\boldsymbol{y}(v_k)\right)} \tag{2.10}$$

    神经网络语言模型之所以能对 $n$ 元条件概率进行更好的建模，缓解数据稀疏问题，是由于它使用词序列的词向量对上文进行表示；而传统语言模型使用的是各词的独热表示作为上文的表示。词的独热表示 (one-hot representation) 是一个 $|\mathbb{V}|$ 维向量。对于词 $v_i$，其独热表示向量中，只有第 $i$ 维是 1，其余各维均是 0。在长度为 $n-1$ 的上文序列中，如果采用独热表示，则是一个 $|\mathbb{V}|^{n-1}$ 维的 0/1 向量，空间非常稀疏；而使用词向量表示时，则是一个 $(n-1)|\boldsymbol{e}|$ 维的实数向量。利用这种低维的实数表示，可以使相似的上文预测出相似的目标词，而传统模型中只能通过相同的上下文预测出相同的目标词。

    神经网络语言模型中的词向量出现在两个地方。在输入层中，各词的词向量存于一个 $|\boldsymbol{e}| \times |\mathbb{V}|$ 维的实数矩阵中。词 $w$ 到其词向量 $\boldsymbol{e}(w)$ 的转化就是从该矩阵中取出一列。值得注意的是，隐藏层到输出层的权重矩阵 $U$ 的维度为 $|\mathbb{V}| \times |\boldsymbol{h}|$，可以将其看做 $|\mathbb{V}|$ 个 $|\boldsymbol{h}|$ 维的行向量，其中的每一个向量，均可以看做某个词在模型中的另一个表示 $\boldsymbol{e}'$。在不考虑 $W$ 的情况下，每个词在模型中有两套词向量，其中 $\boldsymbol{e}(w)$ 为词 $w$ 作为上下文时的表示，而 $\boldsymbol{e}'(w)$ 为词 $w$ 作为目标词时的表示。由于 Bengio 等人的工作只考虑对语言模型的建模，词向量只是其副产品，因此他们并没有指出哪一套向量作为词向量效果更好。在其他关注词向量的工作中，通常只使用 $\boldsymbol{e}$ 作为词向量，具体在后文中会有讨论。

    如上文所述，输出层的分量 $\boldsymbol{y}(w_i)$ 描述的是在上文为 $w_{i-(n-1)}, \ldots, w_{i-1}$ 的条件下，下一个词为 $w_i$ 的可能性，该分量体现了上文序列与目标词之间的关系。通常，$\boldsymbol{y}(w_i)$ 又被称作能量函数，记作 $E(w_i; w_{i-(n-1):i-1})$：

$$\begin{aligned} \boldsymbol{y}(w_i) &= \boldsymbol{b}^{(2)} + \boldsymbol{e}'(w_i)^{\mathsf{T}} \tanh\left(\boldsymbol{b}^{(1)} + H\left[\boldsymbol{e}(w_{i-(n-1)}); \ldots; \boldsymbol{e}(w_{i-1})\right]\right) \\ &= E(w_i; w_{i-(n-1):i-1}) \end{aligned} \tag{2.11}$$

对于整个语料而言，语言模型需要最大化：

$$\sum_{w_{i-(n-1):i} \in \mathbb{D}} \log P(w_i \mid w_{i-(n-1)}, \ldots, w_{i-1}) \tag{2.12}$$

训练时，神经网络语言模型使用随机梯度下降法 [10] 来优化上述训练目标。每次迭代，随机从语料 $\mathbb{D}$ 中选取一段文本 $w_{i-(n-1)}, \ldots, w_i$ 作为训练样本，使用下式进行一次梯度迭代：

$$\theta \leftarrow \theta + \alpha \frac{\partial \log P(w_i \mid w_{i-(n-1)}, \ldots, w_{i-1})}{\partial \theta} \tag{2.13}$$

式中，$\alpha$ 是学习速率；$\theta$ 为模型中的所有参数，包括词向量和网络结构中的权重 $U$、$H$、$\boldsymbol{b}^{(1)}$、$\boldsymbol{b}^{(2)}$。

### 2.2.3　log 双线性语言模型（LBL）

2007 年，Mnih 和 Hinton 在神经网络语言模型（NNLM）的基础上提出了 log 双线性语言模型（Log-Bilinear Language Model，LBL）[79]。LBL 与 NNLM 的区别正如它们的名字所示，LBL 的模型结构是一个 log 双线性结构；而 NNLM 的模型结构为神经网络结构。具体来讲，LBL 模型的能量函数为：

$$\begin{aligned} E(w_i; w_{i-(n-1):i-1}) = \boldsymbol{b}^{(2)} &+ \boldsymbol{e}(w_i)^{\mathrm{T}} \boldsymbol{b}^{(1)} + \\ & \boldsymbol{e}(w_i)^{\mathrm{T}} H \left[ \boldsymbol{e}(w_{i-(n-1)}); \ldots; \boldsymbol{e}(w_{i-1}) \right] \end{aligned} \tag{2.14}$$

LBL 模型的能量函数（公式 2.14）与 NNLM 的能量函数（公式 2.11）主要有两个区别。一、LBL 模型中，没有非线性的激活函数 tanh，而由于 NNLM 是非线性的神经网络结构，激活函数必不可少；二、LBL 模型中，只有一份词向量 $\boldsymbol{e}$，也就是说，无论一个词是作为上下文，还是作为目标词，使用的是同一份词向量。其中第二点（只有一份词向量），只在原版的 LBL 模型中存在，后续的改进工作均不包含这一特点。

之后的几年中，Mnih 等人在 LBL 模型的基础上做了一系列改进工作。其中最重要的模型有两个：层级 log 双线性语言模型（Hierarchical LBL，HLBL）[80] 和基于向量的逆语言模型（inverse vector LBL，ivLBL）[81]。以下分别介绍这两个模型所用的技术。

#### 层级 softmax

HLBL 模型 [80] 采用了 Bengio 在 2005 年提出的层级 softmax 函数 [82]，加速了目标层的求解。传统的 softmax 函数如公式 2.10 ，分母所示的归一化项，由

于需要得到 $y$ 中的每个分量的值，计算非常耗时。而层级 softmax 函数通过构造一个树形结构，使求解某个分量对应的概率值时，只需要 $O(\log(|\mathbb{V}|))$ 次运算，而不需要此前的 $O(|\mathbb{V}|)$ 次运算，极大地降低了模型的时间复杂度。

**噪声对比估算**

    2013 年提出的基于向量的逆语言模型 (ivLBL) [81] 虽然名字上仍然继承了 log 双线性语言模型 (LBL)，实际上已经抛弃了 log 双线性结构，转而采用了向量点积结构。该模型在 Skip-gram 模型（见 2.2.6 节）的基础上，使用噪声对比估算（noise-contrastive estimation，NCE）[34] 加速 softmax 的估计，将 softmax 的复杂度降低到常数级别。

### 2.2.4 循环神经网络语言模型（RNNLM）

    2.2.2 节提到的 NNLM 以及 2.2.3 提到的 LBL 模型均为 $n$ 元模型。Mikolov 等人提出的循环神经网络语言模型（Recurrent Neural Network based Language Model，RNNLM）则直接对 $P(w_i \mid w_1, w_2, ..., w_{i-1})$ 进行建模，而不使用公式 2.4 对其进行简化 [72, 74]。因此，RNNLM 可以利用所有的上文信息，预测下一个词，其模型结构如图 2-2 所示。

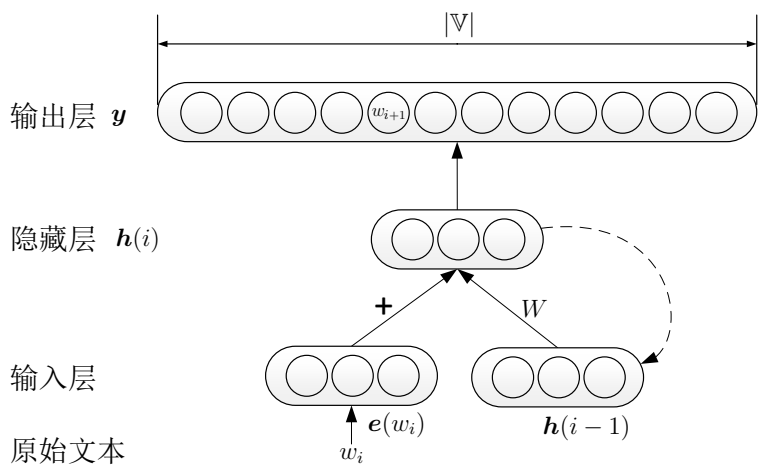

图 2-2   循环神经网络语言模型（RNNLM）模型结构图

RNNLM 的核心在于其隐藏层的算法：

$$\boldsymbol{h}(i) = \phi(\boldsymbol{e}(w_i) + W\boldsymbol{h}(i-1)) \tag{2.15}$$

其中，$\phi$ 为非线性激活函数。该式对应 NNLM 的公式 2.8。但与 NNLM 不同，RNNLM 并不采用 $n$ 元近似，而是使用迭代的方式直接对所有上文进行建模。在公式 2.15 中，$\boldsymbol{h}(i)$ 表示文本中第 $i$ 个词 $w_i$ 所对应的隐藏层，该隐藏层由当前词的词向量 $\boldsymbol{e}(w_i)$ 以及上一个词对应的隐藏层 $\boldsymbol{h}(i-1)$ 结合得到。

隐藏层的初始状态为 $\boldsymbol{h}(0)$，随着模型逐个读入语料中的词 $w_1, w_2, \ldots,$ 隐藏层不断地更新为 $\boldsymbol{h}(1), \boldsymbol{h}(2), \ldots$。根据公式 2.15，每一个隐藏层包含了当前词的信息以及上一个隐藏层的信息。通过这种迭代推进的方式，每个隐藏层实际上包含了此前所有上文的信息，相比 NNLM 只能采用上文 $n$ 元短语作为近似，RNNLM 包含了更丰富的上文信息，也有潜力达到更好的效果。

RNNLM 的输出层计算方法与 NNLM 的输出层一致，可见公式 2.9。

### 2.2.5 C&W 模型

与前面的三个基于语言模型的词向量生成方法不同，Collobert 和 Weston 在 2008 年提出的 C&W 模型 [17]，是第一个直接以生成词向量为目标的模型。

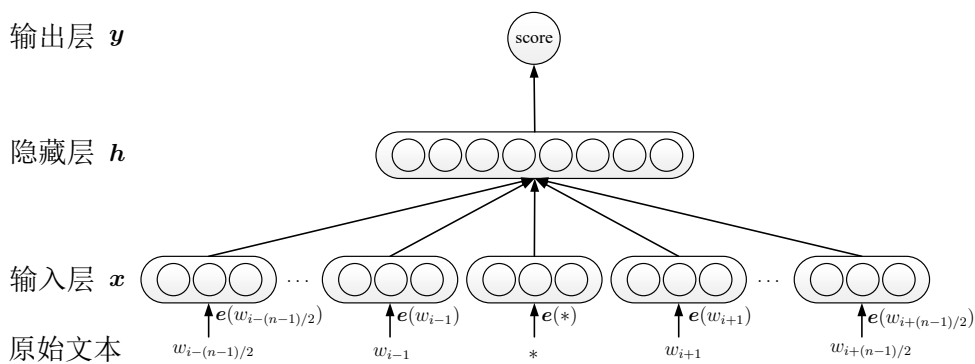

图 2-3 C&W 模型结构图

语言模型的目标为求解 $P(w_i \mid w_1, w_2, ..., w_{i-1})$，其中隐藏层到输出层的矩阵运算是最耗费时间的部分。因此，前面的各个词向量模型中，几乎都有对这一部分做优化的步骤，如层级 softmax、分组 softmax 和噪声对比估算。C&W 模型的目标是更快速地生成词向量，因此它并没有采取语言模型的方式，去求解上述条件概率，转而采用了另一种更高效的方法，直接对 $n$ 元短语打分。对于语料中出现过的 $n$ 元短语，模型会对其打高分；而对于语料中没有出现的随机短语，模型会对其打低分。通过这种方式，C&W 模型可以更直接地学习得到符合分布假说的词向量。

具体而言，对于整个语料，C&W 模型需要最小化：

$$\sum_{(w,c)\in\mathbb{D}}\sum_{w'\in\mathbb{V}}\max(0, 1 - \text{score}(w,c) + \text{score}(w',c)) \tag{2.16}$$

其中，$(w,c)$ 为从语料中选出的一个 $n$ 元短语 $w_{i-(n-1)/2},\ldots,w_{i+(n-1)/2}$，一般 $n$ 为奇数，以保证上文和下文的词数一致；$w$ 为序列中的中间词，即 $w_i$，在该模型中为目标词；$c$ 表示目标词 $w$ 的上下文；$w'$ 为字典中的某一个词。模型采用 pairwise 的方式 [16] 对文本片段进行优化，希望正样本的打分要比负样本的打分至少高 1 分。正样本 $(w,c)$ 来自语料，而负样本 $(w',c)$ 则是将正样本序列中的中间词替换成其它词。形式化地，目标词 $w$ 和上下文 $c$ 分别为：

$$\begin{aligned} w &= w_i \\ c &= w_{i-(n-1)/2},\ldots,w_{i-1},w_{i+1},\ldots,w_{i+(n-1)/2} \end{aligned} \tag{2.17}$$

代入公式 2.16 中，正负样本分别为：

$$\begin{aligned} (w,c) &= w_{i-(n-1)/2},\ldots,w_{i+(n-1)/2} \\ (w',c) &= w_{i-(n-1)/2},\ldots,w_{i-1},w',w_{i+1},\ldots,w_{i+(n-1)/2} \end{aligned} \tag{2.18}$$

在大多数情况下，把一个普通短语的中间词随机替换成其它词，得到的都是不正确的短语，所以这样构造的负样本是有效的（多数情况下确实是负样本，极少数情况下把正确的短语当作负样本，但是不会影响模型整体的效果）。同时，由于负样本仅仅是修改了正样本中的一个词，也不会让分类面距离负样本太远而影响分类效果。

    C&W 模型 (图 2-3) 与 NNLM (图 2-1) 相比，主要的不同点在于 C&W 模型将目标词放到了输入层，同时输出层也从语言模型的 $|\mathbb{V}|$ 个节点变为一个节点，这个节点的数值表示对这组 $n$ 元短语的打分。打分只有高低之分，没有概率的特性，因此无需复杂的归一化操作。C&W 模型使用这种方式把 NNLM 模型在最后一层的 $|\mathbb{V}| \times |\boldsymbol{h}|$ 次运算降为 $|\boldsymbol{h}|$ 次运算，极大地降低了模型的时间复杂度。这个区别使得 C&W 模型成为神经网络词向量模型中最为特殊的一个，其它模型的目标词均在输出层，只有 C&W 模型的目标词在输入层。由这一改变带来的影响，可以参见第三章 3.3.2 节的分析。

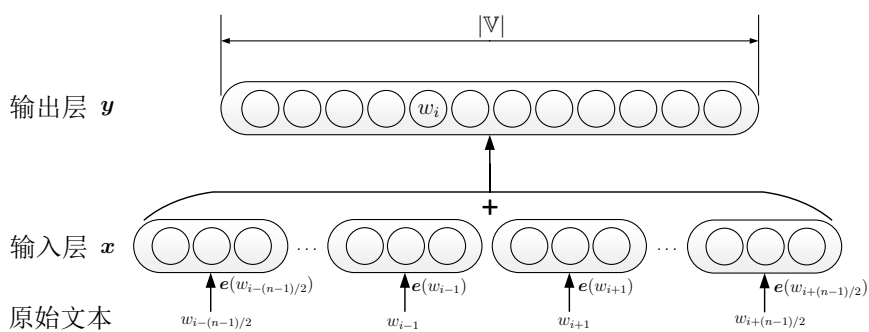

图 2-4 CBOW 模型结构图

### 2.2.6 CBOW 模型和 Skip-gram 模型

Mikolov 等人在 2013 年的文献 [73] 中，同时提出了 CBOW（Continuous Bag-of-Words）和 Skip-gram 模型。他们设计两个模型的主要目的是希望用更高效的方法获取词向量。因此，他们根据前人在 NNLM、RNNLM 和 C&W 模型上的经验，简化现有模型，保留核心部分，得到了这两个模型。

### CBOW 模型

CBOW 模型的结构如图 2-4，该模型一方面根据 C&W 模型的经验，使用一段文本的中间词作为目标词；另一方面，又以 NNLM 作为蓝本，并在其基础上做了两个简化。一、CBOW 没有隐藏层，去掉隐藏层之后，模型从神经网络结构直接转化为 log 线性结构，与 Logistic 回归一致。log 线性结构比三层神经网络结构少了一个矩阵运算，大幅度地提升了模型的训练速度。二、CBOW 去除了上下文各词的词序信息，使用上下文各词词向量的平均值[2]，代替神经网络语言模型使用的上文各词词向量的拼接。形式化地，CBOW 模型对于一段训练样本 $w_{i-(n-1)}, \ldots, w_i$，输入为：

$$\boldsymbol{x} = \frac{1}{n-1} \sum_{w_j \in c} \boldsymbol{e}(w_j) \tag{2.19}$$

---

[2]Mokolov 在文献 [73] 中使用的是求和，但是在后来 word2vec 的工具包中将求和改进为平均值。

由于没有隐藏层，CBOW 模型的输入层直接就是上下文的表示。CBOW 模型根据上下文的表示，直接对目标词进行预测：

$$P(w|c) = \frac{\exp\left(\boldsymbol{e}'(w)^\mathsf{T}\boldsymbol{x}\right)}{\sum_{w' \in \mathbb{V}} \exp\left(\boldsymbol{e}'(w')^\mathsf{T}\boldsymbol{x}\right)} \tag{2.20}$$

上述二式中，目标词 $w$ 和上下文 $c$ 的定义与 C&W 模型（2.2.5 节）中的公式 2.17 一致。对于整个语料而言，与神经网络语言模型类似，CBOW 的优化目标为最大化：

$$\sum_{(w,c) \in \mathbb{D}} \log P(w|c) \tag{2.21}$$

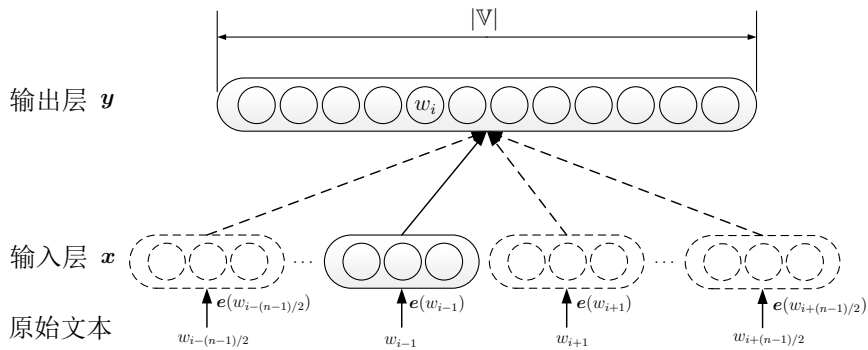

图 2-5　Skip-gram 模型结构图

**Skip-gram 模型**

Skip-gram 模型的结构如图 2-5，与 CBOW 模型一样，Skip-gram 模型中也没有隐藏层。和 CBOW 模型不同的是，Skip-gram 模型每次从目标词 $w$ 的上下文 $c$ 中选择一个词，将其词向量作为模型的输入 $\boldsymbol{x}$，也就是上下文的表示。Skip-gram 模型同样通过上下文预测目标词[3]，对于整个语料的优化目标为最大化：

$$\sum_{(w,c) \in \mathbb{D}} \sum_{w_j \in c} \log P(w|w_j) \tag{2.22}$$

---

[3]本文为了各模型之间的一致性，将 Skip-gram 模型描述成通过上下文预测目标词，而在 Skip-gram 的论文 [73] 中将模型描述成通过目标词预测上下文。由于模型需要遍历整个语料，任意一个窗口中的两个词 $w_a, w_b$ 都需要计算 $P(w_a|w_b) + P(w_b|w_a)$，因此这两种描述方式是等价的。

其中，

$$P(w|w_j) = \frac{\exp\left(\boldsymbol{e}'(w)^{\mathsf{T}}\boldsymbol{e}(w_j)\right)}{\sum_{w' \in \mathbb{V}} \exp\left(\boldsymbol{e}'(w')^{\mathsf{T}}\boldsymbol{e}(w_j)\right)} \tag{2.23}$$

**负采样技术**

Mikolov 等人在 2013 年提出了负采样技术（negative sampling），进一步提升了最后一层的效率 [75]。负采样技术借鉴了 C&W 模型采用的构造负样本的方法，还参考了 ivLBL 模型所用的 NCE 方法，最后构造出了一个优化目标，最大化正样本的似然，同时最小化负样本的似然。

负采样技术与 C&W 模型中相应部分的区别主要是，负采样技术不采用 pairwise 的方式训练，因此，一个正样本可以对应多个负样本，Mikolov 等人在实验中论述了使用多个负样本（一般选 5）能有效提升模型的性能。

负采样技术与 NCE 技术的主要区别是，负采样技术仅仅是优化正负样本的似然，而不对输出层做概率归一化。NCE 技术则是通过噪声样本对概率进行估计。在实验中，Mikolov 等人也论述了负采样技术相比 NCE 技术少了一些约束，对于生成词向量，是有帮助的 [75]。

**二次采样技术**

在大规模语料中，高频词通常就是停用词（如英语中的"the"、汉语中的"的"）。一方面，这些高频词只能带来非常少量的语义信息，比如几乎所有的词都会和"的"共同出现，但是并不能说明这些词的语义相似。另一方面，训练高频词本身占据了大量的时间，但在迭代过程中，这些高频词的词向量变化并不大。Mikolov 等人为了解决这一问题，提出了二次采样技术（subsampling）[75]，具体而言，如果词 $w$ 在语料中的出现频率 $f(w)$ 大于阈值 $t$，则有 $P(w)$ 的概率在训练时跳过这个词。

$$P(w) = 1 - \sqrt{\frac{t}{f(w)}} \tag{2.24}$$

word2vec 工具包的实现与论文 [75] 中的公式稍有不同：

$$P(w) = \frac{f(w) - t}{f(w)} - \sqrt{\frac{t}{f(w)}} \tag{2.25}$$

无论采用论文中的公式，还是工具包实现的公式，词出现的越频繁，就越有可能在训练中被跳过。这种二次采样技术不仅可以提升词向量的训练速度，大多数情况下也能提升词向量的性能 [65, 75]。

### 2.2.7　Order 模型

2.2.6 节提到的 CBOW 模型和 Skip-gram 模型为了有更高的性能，在神经网络语言模型或者 log 双线性语言模型的基础上，同时去掉了隐藏层和词序信息。为了更好地分析词序信息对词向量性能的影响，这里提出一个新模型，名为"Order"，意为保留了词序信息。该模型在保留词序信息的同时，去除了隐藏层，其结构如图 2-6 所示。

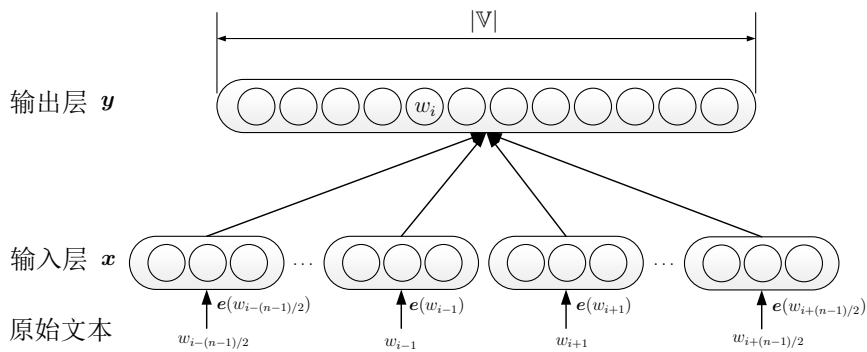

图 2-6　Order 模型结构图

相对 CBOW 模型，Order 模型使用上下文词向量的拼接作为模型的输入，形式化地：

$$\boldsymbol{x} = \left[ \boldsymbol{e}(w_{i-(n-1)}); \ldots ; \boldsymbol{e}(w_{i-(n-1)/2-1}); \boldsymbol{e}(w_{i-(n-1)/2+1}); \ldots ; \boldsymbol{e}(w_i) \right] \qquad (2.26)$$

相对 log 双线性语言模型，Order 模型采用了公式 2.20（与 CBOW 模型一致），直接从上下文的表示预测目标词。

### 2.2.8　词向量模型的理论比较

本节从上下文表示以及上下文与目标词之间的关系两个角度分析各个神经网络词向量模型。

## 上下文表示

上文介绍的各种神经网络词向量模型中，除了 Skip-gram 模型使用词作为上下文表示之外，其它模型均使用 $n$-gram 作为上下文表示，而这些表示使用不同的组合策略构造 $n$-gram 的表示。如 CBOW 模型使用 $n$-gram 中各词词向量的平均值作为上下文表示；Order 模型使用 $n$-gram 中各词词向量的拼接作为上下文表示，这种方法可以看做词向量的线性组合；LBL 模型则是直接对 $n$-gram 中各词的词向量做了线性变换；NNLM 和 C&W 模型更是做了非线性变换。这些不同的策略可以从复杂度的角度进行分析，本文具体从三个角度来观察这些模型：模型结构的复杂度、模型参数的个数以及模型求解的时间复杂度。

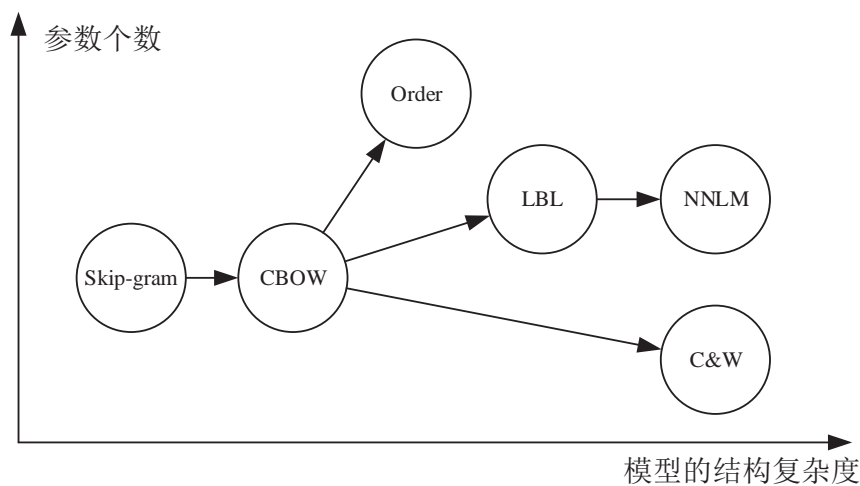

图 2-7　神经网络词向量模型复杂程度对比图

图 2-7 展示了各词向量模型复杂度的关系图。图中箭头方向表示各模型复杂程度的拓扑序，从简单的模型指向复杂的模型。水平方向的相对位置表示模型结构的复杂程度，从简单到复杂。垂直方向的相对位置表示模型的参数个数，越靠上的模型参数越多。

## 模型结构的复杂度

从模型结构上看，CBOW 模型与 Skip-gram 模型相比，采取了更复杂的上下文表示方法，用上下文词向量的线性叠加代替了随机选取其中一个词的词向量。Order 相比 CBOW 模型在上下文表示时，保留了词序信息。LBL 在保留词序信息的同时，还进一步使用线性变换，使模型具有上下文的语义组合能力。NNLM

与 C&W 模型进一步采用了非线性激活函数，使得整个模型为神经网络结构，表达能力强于 LBL 的双线形结构。

这些模型中，虽然从模型结构上看，在图 2-7 中有从左往右的绝对顺序，但是如果看模型整体的复杂度，并不能直接断言 LBL 模型比 Order 更复杂，或者 C&W 模型比 LBL 更复杂。这是因为这些模型在保证词向量维度相同时，使用的参数个数有很大的差异，更复杂的结构或者更多的参数都有可能导致模型的复杂度增加。

## 参数个数

从参数个数上看，参数最少的模型是 C&W。在神经网络词向量模型中，参数的个数主要包含词向量和网络结构中的其它参数这两部分。C&W 模型只有一份词向量，而其它模型均维护了两份词向量，因此 C&W 模型的参数个数最少，为 $|e| \times |\mathbb{V}| + (win+1)|h|$。其中 $|e|$ 为词向量的维度，$|\mathbb{V}|$ 为词表大小，$|h|$ 为隐藏层的维度，$win$ 为上下文窗口的大小。Skip-gram 与 CBOW 模型只使用了两份词向量，而没有其它额外的模型参数，参数个数为 $2|e| \times |\mathbb{V}|$。LBL 和 NNLM 这两个模型在 CBOW 等模型的基础上加入了隐藏层，因此增加了一个输入层到隐藏层的权重矩阵，其参数个数为 $(|e|+|h|) \times |\mathbb{V}| + (win-1)|e| \times |h|$。参数最多的 Order 模型由于既保持了上下文的词序信息，又采用了线性结构，因此在模型中，词当作目标词时的词向量 $e'$，其维度需要与作为上下文时的词向量的维度和一致，因此总参数个数为 $win|e| \times |\mathbb{V}|$。

## 时间复杂度

从模型的时间复杂度上看，文献 [73] 对早期的若干模型均有分析，因此这里只做简要介绍。基于神经网络的词向量模型均通过扫描语料中的每一个词，取该词以及其周围的上下文作为一个样本。因此对比这些模型时，可以只分析训练一个样本的时间复杂度。对于原版的 NNLM 和 LBL 模型，训练一个样本需要的计算为输入层到隐藏层，隐藏层到输出层这两个矩阵运算，其复杂度为 $O((win-1)|e| \times |h| + |h| \times |\mathbb{V}|)$。对于 Order 模型，由于省略了隐藏层，其复杂度为 $O((win-1)|e| \times |\mathbb{V}|)$。对于 CBOW 和 Skip-gram 模型，由于进一步忽略了词序信息，其复杂度为 $O(|e| \times |\mathbb{V}|)$。C&W 模型在结构中省去了对目标词的预测，其复杂度仅为 $O((win+1)|e| \times |h|)$。对于上述各个模型，如果采用层级 softmax 函数做输出层的优化，式子中的 $|\mathbb{V}|$ 可以加速到 $\log(|\mathbb{V}|)$，而如果使用噪声对比

估算（noise-contrastive estimation）或者负采样技术（negative sampling），$|V|$ 可以进一步优化到常数。因此，如果这些模型都使用先进的负采样技术预估输出层，则这些模型在时间复杂度上的排序，与模型结构的复杂度一致，从图 2-7 中看，从左往右时间复杂度依次递增。

**效率和性能**

CBOW、Skip-gram 和 Order 模型相对于其它神经网络模型，均去除了隐藏层。如果有隐藏层，输入层的上下文表示可以通过一个线性变换或者非线性变换得到隐藏层，这种操作属于语义组合操作 [36]。如果没有隐藏层，上下文词之间的关系为普通的线性叠加关系，会丢失部分语义信息。

CBOW 模型和 Skip-gram 模型还通过不同的方法去掉了其它神经网络模型中保留的词序信息。虽然这两个模型根据上下文各词与目标词之间的距离采用了加权策略，可以少量保留词序信息，但是这种策略相对于词向量顺序拼接的方式，可以认为几乎忽略了词序信息。

这些模型采取的简化策略，使其有更高的运行效率，可以在更大规模的语料上训练词向量，但是模型本身对语义捕获的能力也有一些降低。这些简化究竟对词向量的性能有多少影响，需要通过实验来说明。

**目标词与上下文之间的关系**

现有词向量模型的目标词与上下文之间主要有两种关系。从神经网络的目标上看，C&W 模型与众不同。C&W 模型的目标函数是求目标词 $w$ 与其上下文 $c$ 的联合打分，而其他模型均为根据上下文 $c$，预测目标词 $w$。从各个模型的结构，以及对目标词与上下文的处理方式看，C&W 模型将上下文和目标词同时放在输入层，通过神经网络模型优化它们之间的关系。而其它的神经网络模型只把上下文放在输入层，把目标词放在输出层。C&W 模型使用神经网络模型构建目标词和上下文的组合关系，而其它神经网络模型的上下文和目标词之间呈现预测关系。

如果从能量函数来看，C&W 模型和其它模型之间的区别会更为清晰。比如 LBL 模型的能量函数（公式 2.14）中，主要部分为：

$$E(w;c) = \boldsymbol{e}(w)^{\mathsf{T}} H \boldsymbol{e}(c) \tag{2.27}$$

而根据公式 2.23 可得，Skip-gram 模型的能量函数为：

$$E(w; c) = \boldsymbol{e}'(w)^{\mathsf{T}}\boldsymbol{e}(c) \tag{2.28}$$

在这些模型中，上下文对应的向量与目标词对应的向量，通过点积或双线性计算，均有"交互"关系，而在 C&W 模型中，上下文与目标词的向量仅为加法"组合"关系：

$$E(w; c) = A\boldsymbol{e}(w) + B\boldsymbol{e}(c) \tag{2.29}$$

无论是上下文与目标词呈组合关系的 C&W 模型，还是预测目标词，目标词与上下文之间有交互关系的其它神经网络模型，均符合分布假说。但是由于这两种方式的表达能力不同，因此对于语义的捕获能力可能也会有差异。第三章实验部分会对这两种不同方法做进一步的分析。

## 2.3  相关工作

本文将分布表示归类为基于矩阵的分布表示、基于聚类的分布表示和基于神经网络的分布表示。这种分类方式基本沿用了 Turian 等人的分类 [116]。在 Turian 的分类中，这三种方法分别被称作"distributional representation"、"clustering-based word representation"和"distributed representation"。本文和 Turian 的区别在于，本文将这三种方法统称为"distributional representation"，而将其中的第一种方法称作基于矩阵的分布表示。

Baroni 等人的文献 [3] 分析了语义向量模型，因此没有考虑聚类模型。文中，Baroni 将基于矩阵的分布表示称作计数（count）方法，而将神经网络模型称作预测（predict）方法。

Sahlgren 的博士论文 [98] 研究了词空间模型（word space model），具体分析了基于矩阵的分布表示中的"词-文档"矩阵和"词-词"矩阵，并得到结论："词-文档"矩阵主要构建了词的组合关系（syntagmatic），而"词-词"矩阵主要构建了词的替换关系（paradigmatic）。

Turney 和 Pantel 的工作 [118] 总结了向量空间模型（vector space model），他们将向量空间模型分为"词-文档"矩阵、"词-上下文"矩阵和"词对-模板"矩阵这三种，认为"词-文档"矩阵适合用来表示文档，"词-词"矩阵适合表示词，而"词对-模板"矩阵用来表示词对之间的关系。

## 2.4 模型总结

本章介绍了三类不同的分布表示方法：基于矩阵的分布表示、基于聚类的分布表示和基于神经网络的分布表示。其中的经典模型总结于表 2-1 中。这三类模型中，基于聚类的模型较为特别，将词表示为聚类的类标。如果采用层级聚类方法，可以根据聚类类别的公共前缀衡量词之间的相似度。而另外两类模型得到的都是向量表示，可以直接使用余弦距离、欧氏距离等向量空间距离衡量指标来检测词义的相似度。

| 名称 | 上下文 | 上下文与目标词之间的建模（技术手段） |
|---|---|---|
| LSA/LSI [56] | 文档 | 矩阵 |
| HAL [70] | 词 | |
| GloVe [90] | 词 | |
| Jones & Mewhort [45] | n-gram | |
| Brown Clustering [12] | 词 | 聚类 |
| Skip-gram [73] | 词 | 神经网络 |
| CBOW [73] | n-gram（加权） | |
| Order（2.2.7 小节） | n-gram（线性组合） | |
| LBL [79] | n-gram（线性组合） | |
| NNLM [7] | n-gram（非线性组合） | |
| C&W [17] | n-gram（非线性组合） | |

表 2-1　分布表示模型的概要信息

基于矩阵的模型和基于神经网络的模型最终都可以得到词的低维向量表示，这两类方法是否存在一定的联系？答案是肯定的。Levy 和 Goldberg 证明了对"词-词"矩阵做 SVD 分解与 Skip-gram 模型配合负采样技术优化具有相同的最优解 [64]。本文证明了"词-词"矩阵分解与 Skip-gram 模型的原始形式具有相同的最优解，具体证明过程见附录 A。遗憾的是，这两个等价关系需要一个前提条件：最优解可以取到。在实际优化过程中，一般的优化方法只能不断逼近最优解，而很难达到最优解。尽管理论上这两者的最终优化结果应当一致，但是在实践中还是有较大的差异，文献 [3, 64, 65] 比较了这些模型在实践中的差异。2015 年 Li 等人在文献 [67] 中证明了"词-词"矩阵分解与 Skip-gram 模型可以完全等价。这项工作抛弃了 SVD 所用的均方根误差来作为矩阵分解的重构

误差，而是采用了一种对数概率误差。他们的实验也证明，这两者可以达到相同的效果。

这些证明建立了"词-词"矩阵与 Skip-gram 模型之间的联系，这两个模型只是矩阵和神经网络这两种技术手段下的特例，它们都选用了词作为上下文。而神经网络相对矩阵表示的优势正是可以通过组合手段，对上下文进行更为复杂的建模，同时避免维数灾难问题。下一章将会详细分析各个神经网络词向量模型在实践中的差别。

# 第三章　词向量表示技术的实验分析

在使用深度学习技术解决自然语言处理任务时，最基础的问题是词的表示。本章从理论和实验两个角度分析和探讨了构建词向量表示时的关键点：模型、语料以及参数的选择，并给出了若干条生成优质词向量的参考建议。

## 3.1　引言

词向量模型可以从大规模无标注语料中自动学习到句法和语义信息 [76]。近年来，大量研究者投身到设计新型的词向量模型中，基于词向量的神经网络模型也为多项自然语言处理任务带来了性能的提升，甚至在多项任务中达到了目前最好的效果。现有的词向量模型在提出时，作者均声称他们的方法比前人的方法好，然而这些工作在评价词向量时，会挑选比较局限的评价指标，有时候甚至使用了不同的训练语料，因此其评价结果可能会缺乏借鉴意义。现有的词向量模型纷繁复杂，这些模型究竟哪个更好，或者在什么情况下更适合用哪个模型，现有的研究工作仍然缺乏相应的比较分析。本章主要分析词向量生成中的几个关键点，包括词向量模型的选择、语料的选择以及训练参数的选取，对各现有的词向量模型进行全面的比较。

| 模型 | 上下文的表示 | 目标词与其上下文关系 |
|---|---|---|
| Skip-gram [73] | 上下文中某一个词的词向量 | 上下文预测目标词 |
| CBOW [73] | 上下文各词词向量的平均值 | 上下文预测目标词 |
| Order | 上下文各词词向量的拼接 | 上下文预测目标词 |
| LBL [79] | 上下文各词的语义组合 | 上下文预测目标词 |
| NNLM [7] | 上下文各词的语义组合 | 上下文预测目标词 |
| C&W [17] | 上下文各词与目标词的语义组合[1] | 上下文和目标词联合打分 |

表 3-1　各词向量模型在目标词与上下文建模上的异同

为了选择最合适的词向量模型，首先需要对现有词向量模型的结构进行剖析。根据第二章的分析，可以从上下文的表示以及目标词与上下文的关系这两个角度进行对比分析。表 3-1 列举了神经网络词向量模型在这两方面的异同。例

---

[1]该模型联合处理上下文和目标词，不存在一个独立的上下文表示。

如，Skip-gram 模型 [73] 选取了目标词 $w$ 上下文中的某一个词的词向量，作为其上下文表示。CBOW 模型采用上下文各词词向量的平均值作为上下文表示。这两个模型为了加速，均忽略了词序信息。然而，Landauer 在文献 [54] 中曾分析，文本中大约有 20% 的语义来自于词序，而剩下部分来自词的选择。因此，这两个模型可能会丢失一些重要信息。与之相对的，LBL 模型 [79]、NNLM 模型 [7]、C&W 模型均使用上下文窗口中各词词向量的拼接作为上下文的表示，可以保留词序信息。本文希望知道，**使用哪个模型效果更好？具体而言，在多种不同的上下文表示之中，以及在上下文与目标词的两种不同的关系之间，应当如何选择合适的模型？** (问题一)

同时，词向量的精度非常依赖于训练语料的选择，不同大小、不同领域的语料会极大地影响词向量的性能。因此，本文还希望知道，**训练语料的大小及领域对词向量有什么样的影响？** (问题二)

除了模型和语料的选择，现有的词向量算法也非常依赖于参数的选择。最主要的是模型的迭代次数，以及词向量的维度。对于迭代次数，如果迭代次数过少，词向量就会训练不充分，所包含的信息不足；如果迭代次数过多，模型很可能过拟合。对于词向量的维度，本文也希望找到合适的维度。因此，本文尝试分析**在迭代训练中，选择什么样的迭代次数可以获得足够好的词向量，同时避免过拟合？** (问题三) 以及**多少维的词向量效果最理想？** (问题四)

为了更客观地回答上述四个问题，本文选取了三大类指标，共八个具体的任务评价这些词向量。本文认为这些评价指标涵盖了现有词向量的所有用法。第一类指标为评价词向量的语言学特性。本文使用标准的 WordSim353 数据集 [28] 以及 TOEFL 数据集 [55] 评价词向量的空间距离是否和人的直观感受一致。第二类指标中，本文将词向量作为现有自然语言任务中的特征，看其所能达到的性能。具体而言，本文选择了文本分类任务和命名实体识别任务。第三类指标中，本文将词向量作为神经网络模型的初始值，并使用卷积神经网络做情感分类任务，以及使用文献 [18] 的模型做词性标注任务。通过使用这些不同的评价指标对词向量模型进行评价，本文尝试分析出应该怎么选择模型 (问题一) 和参数 (问题三和四)。本文进一步通过使用不同规模和不同领域的语料对词向量进行训练尝试回答第二个问题。

本章的主要贡献为系统化地整理现有的词向量模型，并且通过多种评价指标全面地比较分析各词向量模型与语料的选取。通过这些实验比较和理论分析，

本文给出了若干条生成词向量的参考建议。

1. 选择一个合适领域的语料，在此前提下，语料规模越大越好。使用大规模的语料进行训练，可以普遍提升词向量的性能，如果使用领域内的语料，对同领域的任务会有显著的提升。

2. 选择一个合适的模型。复杂的模型相比简单的模型，在较大的语料中才有优势。简单的模型在绝大多数情况下已经足够好。预测目标词的模型（表 3-1 中除了 C&W 以外的所有模型）比目标词与上下文呈组合关系的模型（C&W 模型）在多个任务中有更好的性能。

3. 训练时，迭代优化的终止条件最好根据具体任务的验证集来判断，或者近似地选取其它类似的任务作为指标，但是不应该选用训练词向量时的损失函数。

4. 词向量的维度一般需要选择 50 维及以上，特别当衡量词向量的语言学特性时，词向量的维度越大，效果越好。

## 3.2　评价方法

为了更全面地对各种不同的词向量模型进行评价，本文考察了各种词向量的用法，并将这些用法分成三大类，分别为：一、利用词向量的语言学特性完成任务；二、将词向量作为特征，提高自然语言处理任务的性能；三、将词向量作为神经网络的初始值，提升神经网络模型的优化效果。本文在这三大类用法的基础上，选取了八个有代表性的具体任务，作为词向量的评价指标。

### 3.2.1　词向量的语言学特性

各词向量模型均基于分布假说设计而成，因此无论哪种词向量模型，都会符合分布假说所提出的性质：具有相似上下文的词，会拥有相似的语义，并且其词向量的空间距离更接近。文献 [3] 通过语义相关性、同义词判别、概念分类和类比等实验论述了词向量具有各种不同的语言学特性。本文从中选取了三个代表性任务。

**语义相关性 (ws)**

衡量语义相关性最经典的是 WordSim353 数据集，该数据集包含了 353 个词对，其中每一个词对有至少十位标注者对其进行 0 到 10 之间的打分，分数越高表示标注人员认为这两个词的语义更相关或者更相似。例如，词对"student, professor"的平均打分为 6.81，而词对"professor, cucumber"的打分为 0.31。评价时，对于每个词对，本文使用所有标注者打分的平均值作为参考得分 $X$，以词对的两个词向量的余弦距离作为模型得到的相关性得分 $Y$，并衡量这两组数值之间的皮尔逊相关系数。皮尔逊相关系数衡量了两个变量之间的线性相关性，值在 $-1$ 到 $1$ 之间，如果模型得到的打分与人工标注的打分一致，得分就越高。具体而言，$X$ 和 $Y$ 之间的皮尔逊相关系数定义为 $X$ 和 $Y$ 之间的协方差与它们标准差的商：

$$\rho_{X,Y} = \frac{\text{cov}(X,Y)}{\sigma_X \sigma_Y} \tag{3.1}$$

**同义词检测 (tfl)**

托福考试（TOEFL）数据集 [55] 包含 80 个单选题，每个题目包含一个问题词以及四个选项，要求从四个选项中选出一个与问题词同义的词语。例如：问题"levied"，选项"imposed"、"believed"、"requested"、"correlated"，正确答案为"imposed"。对于每一个问题，需要计算问题词与选项词对应词向量之间的余弦距离，并选用距离最近的选项词，作为答案。在评价词向量时，本文直接使用 80 个问题的准确率。

**单词类比 (sem、syn)**

英文单词类比数据集由 Mikolov 等人于 2013 年 [73] 提出，该数据集包含了 9000 个语义类比问题以及 1 万个句法类比问题。语义类比问题包括国家首都、家庭成员称谓、国家货币等五类问题，如，"'king' 对 'queen' 如同 'man' 对什么？"，答案为"woman"。句法类比问题有比较级、最高级、名词单复数等九类问题，如 "'dance' 对 'dancing' 如同 'predict' 对什么？"，答案为"predicting"。

为了回答这类类比问题，Mikolov 等人 [73] 根据相似关系词对的词向量之差也相似的特点，提出使用词向量的加减法来完成这一任务。例如，对于问题

"'king' 对 'queen' 如同 'man' 对什么?",该方法直接从词表中寻找与 $\overrightarrow{queen} - \overrightarrow{king} + \overrightarrow{man}$ 最相似的词,作为答案。评价时使用回答问题的准确率。

单词类比任务的数据集相对前两个任务规模较大,因此在实验中,结果较为稳定,该指标也成为评价词向量的经典指标。

### 3.2.2　词向量用作特征

词向量可以从无标注文本中学习到句法和词法的特征,很多现有工作直接使用词向量作为机器学习系统的特征,并以此提高系统的性能。

本文选用两个有代表性的任务,一、将词向量作为唯一特征,完成文本分类任务;二、将词向量作为现有系统的额外特征,完成命名实体识别任务。选用词向量作为唯一特征可以从一个侧面看出词向量的表达能力,而选用词向量作为现有系统的额外特征可以看出词向量所含的信息与现有人工设计特征的区别。

#### 基于平均词向量的文本分类 (avg)

该任务直接以文本中各词词向量的加权平均值作为文档的表示,以此为特征,利用 Logistic 回归完成文本分类任务。其中权重为文档中各词的词频。本文选用了 IMDB 数据集 [71] 做文本分类实验。该数据集包含三部分,其中训练集和测试集各 2.5 万篇文档,用来做文本分类的训练和测试;无标注部分共 5 万篇文档,用于训练词向量。任务的评价指标为文本分类的准确率。

#### 命名实体识别 (ner)

命名实体识别 (Named entity recognition,NER) 在机器学习框架下,通常作为一个序列标注问题处理。在这一评价指标中,本文将词向量作为现有命名实体识别系统 [97] 的额外特征,该系统的性能接近现有系统的最好性能。实验设置与 Turian 等人实现的方式一致 [116]。任务的评价指标为命名实体识别的 F1 值,测试集是 CoNLL03 多任务数据集的测试集。

### 3.2.3　词向量用做神经网络初始值

Erhan 等人在文献 [27] 中论述了,恰当地选取神经网络的初始值,可以让神经网络收敛到更好的局部最优解。在自然语言处理任务中,基于神经网络模型的方法一般都会使用词向量作为其输入层的初始值。

　　在上一类词向量的用法（将词向量作为特征）中，词向量是模型的固定输入值，在模型的训练过程中，输入值不会改变，只有模型中的参数会改变。然而，将神经网络的初始值赋值为词向量之后，神经网络在训练过程中会改变设置的初始值。因此这两类词向量的用法表面上看非常相似，实质上却是不同的。

**基于卷积神经网络的文本分类 (cnn)**

　　卷积神经网络（Convolutional neural networks，CNN）是表示文本的有效模型。2014 年，Lebret 等人 [59] 以及 Kim 等人 [48] 同时提出用于文本分类任务的卷积神经网络。在这一评价指标中，本文选用了这一经典的卷积神经网络。网络结构在 5.2.4 节有详细介绍。

　　本文选取斯坦福情感树库（Stanford Sentiment Treebank）数据集作为文本分类的训练集、验证集和测试集 [110]。由于该数据集规模较小，文本分类的效果受网络初始值的影响较大，导致了评价指标的不稳定。为了更客观地评价卷积网络中，不同词向量对文本分类性能的影响，本文对每一份词向量重复做 5 次实验。在每次实验中，输入层词表示均初始化为这份词向量，网络结构中的其它参数则初始化为不同的随机值。对于每一次实验，本文在训练集上训练卷积神经网络，取验证集上准确率最高的点，并报告其在测试集上的准确率。最后将 5 组实验的测试集准确率的平均值作为最终的评价指标。

**词性标注 (pos)**

　　词性标注（part-of-speech tagging）是一个经典的序列标注问题。在这个任务中，本文使用 Collobert 等人提出的网络 [18]，对句子中的每个词做序列标注。该任务选用华尔街日报数据集 [115]。评价指标为模型在验证集上达到最佳效果时，测试集上的准确率。

## 3.3 实验及分析

　　在这一节中，本文针对不同模型、不同语料、不同任务做了大量实验，并回答前面提出的四个问题。表 3-2 列举了主要实验设置，包括实验中选取的模型、语料，以及实验中使用的主要参数。在具体的实验中，本文并没有穷举所有的设置组合，而是根据实验需要选取其中的一部分参数进行评测。具体设置将在各实验中介绍。

| 类别 | 实验设置 |
| --- | --- |
| 模型 | Skip-gram, CBOW, Order, LBL, NNLM, C&W |
| 语料 | 维基百科（Wiki）：100M, 1.6B<br>纽约时报（NYT）：100M, 1.2B<br>Wiki+NYT（W&N）：10M, 100M, 1B, 2.8B<br>IMDB 电影评论：13M |
| 参数 | 维度：10, 20, 50, 100, 200<br>固定窗口大小：5 |

表 3-2　词向量对比实验设置总表

### 3.3.1　性能增益率

在使用 3.2 节中提到的 8 个不同的指标评价词向量性能时，可能会遇到两个问题。为了更直观地感受这两个问题，可以参考表 3-3a（第 37 页）中的实验结果。

一、不同评价指标的绝对数值差异较大。如 syn、sem 任务的性能一般在 40% 左右，ws 任务的性能集中在 50% 到 60%，而 tfl 的性能大约在 70% 多，pos 的性能都在 95% 以上。由于指标之间的差异，我们只能在同一个指标内对不同的模型进行纵向的比较，而较难对一个模型在不同指标中的表现做横向的比较。

二、不同评价指标内的相对差异变化较大。如 ws 任务的性能中，最好的模型达到了 63.89%，然而最差的模型只有 46.17%，差距有约 18 个百分点；但是在 avg 任务中各模型的性能差异相对较小，最差的模型能到 73.26%，而最好的模型也只能到 74.94%，差距不到 2 个百分点。

正是因为有这两个问题的存在，在评价词向量时，如果两个模型在性能数值上非常接近，我们会很难定量地判断孰优孰劣。例如，在 pos 中，Order 模型与 LBL 模型分别可以达到 96.76% 和 97.77% 的准确率。这里很难说 LBL 模型比 Order 模型更适合做 pos 任务。这种性能上微小的差别可能并不是由于模型的优劣产生的，更可能是由于测试集样本数较少（ws 任务和 tfl 任务）或者二次训练带来的误差（avg、ner、cnn 和 pos 任务）所导致的。

为了解决第一个问题，本文考虑用"性能增益"代替各项任务性能的绝对数值。性能增益是指一个词向量在某任务上的性能比随机词向量在该任务上性能的相对增幅。随机词向量与其他模型生成的词向量一致，也使用 50 维的实数

向量，但是其中各个维度的值，由 $-1$ 到 $1$ 之间的均匀分布随机生成。使用随机词向量在各任务中得到的性能数值表示，哪怕词向量中不包含任何有用的信息，该任务也能达到的性能。因此，某个词向量相对随机词向量的性能增益表示该词向量所包含的信息对这一任务带来的贡献。

单纯使用性能增益并不能解决上面的第二个问题。为了解决第二个问题，本文进一步提出使用"性能增益率"（Performance Gain Ratio）这一评价指标来代替性能增益。性能增益率的思想借鉴了文献 [24]，每个词向量只与同等条件下最好的词向量做对比。本文根据词向量的特殊性质，将词向量 $a$ 相对词向量 $b$ 的性能增益率定义为：

$$PGR(a, b) = \frac{p_a - p_{rand}}{p_b - p_{rand}} \qquad (3.2)$$

词向量 $a$ 对同等条件下最好的词向量 $best$ 的性能增益率 $PGR(a, best)$ 可以简写作词向量 $a$ 的性能增益率 $PGR(a)$：

$$PGR(a) = \frac{p_a - p_{rand}}{p_{best} - p_{rand}} \qquad (3.3)$$

上述二式中 $p_x$ 表示词向量 $x$ 在某项任务上的性能，$p_{rand}$ 表示随机向量在这项任务上的性能。性能增益 $p_a - p_{rand}$ 体现的是词向量 $a$ 相比随机词向量可以带来的性能的提升。类似地，性能增益率体现的是，词向量 $a$ 与词向量 $best$ 相比，所带来性能提升的比例。

由于本文设定词向量的参考标准为同等条件下效果最好的词向量，因此，性能增益率是一个小于等于 $1$ 的数字。如果值为 $1$，则表示该词向量已经达到同等条件下的最佳效果；如果为 $0$，则说明与随机词向量的效果一样，没有带来任何收益；如果是负数，说明该词向量不仅没有对任务带来提升，而且还对任务产生了负面的效果。

本文需要在 $8$ 个不同的指标下，对各词向量进行评测。使用性能增益率可以让所有的评测结果都有一套统一的指标描述性能，并且一般情况下性能的数值大小均在 $0$ 到 $1$ 之间。这一特点一方面可以让我们以最直观的方式了解到每份词向量对任务带来的实际提升有多少；另一方面，这也为多指标联合分析带来了可能，使得我们可以直接分析各个词向量在多个任务下的综合表现。

### 3.3.2　模型比较

为了公平地比较各个不同的模型，本文需要对各模型采取相同的实现，同时使用同样的语料训练。

**实验概览**

在模型实现方面，本文使用的 Skip-gram 模型和 CBOW 模型的实现基于 word2vec 开源工具包[2]。其余模型均在 word2vec 工具包中 CBOW 实现的基础上修改得到。具体来说，在 CBOW 模型中，上下文的表示为上下文若干词的平均词向量；在 Order 模型中，本文将其替换为上下文若干词的词向量的拼接；在 LBL 模型的实现中，本文在 Order 模型实现的基础上加入了一个隐藏层，使得上下文表示先通过一次线性变换进入隐藏层，再对其预测；在 NNLM 模型的实现中，本文在 LBL 模型的线性变换之后加入 tanh 激活函数，使其成为一个非线性变换，整个模型是一个三层的神经网络结构；在 C&W 模型的实现中，本文基于 NNLM 的实现，将预测的目标词从输出层移动到输入层，这样输入层就是目标词与其上下文词的词向量的拼接，而输出层则只保留一个节点，用于表示这组上下文、目标词组合的评分。

对于所有的模型，本文将中间词作为目标词（一般来说，以语言模型为基础的模型，如 NNLM 和 LBL，都使用最后一个词作为目标词，这里也将其改成中间词），目标词上下文各两个词作为其对应的上下文，即 $win = 5$。对于所有模型，本文使用二次采样（subsampling）技术[75]，并设置 $t = 10^{-4}$。二次采样技术的细节可见第二章第 2.2.6 小节。word2vec 工具包与文献[75]中所描述的二次采样公式略有差别，本文在实验中使用 word2vec 工具包所用的公式。同时，为了提高实验效率，本文也对 word2vec 工具包做了适当的改动。word2vec 工具包采用梯度下降法作为其优化算法，同时也采用了学习速率下降的策略。在训练的初始阶段，其学习速率为一个设定的初始学习速率（如 CBOW 模型默认为 0.05）；在训练过程中，学习速率均匀下降，下降幅度与已学习的样本个数呈正比；到训练的最后阶段，学习速率降为 0。在这种学习速率下降策略下，如果想分析迭代 1 次到 $n$ 次对结果的影响，由于迭代的中间结果不能直接使用，程序需要扫描语料 $1 + 2 + \cdots + n = n(n+1)/2$ 次才能真正生成各种迭代次数的训练结果。为了方便分析迭代次数带来的影响，本文使用 AdaGrad[25] 替代原先的

---

方法，更改后，与普通的梯度下降算法一致，只需要扫描 $n$ 次语料，即可完成对迭代次数影响的分析。本文设定 AdaGrad 的学习速率为 0.1，使用该学习速率时，与原方法的学习效果最为类似。实验表明，将原优化方法改成 AdaGrad 之后，效果基本保持不变，有时会有微弱的提升。

在语料方面，本小节使用统一的训练语料：W&N 数据集（维基百科与纽约时报的混合语料，共 28 亿单词，详见 3.3.3 小节）。同时，由于这些模型都采用迭代算法优化，为保证每个模型都能训练到最佳状态，在每一组实验中，本文对每个模型反复迭代，直到该模型在所有的任务上性能均已经收敛或达到过峰值，并取迭代过程中的最佳值作为模型在该任务上的性能。因此，对于各项任务，可能会选取不同的迭代次数的词向量。

表 3-3 展示了上述实验结果。其中表 3-3a 展示了各任务中性能的绝对数值，表 3-3b 则是转换后的各模型在各任务中的性能增益率。表格中模型为"随机"一行表示，如果采用随机的词向量，各项任务可以达到的性能。从实验结果中可以看出：一、与随机词向量相比，无论哪个模型，在哪个任务上，效果均有显著的提升。也就是说，这些词向量模型均可以在一定程度上捕获句法和语义的特征；并且可以用来提升自然语言处理任务的效果。二、从总体上看，各个任务的最佳结果（加粗部分）由不同的模型得到。从 *Skip-gram* 到 *LBL*，均有取得若干最佳结果。

## 上下文的表示

分析上下文的表示时，我们主要考虑除了 C&W 以外的模型，这些模型的唯一差异就是上下文的表示方式不同。

为了说明不同上下文表示之间的差别，本文在不同规模的语料下对这些模型进行比较。具体而言，本文加入了 W&N 数据集的三个子集，分别包含 1000万（10M）单词、1 亿（100M）单词和 10 亿（1B）单词。这三个子集与完整集合 28 亿（2.8B）单词一起，用于分析不同种类上下文表示的影响。

在这个实验中，需要从模型、语料规模、任务这三个维度中发现规律，为了把注意力放在模型和语料规模上，本文利用性能增益率将各个任务的性能归一化。如果一个模型在某任务上达到 95% 的性能增益率，本文认为这个模型在该任务上足够好。选取 95% 可以在一定程度上消除小规模评价数据集以及二次训练带来的误差。比如，在表 3-3b 中，NNLM 模型在 cnn 任务和 pos 任务上均

| 模型 | syn | sem | ws | tfl | avg | ner | cnn | pos |
|------|-----|-----|-----|-----|-----|-----|-----|-----|
| 随机 | 0.00 | 0.00 | 0.00 | 25.00 | 64.38 | 84.39 | 36.60 | 95.41 |
| Skip-gram | 51.78 | **44.80** | **63.89** | 76.25 | **74.94** | **88.90** | 43.84 | 96.57 |
| CBOW | **55.83** | 44.43 | 62.21 | **77.50** | 74.68 | 88.47 | 43.75 | 96.63 |
| Order | 55.57 | 36.38 | 62.44 | **77.50** | **74.93** | 88.41 | **44.77** | **96.76** |
| LBL | 45.74 | 29.12 | 57.86 | 75.00 | 74.32 | 88.69 | 43.98 | **96.77** |
| NNLM | 41.41 | 23.51 | 59.25 | 71.25 | 73.70 | 88.36 | 44.40 | 96.73 |
| C&W | 3.13 | 2.20 | 46.17 | 47.50 | 73.26 | 88.15 | 41.86 | 96.66 |

(a) 性能实际值

| 模型 | syn | sem | ws | tfl | avg | ner | cnn | pos |
|------|-----|-----|-----|-----|-----|-----|-----|-----|
| Skip-gram | 93 | **100** | **100** | 98 | **100** | **100** | 89 | 85 |
| CBOW | **100** | 99 | 97 | **100** | 98 | 90 | 88 | 90 |
| Order | **100** | 81 | 98 | **100** | **100** | 89 | **100** | 99 |
| LBL | 82 | 65 | 91 | 95 | 94 | 95 | 90 | **100** |
| NNLM | 74 | 52 | 93 | 88 | 88 | 88 | 95 | 97 |
| C&W | 6 | 5 | 72 | 43 | 84 | 83 | 64 | 92 |

(b) 性能增益率

表 3-3　各模型在完整 W&N 语料下的最佳性能（百分比）

达到或超过了 95% 的性能增益率，因此 NNLM 在使用完整 W&N 语料训练时，在两个任务上性能足够好。在评估性能增益率时，各模型只与同样配置的模型做比较，用 10M 语料训练得到的词向量，只与 10M 语料训练得到的最好词向量进行比较，其它规模的语料也类似。

根据第二章 2.2.8 小节中的分析，本文从模型复杂度的角度分析了不同的上下文表示。基于这种视角，可以更好地解释实验结果。表 3-4 展示了在四种不同规模的语料下训练，各模型在多少任务中表现足够好。每个单元格中 *a*+*b* 表示某模型（行）使用某语料（列）训练得到的词向量，在前四个评价语言学特性的任务中，有 *a* 个任务表现足够好；在后四个评价其用于特征或用于神经网络初始值效果的任务中，有 *b* 个任务表现足够好。根据表格中的数据，本文做了以下分析。

一、简单模型在小语料上整体表现更好，而复杂的模型需要更大的语料作支撑。首先从语料大小的角度观察，在 10M 规模的训练语料下，结构最简单的

| 模型 | 10M | 100M | 1B | 2.8B |
|------|-----|------|-----|------|
| Skip-gram | 4+2 | 4+2 | 2+2 | 3+2 |
| CBOW | 1+1 | 3+3 | 4+1 | 4+1 |
| Order | 0+2 | 1+2 | 2+3 | 3+3 |
| LBL | 0+2 | 0+2 | 0+2 | 1+2 |
| NNLM | 0+2 | 0+3 | 0+3 | 0+2 |

表 3-4　各模型在不同规模语料下性能增益率超过 95% 的次数

Skip-gram 模型有 6 个任务足够好，是最佳的选择；当语料扩大到 100M 时，稍微复杂一些的 CBOW 模型开始体现出优势；如果继续扩大语料，能保留词序信息的 Order 模型效果超过了 Skip-gram 和 CBOW 模型。然后观察结构最复杂的 LBL 模型与 NNLM 模型，从中可以大致看出，随着语料规模逐渐变大，这两个模型的相对效果在逐渐变好。实验中，即使在最大规模的语料下，这两个模型也没有达到所有模型中的最好效果，但是从趋势看，这两个保留语义组合关系的模型，仍然有潜力在更大规模的语料下超越其它更简单的模型。这一部分的实验只是探讨了语料规模与模型复杂度之间的关系，实际上从语料的角度来看，语料领域的影响比语料规模或者模型的影响更大，后文 3.3.3 小节会有具体探讨。

二、对于实际的自然语言处理任务，各模型的差异不大。各模型在实际的自然语言处理任务中（加号后面的数字）表现均比较类似。结合表 3-3 中的结果，在完整 W&N 语料下训练时，简单模型和复杂模型的性能差距也非常的小。因此，一般情况下，将 Skip-gram 和 CBOW 等简单模型生成的词向量，用于自然语言处理任务，就可以得到一个令人满意的效果。

## 上下文和目标词的关系

现有词向量模型中，目标词与上下文之间主要有两种关系。C&W 模型对目标词与其上下文联合打分；而其他模型均为根据上下文，预测目标词。因此，在分析上下文和目标词之间的关系时，我们主要关注 C&W 模型与其它模型的差异。实验结果显示，C&W 模型在语言学任务上（syn、sem、ws 和 tfl），效果均不如其它模型。尤其在类比任务 syn 和 sem 中，C&W 模型的性能接近于随机词向量，可以看出 C&W 的词向量几乎不存在线性减法关系。与此同时，C&W 模型在其它任务上表现并不差。为了深入分析 C&W 模型与其它模型的区别，本

文选取了若干词，并列举了用 CBOW 和 C&W 模型得到的词向量中，与这些词最相似的词（词向量空间距离最近的词，后文简称"最近邻"）。表 3-5 展示了对比结果。

| 模型 | Monday | commonly | reddish |
|------|--------|----------|---------|
| CBOW | Thursday<br>Friday<br>Wednesday<br>Tuesday<br>Saturday | generically<br>colloquially<br>popularly<br>variously<br>Commonly | greenish<br>reddish-brown<br>yellowish<br>purplish<br>brownish |
| C&W | 8:30<br>12:50<br>1PM<br>4:15<br>mid-afternoon | often<br>generally<br>previously<br>have<br>are | purplish<br>pendulous<br>brownish<br>orange-brown<br>grayish |

表 3-5　若干词及其用不同模型得到的最近邻对比表

从表中可以发现，用 CBOW 模型训练时，"Monday"的最近邻是一星期中的其它几天，而 C&W 模型得到的的却是一天中的时刻。类似地，"commonly"的最近邻用 CBOW 得到的都是一些语义类似的词，使用时可以替换"commonly"，而 C&W 模型得到的有些是和 commonly 搭配使用的词。"reddish"的最近邻多为颜色，除了 C&W 模型得到的"pendulous"，该词一般与"reddish"一起使用，形容花朵。

根据文献 [98] 中的论述，CBOW 这类通过上下文预测目标词的模型直接对二阶关系进行建模，而一般认为二阶关系就是替换关系（paradigmatic）[57]。简单地说，这些模型通过相似的目标词，构建出上下文之间的相似关系。然而，C&W 模型没有显式地对二阶关系进行建模。实验中也可以看出，C&W 的最近邻并不全是替换关系的词，而 CBOW 模型得到的几乎全是替换关系的词。本文的结果初看与 C&W 模型原始论文 [18] 的结果有差异，在该论文的表 7 中，C&W 模型的最近邻均为替换关系。实际上，文献 [18] 的结果并不是纯 C&W 得到的结果，而是在 C&W 模型之后，进一步使用了四个有监督任务（词性标注、命名实体识别等）调整参数得到的词向量。有监督任务与 CBOW 这类预测目标词的方法类似，也会使相似的词向量具有替换关系。因此，本文认为，C&W 之所以与其它模型的效果有较大的差异，主要是因为 C&W 模型将目标词放在了

输入层，这种结构对语义关系（主要是替换关系）的建模能力不如 CBOW 这类预测目标词的结构。

## 结论

经过两个实验的比较，现在可以回答引言中提出的第一个问题：**使用哪个模型效果更好？具体而言，在多种不同的上下文表示之中，以及在上下文与目标词的两种不同的关系之间，应当如何选择合适的模型？**

一、对于小语料，像 *Skip-gram* 这样的简单的模型会取得更好的效果。对于更大的语料，*CBOW* 和 *Order* 这样对上下文更复杂建模的模型，会有更好的效果。

二、对于实际的自然语言处理任务（比如将词向量用作现有任务的特征，或者用于神经网络模型的初始值），使用 *Skip-gram*、*CBOW* 和 *Order* 这样的简单模型就已经足够好。

三、对于评价语言学特性的任务（比如词汇相似度），通过上下文预测目标词的模型，比上下文与目标词联合打分的 *C&W* 模型效果更好。其中比较特别的是，*C&W* 模型得到的词向量完全不包含线性平移关系。

### 3.3.3  语料影响

本小节从语料规模和语料的领域两个角度着手，分析语料对词向量的影响。

## 实验概览

为了充分研究语料领域的影响，本文选取了三份不同领域的语料，其中包括 16 亿词的维基百科语料（Wikipedia）[3]、12 亿词的纽约时报语料（New York Times）[4]以及 1300 万词的 IMDB 电影评论语料。同时，为了探索混合领域语料的效果，本文还将维基百科与纽约时报语料合并，称作 W&N 语料，该语料总共包含 28 亿词，是本次实验中最大的语料。表 3-6 展示了上述各语料的概要信息。总体而言，纽约时报语料的文档长度较长，而维基百科语料由于涉及的知识面更为广泛，其唯一词数也较多。

---

[3]本文使用了 2013 年 9 月 7 日的维基百科全站存档，使用更新的存档效果应当会更好。https://dumps.wikimedia.org/enwiki/

[4]包含 1987 年 1 月到 2007 年 7 月的纽约时报新闻。https://catalog.ldc.upenn.edu/LDC2008T19

| 语料 | 词数 | 文档数 | 唯一词数 | 词表内词数 |
|------|------|--------|----------|------------|
| 维基百科（Wiki） | 1,705,736,997 | 4,335,623 | 8,387,089 | 1,643,119,281 |
| 纽约时报（NYT） | 1,207,480,927 | 1,855,658 | 3,321,810 | 1,190,382,040 |
| IMDB 电影评论 | 13,419,330 | 50,000 | 85,092 | 13,243,538 |

表 3-6 训练词向量的各语料集概要信息

为了研究语料规模的影响，本文对三个较大的语料（维基百科、纽约时报、W&N 混合语料）选取其不同规模的子集进行分析。对于 W&N 语料，本文选取了 10 亿（1B）词、1 亿（100M）词、1000 万（10M）词三个子集。对于另外两个语料，本文各选取了 1000 万（10M）词的子集。表 3-2（第 33 页）中列举了各语料及其子集的设置。对于同一个语料而言，小子集是大子集的子集。也就是说，W&N 语料中，1000 万词子集的内容在 1 亿词的子集中均存在，1 亿词的子集的内容也可以在 10 亿词的子集中找到。另外两个语料也类似。

为了尽可能客观地分析真正由语料带来的影响，本文还对各语料做了以下两个约束。

一、为了消除词表选择对结果带来的影响，本文对所有实验设置了固定的词表。词表中的词保证在维基百科语料与纽约时报语料中至少各出现 23 次，使得最终词表大小约为 20 万词（201369）。对于任何一份语料，词表外的词会被直接忽略。表 3-6 中"词表内数"一列即为忽略词表外的词之后，语料中剩余词的数量。

二、打乱语料的顺序。word2vec 这类工具包采用从前往后的顺序依次学习语料中的各篇文档，这种方式训练得到的词向量会偏向语料中靠后部分的语义。如果语料中各篇文档的顺序以默认顺序排列，如纽约时报以时间顺序排列，就可能导致某些词的词向量所包含的语义更接近现代的语义，而不是几十年前的语义。这可能会带来一些潜在的不公平。为了尽可能降低这种情况的发生，本文对所有的语料进行文档级的随机打乱。只保证一篇文档中的词按照原来的顺序书写，而文档和文档之间的顺序是随机的。

使用不同语料训练得到的词向量，在各任务上的效果如表 3-7。根据 3.3.2 小节中的结论，在本小节所用的语料规模下，CBOW 模型的平均表现最好。因此本实验选取 CBOW 模型训练所有词向量，使用其它模型也能得到一致的结论。表中各任务的性能使用性能增益率衡量，其参考值为不同语料下能达到的最高

| 语料 | 规模 | syn | sem | ws | tfl | avg | ner | cnn | pos |
|------|------|-----|-----|-----|-----|-----|-----|-----|-----|
| 纽约时报 | 1.2B | 93 | 52 | 90 | 98 | 50 | 76 | 85 | 96 |
| | 100M | 76 | 30 | 88 | 93 | 46 | 77 | 83 | 86 |
| 维基百科 | 1.6B | 92 | **100** | **100** | 93 | 51 | **100** | 86 | 94 |
| | 100M | 74 | 65 | 98 | 93 | 47 | 88 | 90 | 83 |
| W&N | 2.8B | **100** | 89 | 95 | 93 | 50 | 97 | 91 | **100** |
| | 1B | 98 | 87 | 95 | **100** | 48 | 98 | 90 | 98 |
| | 100M | 79 | 63 | 97 | 96 | 51 | 85 | 92 | 86 |
| | 10M | 29 | 27 | 76 | 60 | 42 | 49 | 77 | 42 |
| IMDB | 13M | 32 | 21 | 55 | 82 | **100** | 26 | **100** | -13 |

表 3-7    使用不同语料训练 CBOW 模型时各任务的性能

值。因此每个评价指标的最大性能增益率就是 100%，对应表格中的每一列的最大值都是 100。以下分别分析语料的规模和领域对各任务的影响。

**语料规模**

　　同领域的语料，语料越大效果越好。对比纽约时报数据集与其 1 亿（100M）词的子集，或者对比维基百科语料与其子集，又或者 W&N 语料与其不同规模的子集，可以发现在几乎所有情况下，同领域的大语料比小语料效果要好。表格中有少量例外，可以看出即使在例外的情况下，小语料与其对应的大语料相比，效果的优势也非常微弱。这些例外很可能是因为训练词向量时的波动或者评价指标的不稳定导致的。

　　值得注意的是，在 *syn*（句法问题的单词类比任务）中，语料的规模直接决定了最终的性能，而语料领域几乎对结果没有影响。例如，三个语料的 100M 词子集效果非常类似；纽约时报、维基百科的完整语料以及 W&N 混合语料的 1B 词子集也有类似的性能；W&N 语料的 10M 词子集与 13M 词的 IMDB 语料也有类似的性能。syn 任务中的句法问题主要为类似 "year:years law:＿"、"good:better rough:＿" 的词形变换问题。尽管不同的语料所涉及的领域，甚至使用的体裁差异较大，但它们均遵循基本的英语语法。因此，不同领域的语料训练得到的词向量拥有相似的句法特征，在评价句法特性的任务上，表现也非常接近。

**语料领域**

对于大多数任务（除了 syn 以外的任务），语料领域对词向量在各任务中的表现起了主导作用。对于不同的任务，语料的领域有不同的影响。

一、在评价词向量语义特征的任务中，使用维基百科会比纽约时报有更好的效果，如 sem（语义问题的类比任务）和 ws（语义相关性任务）。在这两个任务中，维基百科语料的 100M 子集，性能也已经超过了 1.2B 的纽约时报语料。本文认为维基百科语料包含了更全面的知识，语义信息更为丰富，因此用维基百科语料训练得到的词向量，包含的语义特征也更为丰富，在评价语义特性的任务中，效果也更好。

二、领域内的语料对相似领域任务的效果提升非常明显，但在领域不契合时甚至会有负面作用。使用 IMDB 语料训练的词向量，在 avg 和 cnn 任务上表现优于其它语料，但是在 ner 和 pos 任务上表现很差。IMDB 语料包含了 IMDB 网站上的 50 万条电影评论，这与 avg 和 cnn 两个任务所用的训练集和测试集的数据源相同，均来自 IMDB 电影评论网站。在这两个任务中，IMDB 语料的效果甚至超过规模大两个数量级的领域外语料。尤其是在 avg 任务中，词向量作为文本分类的唯一特征，IMDB 训练的词向量，性能增益率大约是其它语料的两倍。但是另一方面，在 ner 和 pos 任务中，IMDB 训练的词向量效果非常差，在 pos 任务上甚至对最终性能有负面影响。本文认为 IMDB 语料在 ner 和 pos 任务中效果差主要是因为该语料为网络评论语料，与维基百科和纽约时报相比，书写相对不规范，可能只包含少量对命名实体识别（ner）和词性标注（pos）有用的信息。

从上述实验中可以看出语料的领域对任务性能的影响非常明显，那么为什么领域内的语料会对性能有这么大的提升呢？为了更直观地感受这一原因，本文列举了若干词及其在不同语料下的最近邻（词向量最相似的词），结果见表 3-8。"movie" 一词的最近邻在 IMDB 语料下主要是 "this"、"it"、"thing" 这类词，也就是说，"movie" 一词在 IMDB 语料中甚至可以看做是停用词。"Sci-Fi" 在 IMDB 语料下的最近邻均为 "科幻电影" 其他形式的缩写，而在 W&N 语料下，则是其他类型的电影。"season" 在 IMDB 下的含义表示电视剧的 "季"，而在 W&N 下主要表示比赛的赛季。从这些例子中可以看出，用领域内语料训练的词向量之所以能提升对应领域的任务，其中一个主要原因是通过领域内语料训练的词向量，包含的语意与词在该领域内的含义更为一致。

| Corpus | movie | Sci-Fi | season |
|--------|-------|--------|--------|
| IMDB | film | SciFi | episode |
| | this | sci-fi | seasons |
| | it | fi | installment |
| | thing | Sci | episodes |
| | miniseries | SF | series |
| W&N | film | Nickelodeon | half-season |
| | big-budget | Cartoon | seasons |
| | movies | PBS | homestand |
| | live-action | SciFi | playoffs |
| | low-budget | TV | game |

表 3-8　若干词在 IMDB 和 W&N 语料下的最近邻

　　神经网络词向量模型的训练均基于分布假说，因此，词的语义直接由其上下文决定。在不同领域的语料中，针对同一个词的上下文分布差异可能会非常大，特别是多义词。领域内的语料往往只刻画出某个词在领域内的含义，而这对于同领域的任务是恰好合适的。

**规模和领域的权衡**

| IMDB＼W&N | 20% | 40% | 60% | 80% | 100% |
|-----------|-----|-----|-----|-----|------|
| +0% | 91 | 94 | 100 | 100 | 100 |
| +20% | 79 | 87 | 91 | 96 | 99 |
| +40% | 68 | 86 | 88 | 92 | 98 |
| +60% | 65 | 79 | 85 | 88 | 93 |
| +80% | 64 | 75 | 84 | 87 | 92 |
| +100% | 64 | 70 | 83 | 86 | 88 |

表 3-9　混合语料训练的词向量，在 avg 任务上的效果

　　前面两个小节的实验表明，领域内的数据越多，对于同领域任务的效果也就更好。然而在实际情况下，对于一个特定领域的任务，领域内的语料往往数量有限。如果需要有更大规模的语料，则不得不引入领域外的语料。也就是说，语料的领域纯度与语料规模是互相冲突的。在这种情况下，应当尽可能保持语料的领域纯度，还是增加领域外的语料来扩大语料的规模？为了回答这一问题，

本文设计了如下实验：选取 13M 词的 IMDB 数据集以及 W&N 语料的 13M 词子集，通过这两个数据集的不同比例的融合，探索语料领域和规模哪个更重要。实验结果如表 3-9 所示，表格中的结果为混合语料用 CBOW 模型训练，在 avg 任务下的效果。表格中的每一列表示使用了多少比例的 IMDB 语料，每一行则表示加入了多少比例的 W&N 的 13M 子集。表格中的每一列从上到下，语料规模都在增大，而语料的领域纯度都在降低。实验结果表明无论选取多大规模的 IMDB 语料，增加 W&N 语料都会一致地降低 avg 任务的性能。这一实验中，语料的领域纯度比语料规模更重要。

除了这一实验以外，表 3-7 中的结果以及前面两个小节的分析也可以看出语料的领域更为重要。如 sem、ws、ner 任务中，100M 的维基百科语料效果超过了 1.2B 的纽约时报语料；而且在这三个任务中，维基百科和纽约时报的混合语料，均不如单纯使用维基百科语料。

**结论**

经过上述实验及分析，现在可以回答引言中的第二个问题**训练语料的大小及领域对词向量有什么样的影响？** 本文认为语料领域比语料规模更重要，具体而言，可以分为以下几点：

一、使用领域内的语料，对同领域的任务有明显的帮助。领域内的语料可以让词向量拥有领域内的语义，对同领域任务的促进是最明显的。而且，使用纯领域内语料比混合领域外的大规模语料的效果更好。

二、如果选择了不合适的语料，很可能没有办法从语料中得到与任务相匹配的语义或者用法，会对相应的任务起到负面效果。如使用网络评论语料训练的词向量，用于词性标注任务时，性能下降。

三、对于同领域的语料，语料规模越大，词向量性能越好。

### 3.3.4  参数选择

**迭代次数**

现有神经网络词向量模型均采用迭代方式训练。与其它迭代训练的机器学习方法类似，词向量模型的训练结果受迭代次数的影响较大。一般来说，如果迭代次数较少，会导致训练不充分；然而如果迭代次数过多，模型可能会过拟合。机器学习领域应用最广泛的迭代终止条件是，当验证集损失（损失函数的

值) 达到最低值时，迭代终止 [95]。在词向量训练过程中，损失函数刻画了模型预测"目标词"的精度 (或者上下文与目标词的匹配程度)。然而真实的任务并不要求词向量能够多么精确地预测目标词，而是希望词向量具有更好的语言学特性，以及能更好地辅助其它自然语言处理任务。因此，损失函数仅仅是实际任务的一个代理，在某些情况下，损失函数与实际任务的性能可能不一致。本小节从这一经典方法开始，探索适合词向量训练的迭代停止条件。

本小节的实验选取 95% 的语料作为训练集，剩下 5% 作为验证集。图 3-1 描绘了 3 个具有代表性的训练过程，分别为使用 CBOW 模型在 W&N 语料的三个子集 (10M 单词、100M 单词、1B 单词) 上的训练过程。横坐标表示迭代次数，纵坐标为各任务的性能以及验证集损失。

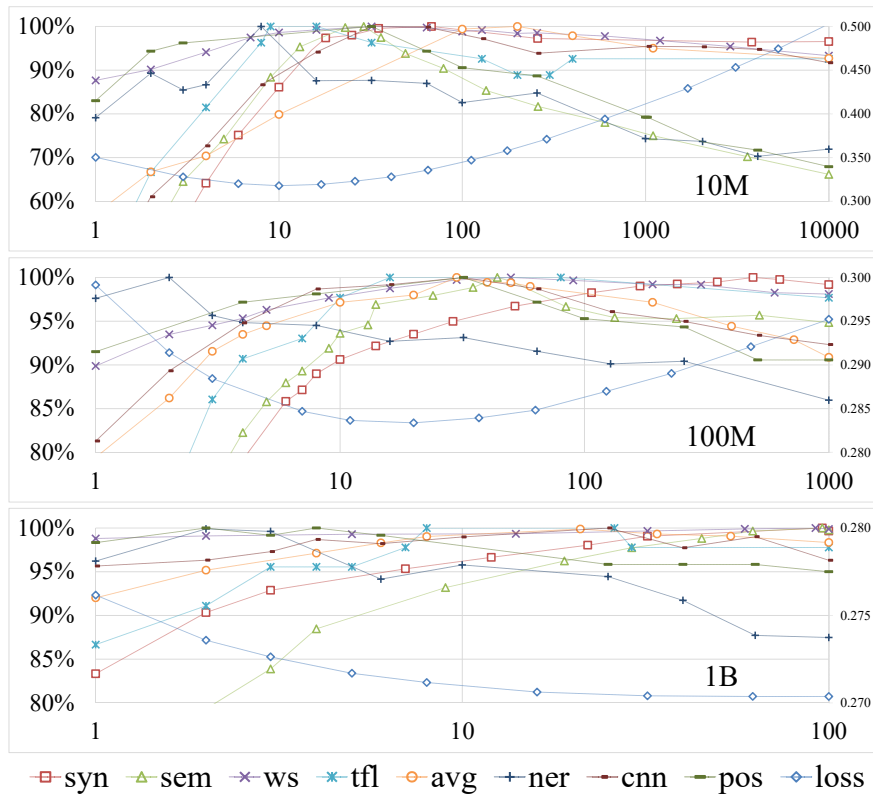

图 3-1    验证集损失及各指标的性能增益率随着迭代次数的变化曲线

从这些例子中可以发现，验证集损失的确与实际任务的性能不一致。在 100M 词的子集中，验证集损失在第 20 次迭代达到峰值，也就是说，在 20 次迭代之后，模型已经过拟合了训练数据，但是各任务的性能在之后的几次迭代中

仍然持续上升。与之相对的，在 1B 词的子集中，模型直到最后都没有过拟合训练集，但是 ner 与 pos 任务的性能在若干次迭代之后就开始下降。这些例子说明，在训练词向量时，使用验证集损失作为迭代停止条件，可能不是一个合适的选择。

值得注意的是，相比任务性能与验证集损失之间的差距，各个任务之间的差异反而更小。该结果表明，可以使用一个简单的任务去检测词向量是否在其它任务上迭代到最佳状态。为了验证这一方法的可行性，本文遍历了 6 种模型，在 3 个 W&N 语料的子集上训练，用 8 个任务检验了效果，总共 144 种搭配。表 3-10 报告了选取各任务以及验证集损失作为参考指标时，在多少种情况下词向量可以训练得足够好（达到迭代中峰值效果的 95% 以上）。如果使用经典的验证集损失作为指标，最后只有 89 组实验足够好，然而使用 tfl 任务（这些任务中最简单的任务）的峰值作为迭代停止条件时，总共有 117 组实验足够好。

| 验证集损失 | syn | sem | ws | tfl | avg | ner | cnn | pos |
|---|---|---|---|---|---|---|---|---|
| 89 | 105 | 111 | 103 | 117 | 104 | 91 | 103 | 101 |

表 3-10　不同迭代停止条件下，词向量足够好的实验数量（共 144 组）

当针对某个任务训练词向量时，使用任务对应的验证集作为迭代终止条件是最好的选择，因为这一目标与最终目标是一致的。然而在某些实际情况下，测试一遍任务的验证集可能会非常耗时，比如 cnn 和 pos 这类任务，评估性能需要数十分钟时间（文本分类或者词性标注的训练过程非常耗时），而 tfl 任务只需要几秒即可完成。因此，这一策略为性能的峰值提供了一个比较不错的近似，尤其在目标任务非常耗时的情况下较为有用。

另外值得一提的是，尽管众所周知，迭代算法一般需要多次迭代才能达到最佳效果，本小节的实验结果也支持这一结论。但是在 word2vec 工具包的早期版本中[5]，训练 Skip-gram 模型和 CBOW 模型均只使用一次迭代，而由于该工具包是目前最常用的训练词向量的方法，因此大量基于 word2vec 的工作并没能达到其最佳性能。特别是在一些比较工作中（如文献 [90]），在比较其它模型与 Skip-gram 和 CBOW 模型时，对 Skip-gram 模型和 CBOW 模型只迭代了一次，得到的结果并不准确。

---

[5] 该问题仅存在于最早的版本中，在 2014 年 9 月的更新中，word2vec 工具包已支持多次迭代。

现在可以回答引言中提出的第三个问题，**在迭代训练中，选择什么样的迭代次数可以获得足够好的词向量，同时避免过拟合？** 本文认为在大多数情况下，可以选取一个简单任务的性能峰值作为训练词向量的迭代终止条件。在条件允许的情况下，选择目标任务的验证集性能作为参考标准，是最合适的选择。

**词向量维度**

本小节通过选择不同的模型和任务，分析词向量维度对性能的影响。实验发现，对于分析词向量语言学特性的任务，有着一致的结论：维度越大，效果越好。图 3-2(a) 以 tfl 任务为例，绘制了性能随着维度的变化曲线图。这一结论在文献 [73] 的实验中证实，当词向量维度到达 600 维时，其语义特性仍然在变好。由于训练更高维度的词向量非常耗时，目前尚不能确定这一结果成立的上界。与分析词向量语言学特性的任务不同，在使用词向量提升自然语言处理任务的指标中，词向量维度到达 50 之后，效果提升非常微弱。图 3-2(b) 以 pos 任务为例，展示了这类实验结果。**词向量的维度选择多少维比较合适？** 本文认为，对于分析词向量语言学特性的任务，维度越大效果越好（除 C&W 模型以外，在 3.3.2 节中已有解释）；对于提升自然语言处理任务而言，*50 维词向量通常就足够好*。

## 3.4 相关工作

本节将列举词向量比较方面的相关工作，并分析这些工作得到的结论与本章实验结论的异同。

### 3.4.1 模型比较

在比较词向量的各项工作中，与本文最类似的工作是 Turian 等人在 2010 年发表的文献 [116]。他们在文中比较了 HLBL 模型与 C&W 模型在命名实体识别（ner）任务和短语识别（chunking）任务中的表现，这两个任务均在现有自然语言工具的基础上加入词向量作为额外的特征，以此评估词向量对系统性能的提升。他们的实验中，使用了一个 6300 万（63M）词的小规模语料，并发现 HLBL 模型与 C&W 模型在两个任务中表现相当。

Baroni 等人在 2014 年发表的文献 [3] 中比较了"计数"模型与"预测"模型在若干语义相似度任务中的表现。他们在文中将基于统计"词-上下文"共现

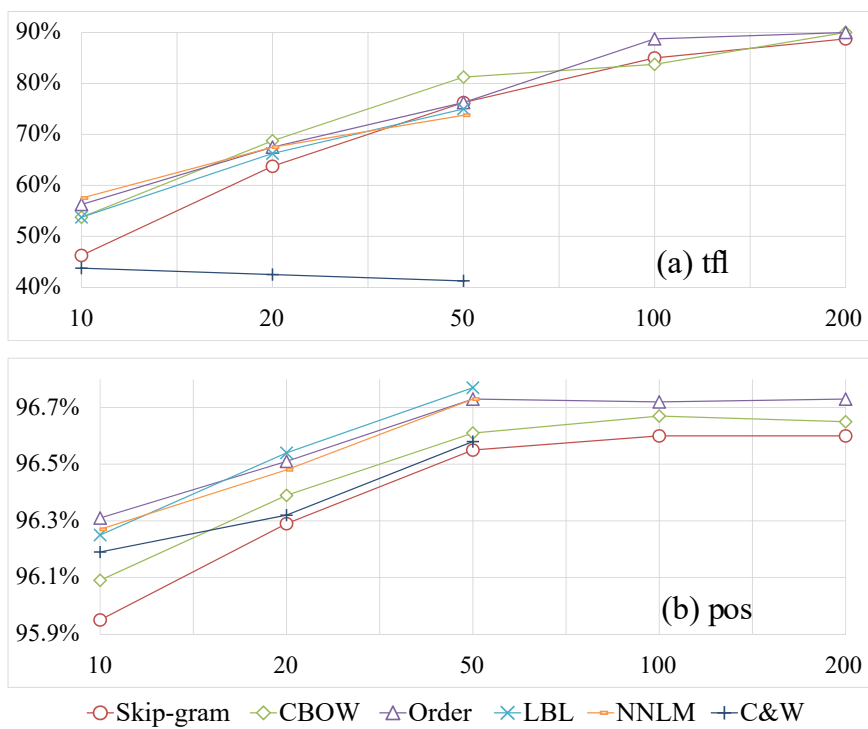

图 3-2 tfl 和 pos 任务的性能随词向量维度的变化曲线

矩阵,以及在其基础上进行矩阵分解的方法,统称为计数模型;并将基于神经网络的词向量模型统称为预测模型。Baroni 等人在实验中使用"词-上下文"共现矩阵的原始形式、SVD 分解、NMF 分解(非负矩阵分解)[62, 68] 作为计数模型的代表,使用 CBOW 模型作为预测模型的代表。他们对这两类模型进行多种语义特性的评价,包括语义相关性、同义词判别、概念分类、类比等。其实验表明,预测模型对在各项指标中比计数模型有显著的优势。

然而,Milajevs 等人在 2014 年发表的文献 [77] 中,给出了相反的结论。文中指出,他们在实验中,尝试了向量逐元素相加、逐元素相乘等基本的语义组合方式,使用这些组合方式表示短语以及句子的语义,并通过短语语义相似度等指标进行评测。实验结果表明,基于共现矩阵的词表示方法相比神经网络的词向量模型,其基本的语义组合能力更强。

Levy 等人在 2015 年发表的文献 [65] 中尝试了多种不同的模型参数,包括动态窗口或者固定窗口的选择、重新采样技术的效果、低频词的处理方式、上下文的平滑方案等。实验发现,大部分参数设置的技巧,对基于共现矩阵的模型以及基于神经网络的模型同时有效。

### 3.4.2　语料影响

语料规模方面，Mikolov 等人在文献 [73] 中发现，语料规模越大，CBOW 模型在类比任务（本文中的 syn 任务和 sem 任务）中效果更好。Pennington 在文献 [90] 中指出，对于 GloVe 而言，语料规模越大，句法问题的类比任务（syn）效果越好，但是语义问题的类比任务（sem）却不一定。

语料领域方面，Stenetorp 等人在 2012 年 [111] 发现，使用领域内的语料训练得到的词表示，相比使用新闻语料训练得到的词向量，在进行生物医药领域的命名实体识别任务时，有明显的优势。他们在实验中使用的词表示为布朗聚类（Brown Clustering），与基于神经网络的模型不同，布朗聚类得到的词表示并非是一个低维的实数向量。但是类似的是，无论是基于聚类的词表示方法还是基于神经网络的词向量表示方法，他们都是根据语料统计建模得到的，因此语料的影响也是类似的。

## 3.5　本章小结

词向量包含了丰富的词义信息，对词义分析以及各项自然语言处理任务均有一定的帮助。长期以来人们通过增加语料、改进模型等手段，试图寻找一种通用有效的词向量。然而本文通过大量实验，发现对于不同的任务，最好的词向量所用的模型、语料、参数均各不相同。也就是说，对于所有任务都有效的通用词向量，可能是不存在的。从模型的角度看，不同的模型建模了不同的语义关系，如 C&W 模型可以对组合关系进行建模，而 CBOW 等模型可以更好地对替换关系进行建模。从语料的角度看，特定领域的任务需要词在该领域中的含义，而往往只有领域内语料，才能将多义词训练成为领域内的语义。从任务的角度看，不同的任务需要词向量具有不同的性质，这些性质之间可能存在矛盾，如 ner、pos 任务希望词向量包含规范的语法信息，而 cnn 任务需要词向量具有网络语言的灵活表示。因此无论从模型、语料还是任务的角度看，通用有效的词向量，都是几乎不可能获得的。尽管不存在对所有任务有效的词向量，对于特定任务，依然可以使用本章得到的结论，生成一份有效的词向量。

1. 选择一个合适领域的语料，在此前提下，语料规模越大越好。使用大规模的语料进行训练，可以普遍提升词向量的性能，如果使用领域内的语料，对同领域的任务会有显著的提升。

2. 选择一个合适的模型。复杂的模型相比简单的模型，在较大的语料中才有优势。简单的模型在绝大多数情况下已经足够好。预测目标词的模型比目标词与上下文呈组合关系的模型（C&W 模型）在多个任务中有更好的性能。

3. 训练时，迭代优化的终止条件最好根据具体任务的验证集来判断，或者近似地选取其它类似的任务作为指标，但是不应该选用训练词向量时的损失函数。

4. 词向量的维度一般需要选择 50 维及以上，特别当衡量词向量的语言学特性时，词向量的维度越大，效果越好。

# 第四章  基于字词联合训练的中文表示及应用

不同于英文，中文中最自然的语言单位是"字"。现有工作在学习中文文本的向量表示时往往直接沿用了英文的处理方式，而忽略了中文的特殊性。本章根据中文特点，提出了基于字词联合训练的中文字、词表示学习方法。本文还提出了一个基于神经网络的中文分词模型，探索了神经网络在中文分词中的效果，也在该任务上评价字词联合训练得到的字表示效果。

## 4.1  引言

词是"最小的能独立运用的语言单位"[1]，由于中文具有大字符集连续书写的特点，如果不进行分析，计算机则无法得知中文词的确切边界，从而很难理解文本中所包含的语义信息。因此，中文分词是自然语言处理中的一个关键基础技术，是其他中文应用，例如命名实体识别、句法分析、语义分析等任务中前期处理的关键环节，其性能的优劣对于中文信息处理尤为重要。

中文分词的研究在近二十年来取得了丰富的成果。早期采用的是基于词典的匹配方法，如：最大正向匹配、最大逆向匹配、双向匹配等。然而，由于语言的复杂性，中文文本中存在大量的词边界歧义与未登录词 (OOV)。仅仅使用基于词典的匹配方法无法有效地解决以上两个中文分词中的关键难点问题。所以越来越多的方法关注基于字的中文分词。基于字的中文分词方法的基本假设是一个词语内部字符高内聚，而词语边界与外部汉字低耦合。每一个词都可以通过其所在的上下文特征进行表示，通过统计模型可以很好的判别当前字在构词过程中的作用 (词的开始、中间、结束或是单字词)。大量实验表明，这种基于字的中文分词方法要明显优于基于词典匹配的分词方法。然而，现有基于字标注的分词方法通常使用词 (unigram)、二元词组 (bigram) 等特征，这些特征都是相互孤立的，且本身并不具有语义信息。比如"江"、"河"、"湖"、"海"在表示地名时用法非常类似，但是现有模型会分别对这四个字进行建模，不能很好地利用字之间的联系。

近些年随着深度学习的兴起，基于神经网络的特征学习方法为自然语言处理带来了新的思路。如本文其它章节所介绍的，现有基于神经网络的词向量和文档向量表示技术已经为命名实体识别、词性标注、文本分类等多项任务带来

了性能上的提升。本文基于这一想法，提出了基于字表示的分词模型。该模型利用神经网络字表示对字与字之间的关系进行建模，在较少参数的情况下，依旧保持了较好的分词精度。

在基于字表示的分词模型中，获得有效的字表示是其中非常关键的一步。但是现有的中文字表示工作中，大多数方法直接沿用了第二章中所介绍的英文词向量的生成方法，将字当做模型的处理单元，建立每个字与其上下文字之间的关系 [131]。但是，在中文中，如果想要通过词向量模型，直接获取字的"语义"，可能会遇到一些障碍。如文献 [1] 所述，词是能独立运用的最小语言单位，语义是词或者词组与他们之间含义的关系，直接从字的层面分析语义，可能意义并不大。由于这一原因，本文设计了基于字词联合训练的中文表示技术，该方法使用词的语义空间对字进行建模，为字带来更好的表示。

另一方面，对于中文的词表示，大多数工作同样直接使用了处理英文的方法。然而中文的符号系统是汉字，如果可以有效利用字与字之间的关系，以及字与词之间的关系，势必能为中文词表示带来提升。Chen 等人的工作 [15] 通过假设词的语义由其中字的意思以及词特有的意思融合而成，构造了 CWE 模型。该模型已经超越了直接使用英文词向量模型的效果。本文提出的字词联合训练模型通过对汉字的有效建模，让这些汉字建立了一些词之间的关系，使得词的上下文更丰富，从而提升了词表示的语义。实验结果表明，该模型的效果超过了现有利用字信息提升词义的方法。

本章后续内容安排如下：第 4.2 节介绍了分词和表示学习的相关工作；第 4.3 节介绍了基于字词联合训练的表示方法；第 4.4 节介绍了基于字表示的分词算法框架；第 4.5 节为实验及分析；最后对本章工作进行总结。

## 4.2　相关工作

### 4.2.1　表示学习

传统机器学习方法中，特征的选取是其中最耗时的一个环节。这一环节需要耗费领域专家大量的精力，针对具体任务进行分析，从而设计出有效的特征。表示学习的提出缓解了这一严峻的问题。表示学习的目标是通过算法自动学习得到特征表示，使得机器学习算法可以更高效地运转。

在自然语言中，最基础的语义单元为词。研究人员已经提出了大量词表示的学习方法，如：Skip-gram [73]、CBOW [73]、NNLM [7] 等模型，这些模型在

本文第二章进行了综述，并在第三章做了详细的实验比较。其中无论哪个模型，均基于分布假说，对目标词与其上下文词之间的关系进行建模。因此这类模型的最终效果受到分布假说的制约。

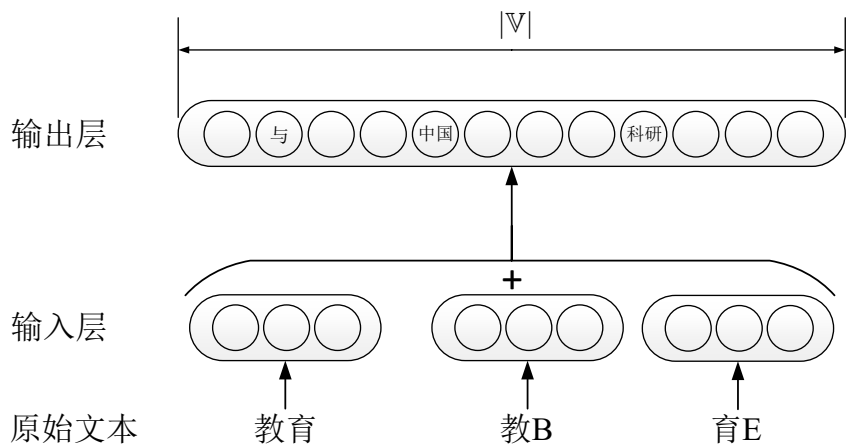

图 4-1　CWE+P 模型结构图

与本文同期，已经有两项工作跳出了分布假说的框架，将词的内部元素考虑到词表示中。在中文词表示方面，Chen 等人在 2015 年发表的文献 [15] 中提出了 CWE+P 等模型 (结构如图 4-1)，这一模型改进自 Skip-gram 模型 (也适用于 CBOW 模型)，将一个词拆分成两部分：词本身和组成这个词的汉字。训练过程中，使用词本身的向量以及组成这个词的各个字向量的平均值表示这个词的语义。

在英文词表示方面，Sun 等人在 2016 年发表的文献 [112] 中提出了 SEING 模型 (结构如图 4-2 所示)。该模型认为英文中具有相同语素（morpheme）的单词具有相似的语义，因此在建模时使用 Skip-gram 的思路，对于目标词不仅预测上下文的词，也预测目标词的所有语素。这套方法也可以近似沿用到中文处理，将英文单词的语素类比为中文的汉字。

文献 [15] 与 [112] 均为对词表示的改进。然而在中文分词、词性标注等任务中，对汉字的表示也是一个必不可少的环节。现有的字表示方法，都直接沿用了英文中对单词的建模方法，将训练语料拆分成字级别，建立语料中每个字与其上下文其它字之间的关系 [131]。对于字表示的方法，需要有进一步的探索。

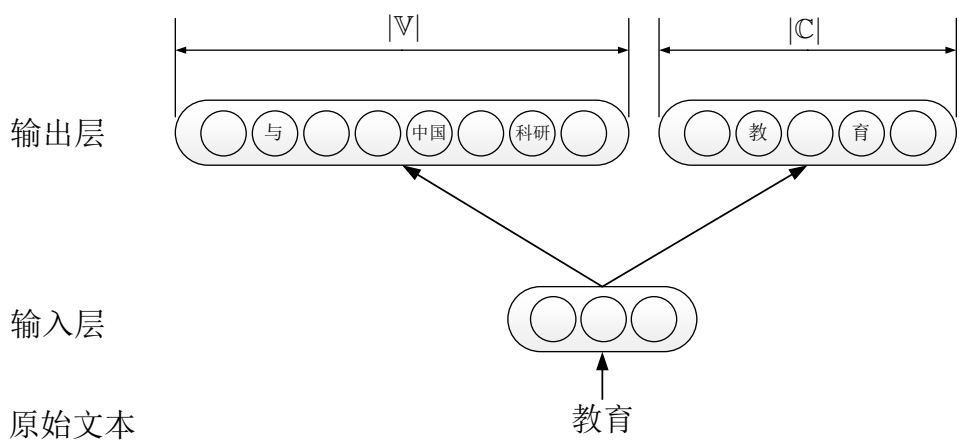

图 4-2 SEING 模型结构图

### 4.2.2 中文分词

传统中文分词方法依赖词典匹配，并通过贪心算法截取可能的最大长度词进行有限的歧义消除。常用的贪心策略有正向最大匹配法、逆向最大匹配法和双向匹配法等。然而，基于词典方法存在两个明显的缺陷，即不能很好地处理词边界歧义和未登录词 (OOV)。为了解决中文分词的这两个关键问题，许多研究工作集中到了基于字标注的机器学习中文分词方法。

基于字标注的中文分词方法基本假设是一个词语内部文本高内聚，而词语边界与外部文字低耦合。通过统计机器学习方法学习判断界是当前中文分词的主流做法。现有工作大多使用序列标注模型执行 BMES 标注。Xue 等人提出了基于 HMM 模型的字标注中文分词方法 [125]。刘群等提出一种基于层叠隐马模型的汉语词法分析方法 [132]。该方法引入角色 HMM 识别未登录词，使用 Viterbi 算法标注出全局最优的角色序列。同时，该方法还提出了一种基于 N-最短路径的策略进行切分排歧。Wang 等人使用基于字分类的 CRF 模型进行中文词法分析 [89]。对基于字标注中文分词方法的改进包括引入更多的标签和设计更有效的特征 [114, 129]、联合使用产生式模型和判别式模型以融合两者的优点 [119] 以及将无监督方法中使用的特征引入有监督方法中 [130] 等。这些传统统计机器学习方法依赖于人工设计的特征，设计特征需要大量的人工参与，设计有效的特征非常费时费力。而将表示学习方法引入机器学习中，可以将特征表示这一步交给算法完成，在一定程度上减少人工，提升效率。

## 4.3 基于字词联合训练的中文表示技术

分布假说认为，词的语义由其上下文决定。根据分布假说构造的词向量模型也在各项任务中取得了一定的成果。但是在中文里，最自然的语言单位是"字"。不同于富含语义信息的词，字仅为记录汉语用的符号系统，本身不具备语义[1]。在一些现有的中文向量表示的工作中，直接将分布语义推广到字表示中，使用上下文中各个字的分布作为当前字的表示[131]。但由于字本身不具备语义，这种方式的效果会受到一定的约束。

为了让字的表示具有更丰富的语义信息，本文借鉴了分布假说的思想，提出利用某个字上下文中各个词的分布，作为这个字的表示。虽然字本身仍然不具备语义信息，但是利用这种表示，把字放入词的语义空间中，通过字词联合训练，可以更有效地对字进行建模。

本文第二章和第三章对若干现有的词向量模型进行了阐述和分析，这里本文选取其中一个对上下文建模最为直接的模型，Skip-gram 模型，并在此基础上实现字词联合训练的想法。如公式 4.1（同公式 2.22）所示，Skip-gram 模型的优化目标为，$w$ 的上下文中的某个词 $w_j$ 对词 $w$ 的条件概率：

$$\sum_{(w,c)\in\mathbb{D}}\sum_{w_j\in c}\log P(w\mid w_j) \tag{4.1}$$

为了实现字词联合训练，本文提出同时优化上下文中某个词 $w_j$ 对目标词 $w$ 的条件概率，以及上下文中各个的汉字 $ch_k$ 对目标词 $w$ 的条件概率：

$$\sum_{(w,c)\in\mathbb{D}}\sum_{w_j\in c}\left((1-\beta)\log P(w\mid w_j)+\beta\frac{1}{|w_j|}\sum_{ch_k\in w_j}\log P(w\mid ch_k)\right) \tag{4.2}$$

式中，$ch_k$ 表示词 $w_j$ 中的汉字，$|w_j|$ 表示词 $w_j$ 的字数。其中归一化项 $\frac{1}{|w_j|}$ 用于使不同字数的词在训练中拥有同样的地位。模型结构如图 4-3。

在字词联合训练中，不仅每个词具有对应的词向量，每个字也具有对应的字向量。词向量和字向量的维度相同，并且根据目标函数（公式 4.2），字和词的向量表示在同一个语义空间中。

模型的训练方法与第二章中介绍的各个词向量模型相同，采用随机梯度下降法求解各个词以及各个字的向量表示。

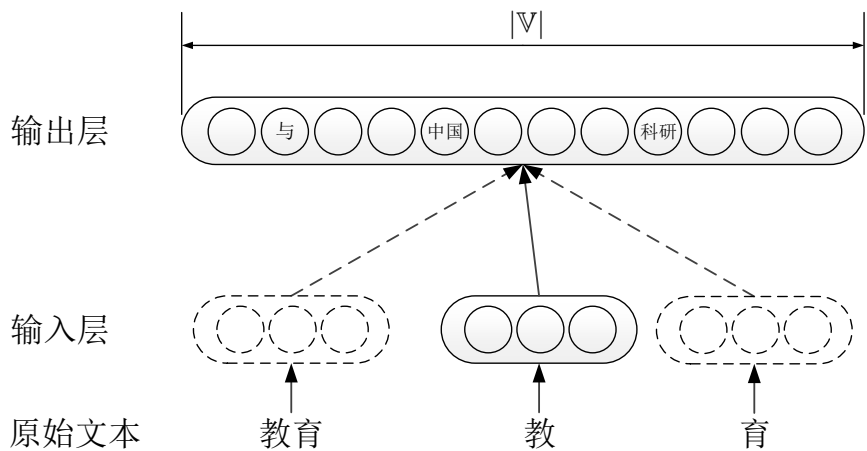

图 4-3  字词联合训练模型结构图

字词联合训练模型相比单独训练字向量，在时间复杂度上只有常数的增加，增幅大约为语料中词的平均字数，一般在 3 左右。因此该方法仍然可以推广到大规模语料中使用。

## 4.4  基于字表示的分词模型

与其它基于字标注的分词方法相似，本文提出的分词模型也采用 BMES 体系对汉字进行标注。对于单字词，其标签为 S；对于多字词，词中的第一个汉字标签为 B，最后一个汉字标签为 E，中间字的标签为 M。对训练数据的每个字进行标注后，本文采用一种 3 层神经网络结构对每个字进行训练，其结构如图 4-4。

对于句子中的每个字的标签分类任务，本文选当当前字以及上下文窗口中，共 $win$ 个字作为特征。其中上文和下文均为 $(win - 1)/2$ 个字。图中最下方为这 $win$ 个字的原始文本 $w_1, w_2, \ldots, w_{win}$，经过第一层，将每个字转换成其字向量表示 $\bm{e}(w_i)$，并把 $win$ 个字连接成一个 $win \times |e|$ 维的向量 $\bm{x}$。该向量是神经网络的输入层：

$$\bm{x} = [\bm{e}(w_1); \ldots; \bm{e}(w_{win})] \tag{4.3}$$

隐藏层 $\bm{h}$ 的设计与普通的前馈神经网络一致，输入层的各个节点与隐藏层的 $|\bm{h}|$ 个节点之间两两均有边连接。隐藏层选用 tanh 函数作为激活函数：

$$\bm{h} = \tanh(\bm{b}^{(1)} + H\bm{x}) \tag{4.4}$$

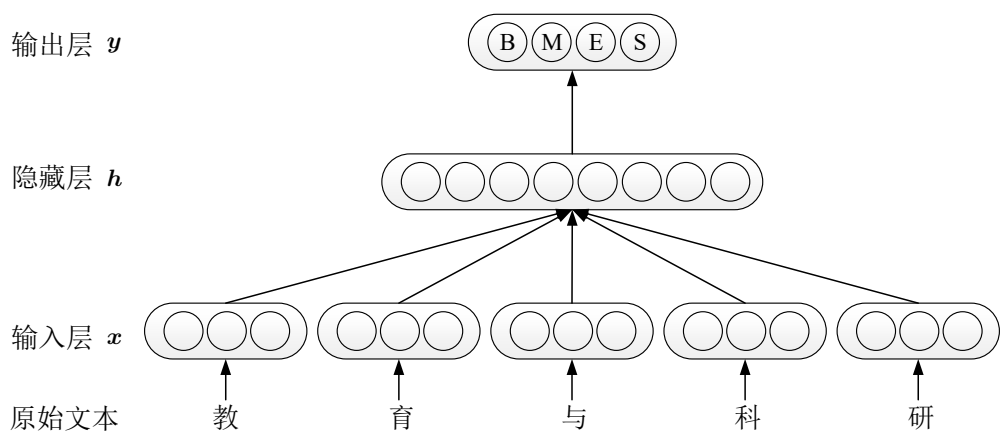

图 4-4 分词算法基本网络结构图

其中 $H$ 为输入层到隐藏层的权重，$\boldsymbol{b}^{(1)}$ 为偏移向量。使用类似的方法，可以将隐藏层转为输出层：

$$\boldsymbol{y} = \boldsymbol{b}^{(2)} + U\boldsymbol{h} \tag{4.5}$$

输出层一共有 4 个节点，使用 softmax[11] 归一化后，分别表示这个字被打上 B、M、E、S 标签的概率：

$$P(i \mid w, \theta) = \frac{\exp\left(\boldsymbol{y}_i\right)}{\sum_{k=1\ldots 4} \exp\left(\boldsymbol{y}_k\right)} \tag{4.6}$$

其中参数 $\theta$ 包含各个字的字向量 $\boldsymbol{e}$，以及网络中的参数 $H$、$U$、$\boldsymbol{b}^{(1)}$、$\boldsymbol{b}^{(2)}$。

对于整个训练语料，本文使用最大似然估计法，即求出一组参数 $\theta$，最大化：

$$\theta \mapsto \sum_{(w,tag_w)\in\mathbb{D}} \log P(tag_w \mid w, \theta) \tag{4.7}$$

其中 $\mathbb{D}$ 是训练语料集，$tag_w$ 是字 $w$ 的正确标签。

本文使用随机梯度下降法 [10] 来优化上述训练目标。每次迭代，随机选取一个样本 $w, tag_w$，使用下式进行一次梯度迭代。式中，$\alpha$ 是学习速率。

$$\theta \leftarrow \theta + \alpha \frac{\partial \log P(tag_w \mid w, \theta)}{\partial \theta} \tag{4.8}$$

## 4.5 实验及分析

为了充分说明字词联合训练的有效性，本文从字的表示和词的表示两方面分别评价字词联合训练方法。下文 4.5.1 小节介绍了字词联合训练的实验设置，包括训练语料与参数；4.5.2 小节介绍了对其中字向量部分的评价；4.5.3 小节介绍了对其中词向量部分的评价；4.5.4 小节探讨了如果在字词联合训练的基础上额外加入上下文的字，会对模型产生什么影响。

### 4.5.1 字词联合训练实验设置

在训练字词向量时，本文使用了两个语料："小语料"为北京大学标注的 SIGHAN 2005 分词数据的训练集，也是 4.5.2 小节中分词实验的训练集，共 179 万字。该语料中的词为人工标注的标准词，与分词实验中的词一致。"大语料"为维基百科的中文语料，共 1.6 亿字。本文使用 ICTCLAS[1]工具包对其进行分词。由于使用工具包进行分词，语料中存在一定的分词错误现象。

本文选取这两个数据集主要出于以下两方面考虑。一、对于评价字表示的实验，"小语料"属于领域内的小规模标准数据集，而"大语料"属于大规模少量噪声的领域外数据集。本文希望通过这两个数据集分析有少量噪声的大规模语料对字向量性能的影响。二、对于评价词表示的实验，本文只使用"大语料"训练字词联合表示。根据第三章的实验结果，使用维基百科语料训练的词向量对于语义类任务的效果最理想，因此在评价字词联合训练的词表示时，也使用通过维基百科训练的词向量。

根据第三章中的经验，本章训练字词向量时，所有字、词向量的维度均为 50；上下文窗口大小为 5；同时，也采用负采样技术对模型进行优化。

### 4.5.2 字表示的实验

根据第三章中的分析，在自然语言处理任务中，如果使用好的词向量作为神经网络模型的初始值，能使模型收敛到更好的局部最优解。因此本文将字向量作为神经网络分词模型（第 4.4 节）的初始值，对不同字向量的效果进行评价。

---

[1]http://ictclas.nlpir.org

**实验设置**

本节采用分词任务对字向量进行评价。分词数据集采用 SIGHAN 2005 bake-off 评测中，北京大学标注的语料。原始语料只包含了训练集与测试集，在实验前，本文将原始语料的训练集前 90% 当作实际的训练集，最后 10% 当作验证集。测试集保持不变。最后训练集共有 1626187 个字，验证集包含了 160898 个字，测试集有 168973 字。

在分词语料中，英文与数字的出现次数较少（甚至有可能 26 个英文字母中有的字母未在训练集中出现过）。为了简化处理流程，本文使用了一个简单的数据预处理步骤，将所有的连续数字字符替换成一个专用的数字标记"NUMBER"，将所有连续的英文字母替换成一个专用的英文单词标记"WORD"。如训练语料"中国/教育/与/科研/计算机网/ (/Ｃ Ｅ Ｒ Ｎ Ｅ Ｔ/) /已/连接/了/２ ０ ０/多/所/大学"经过预处理步骤将会变成"中国/教育/与/科研/计算机网/ (/WORD/) /已/连接/了/NUMBER/多/所/大学"。其中 NUMBER 和 WORD 在训练时都当作一个字符来考虑。这种方法在一定程度上丢失了部分语义信息，会对分词精度产生负面的影响。但是在训练语料不充分的情况下，该预处理可以简化后续步骤，将实验重心放在处理汉字上。

基于字标注的分词模型需要确定上下文窗口的大小，即设定上下文窗口中共多少个字会对当前字的标签产生主要影响。黄昌宁和赵海在文献 [135] 中通过大量实验表明窗口 5 个字可以覆盖真实文本中 99% 以上的情况。因此本文也取上下文窗口为 5，即使用上文 2 个字、下文 2 个字与当前字。这一参数不仅用于神经网络分词模型，同时也用于对比方法中的其它模型。基于字标注的分词模型得到的结果为对各个字打各个标签的概率，当模型对文本中每个字都算出标签概率后，本文使用 Viterbi 算法搜索最优路径，得到最终的分词结果。

**对比方法**

本节实验主要为了证明字词联合训练对字表示的促进作用。因此主要对比模型为不使用字词联合训练的 Skip-gram 模型。在这一对比实验中，首先需要将语料拆分成字，然后直接用 Skip-gram 模型在汉字级别对字进行建模。另一方面，为了说明分词模型的效果，本文同时选用了分词中较常用的最大熵模型，特征选用一元及二元特征。对于语料中的某个字 $ch_k$，其特征向量具体包括：

- 一元特征 $ch_i$，其中 $i$ 为 $\{k-2, k-1, k, k+1, k+2\}$，如果 $ch_i$ 超出了句子的边界，则使用一个特殊的符号"PADDING"来代替。
- 二元特征 $ch_i\_ch_{i+1}$，其中 $i$ 为 $\{k-2, k-1, k, k+1\}$，如果 $c_i$ 或 $c_{i+1}$ 超出了句子的边界，则忽略这个特征。

以上各特征的权重均为 1。具体而言，本文设计了两个基准实验，第一个实验只使用了上述的一元特征，在后文中称作"最大熵一元特征"；第二个实验同时使用了一元特征和二元特征，在后文中称作"最大熵二元特征"。

**实验结果及分析**

| 模型 | 准确率 | 召回率 | F 值 |
|---|---|---|---|
| Sighan2005 最佳成绩（封闭） | 94.6 | 95.3 | 95.0 |
| Sighan2005 最佳成绩（开放） | 96.8 | 96.9 | 96.9 |
| 最大熵一元特征 | 86.8 | 86.6 | 86.7 |
| 最大熵二元特征 | 95.3 | 94.5 | **94.9** |
| 随机字向量 | 93.43 | 92.91 | 93.17 |
| 小语料字向量 | 93.76 | 93.17 | 93.46 |
| 大语料字向量 | 93.66 | 93.29 | 93.48 |
| 小语料字词向量 | 93.96 | 93.21 | 93.58 |
| 大语料字词向量 | 94.02 | 93.28 | **93.65** |

表 4-1　各模型在中文分词任务上的表现

为了展示不同字向量对分词系统的性能影响，本文设计了多组对比实验：

- 随机字向量：使用随机赋值的字向量，对 4.4 节中的模型进行初始化。
- 小语料字向量：使用 Skip-gram 模型直接在小语料上训练字向量，作为网络的初始值。
- 大语料字向量：使用 Skip-gram 模型直接在大语料上训练字向量，作为网络的初始值。
- 小语料字词向量：使用字词联合训练在小语料上训练字向量，作为网络的初始值。
- 大语料字词向量：使用字词联合训练在大语料上训练字向量，作为网络的初始值。

　　表 4-1 中列举了本文所做的 7 组实验。其中"最大熵一元特征"和"最大熵二元特征"为上一小节中描述的两个基准实验，后面各实验为不同的字向量对分词性能的影响。

　　通过这些实验，可以发现，在神经网络模型中，无论使用哪种技术生成的字向量，均能有效提升模型的性能。为了更直观地评估各种字向量对分词模型的帮助，本文使用第三章第 3.3.1 小节（第 33 页）中提出的性能增益率来评价这几种不同的字向量。以随机字向量作为基准，四种字向量中 F 值最高的大语料字字向量作为对比，各字向量的性能增益率为：小语料字字向量 60%，大语料字字向量 65%，小语料字词向量 85%，大语料字词向量 100%。

　　本文对比在两种语料下，字词联合训练与单独训练字表示。实验表明，字词联合训练在两种语料下，在性能增益率上分别有 25 到 35 个百分点的提升，充分说明了字词联合训练对于中文字表示有非常显著的提升。

　　对比不同语料下，字向量带来的性能提升，可以发现，在单独对字进行训练时，大语料能对字向量带来微弱的提升；而如果使用字词联合训练，大语料相对小语料能明显提高字向量的性能。本文认为之所以使用字词联合训练时，大语料更能体现出其效果，是因为如果只针对字进行训练，字向量的上下文空间局限在字符层面，相比而言，字词联合训练时，字向量的上下文空间为词的语义空间，信息更为丰富。而维基百科大语料中本身拥有了大量的语义信息，只有通过表达能力强的表示才能充分捕获到其中的信息。

　　综合对比基于字向量的神经网络分词模型与传统的机器学习分词方法（基准实验），神经网络分词模型虽然也只使用了各个词中的字的表示作为输入，相比使用一元特征的传统分词方法，有非常显著的提升。然而不可否认，神经网络分词模型相比二元特征仍然有一定的差距。这些差距可能需要更有效的字表示技术以及更好的神经网络模型来弥补。Collobert 在文献 [18] 中也得到了类似的结论，在词性标注、命名实体识别等序列标注任务中，他们的神经网络方法与传统基于特征工程的机器学习方法仍然有一些差距。他们在实验中指出，神经网络模型配合少量人工的先验知识，就可以达到以往通过人工精心设计特征才能达到的性能。

### 4.5.3　词表示的实验

　　为了说明字词联合训练对词表示的作用，本节使用语义相关性任务和文本分类任务对联合训练中得到的词表示进行评价，并与 Chen 等人 [15] 与 Sun 等

人 [112] 的模型进行对比分析。

## 评价方法

与第三章中对英文词向量的评价方法一致，本节对中文词向量的评价同样从三个方面进行：一、利用词向量的语言学特性完成任务；二、将词向量作为特征，提高自然语言处理任务的性能；三、将词向量作为神经网络的初始值，提升神经网络模型的优化效果。本文从这三类评价方法中，各选取一个任务评价不同模型得到的词向量。

在评价词向量的语言学特性方面，本节选取了语义相关性任务。该任务与第三章 3.2 小节提到的 ws 任务类似，这里选用文献 [15] 提供的两个中文语义相关性数据集进行评价。这两个数据集分别含有 297 个词对以及 240 个词对，在下文中简写为 ws297 和 ws240。数据集中的每个词对都有若干位标注者对其进行打分（两个数据集的打分范围分别为 0 到 10 以及 0 到 5），分数越高表示标注人员认为这两个词的语义更相关或者更相似。例如，词对"饮料、汽车"的平均打分为 1.23，而词对"学校、学生"的打分为 8.71。评价时，对于每个词对，本文使用所有标注者打分的平均值作为参考得分 $X$，以词对中两个词的词向量的余弦距离作为模型得到的相关性得分 $Y$，并衡量 $X, Y$ 这两组数值之间的皮尔逊相关系数。具体定义可见 3.2 小节中的公式。

在将词向量作为特征提升系统性能方面，本节选用基于平均词向量的文本分类方法。该方法与第三章 3.2 小节提到的 avg 任务一致，后文同样简写为 avg。实验选取复旦文本分类语料[2]，使用 Logistic 回归模型完成文本分类任务。

在将词向量作为初始值提升神经网络模型性能方面，本节选取卷积神经网络和循环卷积网络完成文本分类任务。其中卷积神经网络与第三章 3.2 小节提到的 cnn 任务一致，而循环卷积网络则是本文第五章提出的神经网络文档表示模型，这两个任务在后文中简写为 cnn 和 rcnn。实验选用的语料同样是复旦文本分类语料，模型参数与第五章中所用参数一致。

## 对比方法

本节实验希望验证字词联合训练对于词的语义有提升。为了说明这一点，实验的主要对比方法为传统的 Skip-gram 模型，该模型直接对词进行建模，而不

---

[2]http://www.datatang.com/data/44139, http://www.datatang.com/data/43543

考虑字在其中的作用。本文提出的字词联合训练在 Skip-gram 模型的基础上加入了对字的建模，因此对比 Skip-gram 模型的提升，就是字词联合训练对词表示带来的实际提升。

进一步地，为了说明字词联合训练的有效性，本文还将 Chen 等人的 CWE+P 模型 [15] 与 Sun 等人的 SEING 模型 [112] 纳入对比。Chen 等人的模型同样也是对 Skip-gram 模型的改进，这种方法将一个词拆分成两部分：词和组成这个词的各个汉字，在表示这个词的语义时，使用词本身的向量以及组成这个词的各个字向量的平均值。Sun 等人的方法原本只对英文进行建模，该模型认为英文中具有相同语素的单词具有相似的语义，因此在建模时使用 Skip-gram 的思路，对于目标词不仅预测上下文的词，也预测目标词的所有语素。这套方法也可以近似沿用到中文处理，将英文单词的语素类比为中文的汉字。

**实验结果及分析**

| 模型 | ws240 | ws297 |
|---|---|---|
| CWE+P [15] | 44.07 | 53.43 |
| SEING [112] | 42.84 | 49.80 |
| Skip-gram | 43.51 | 52.53 |
| 字词联合训练 | **46.68** | **54.45** |

表 4-2　各模型在中文语义相关性任务上的表现（×100）

表 4-2 展示了不同方法在词类比任务中的效果。实验中，设定字词联合训练的参数 $\beta = 0.5$，即各模型对字和词的建模比例为 $1:1$。对于 CWE+P 模型与 SEING 模型，文献 [15] 和 [112] 并没有对字和词的建模比例进行考虑，也可以认为这两个模型对字词的建模比例为 $1:1$。

为了更进一步说明本文提出的字词联合训练的有效性，这里调节公式 4.2 中的 $\beta$，分析在不同的 $\beta$ 下，字词联合训练的效果。作为对比，实验中同样调节了 CWE+P 与 SEING 模型的字词建模比例。对于 CWE+P 模型，可以直接使用 CWE 开源工具包[3]，设定 char-rate 参数为 $\beta/(1-\beta)$；对于 SEING 模型，计算每个样本时，设定有 $\beta$ 的比例对词中的字进行预测，有 $1-\beta$ 的比例对词的上下文进行预测。图 4-5 展示了实验结果。参数 $\beta$ 的含义为在字词联合训练中，对

---

[3] https://github.com/Leonard-Xu/CWE

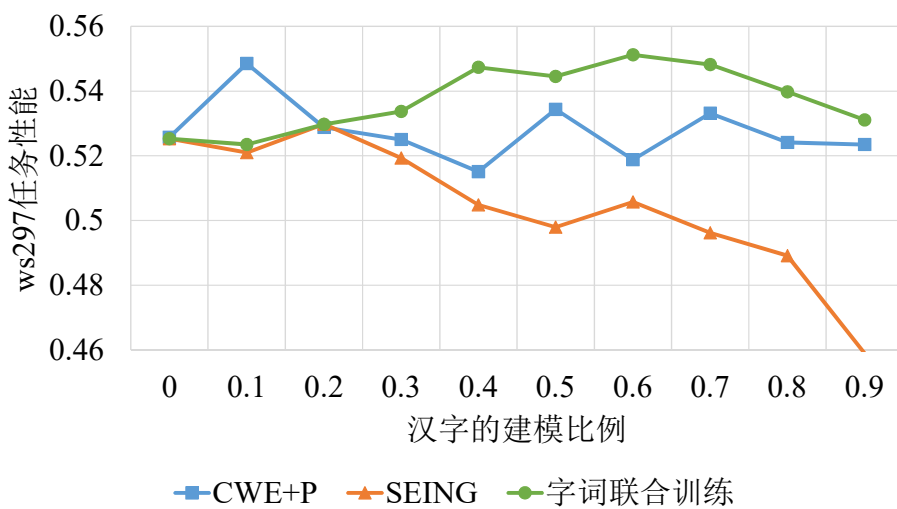

图 4-5　各模型汉字的建模比例对词义的影响

字的建模占多大的比例。如果 $\beta = 0$ 则只对词进行建模，三个模型均等价于原始的 Skip-gram 模型；如果 $\beta = 1$ 说明模型只对字进行建模。实验中尝试了 $\beta$ 从 0 到 0.9 中的各个值。从整体上看，逐渐增加汉字建模比例时，字词联合训练对词义的建模呈现出先上升后下降的趋势；CWE+P 模型也有类似的趋势，峰值处汉字的建模比例为 10%（性能为 0.5486，略低于字词联合训练）；而 SEING 模型几乎随着对汉字建模比例的增加，效果一直变差。

本文提出的字词联合训练与 CWE+P 模型或者 SEING 模型均为对字（或者语素）和词同时建模，虽然建模方式有所不同，但是为什么这三种方法在实验中呈现出较大的差异？尤其是 SEING 模型，对于英文词义表示有显著的提升 [112]，然而在中文词义相似度任务中却起到了负面效果。为了回答这一问题，本文首先分析这三种模型对字词建模的区别。

**CWE+P** 模型的基础假设是词的语义可以拆解成两部分：词中的各个汉字（模型名称中的 +P 表示汉字需加上前后缀的标记）的语义，以及这个词特有的语义。在文献 [15] 中，作者将这两部分组合成实际词向量时，使用了向量加法。在这种设定下，字词所占语义比例需要仔细调节，如果比例过高，则词的语义几乎直接由字决定，在现代汉语中这一点并不成立。从前面的实验中也看出，字的比例在 10% 左右时可以让词义达到最佳效果，原文中设定的 50% 很可能不是最佳参数。从另一个角度看，由于目前 CWE 模型只使用了朴素的向量加法作为词义合成，如果选用更合理的组合函数，应当能达到更好的效果。

**SEING** 模型的基础假设是，具有相似语素的单词，有相似的含义，类比到中文即具有相似汉字的词，有相似的含义。从前面的实验中可以看出，这一假设从英文沿用到中文时，效果并不理想。造成中英文区别的原因可能有两点：一、中文的语素（morpheme）并非直接为汉字。虽然大多数情况下中文汉字就是语素，但是也存在一些多个字组成的语素（多音节语素），如"苏打"、"巧克力"。二、英文中的语素多为表义的词根、固定的前缀后缀等（如 unbreakable 可以分解成 un、break、able 三个语素），而汉字在组词时，相比英文更为灵活，大多数汉字既可以作为前缀，又可以作为后缀（如星期天、天空），即使同作为前缀或后缀，意思也不尽相同（如夏天、晴天）。综合分析这两点，从统计上看，第一点中提到的多音节语素在中文语料中的比例较低，而第二点区别在中英文中普遍存在，更可能是影响 SEING 模型性能的核心原因。因此，将 SEING 模型的基础假设直接沿用到中文时，即具有相同汉字的词拥有相似的语义，或许并不合适。

本文提出的**字词联合训练**从模型上看，只增加了对汉字上下文的建模，并没有修改词的建模。但是实际上，由于汉字的出现，一些原本没有联系的上下文，被联系了起来。也就是说，汉字对词起到了平滑作用，在一定程度上扩充了上下文，从而提升了词表示。与此同时，由于汉字的上下文也是根据分布假说构造得到的，因此即使汉字的建模比例不合理，一般情况下也不会产生负面作用。

综合分析这三个模型，CWE+P 模型与本文的字词联合训练模型，遵循了分布假说，词义由其上下文决定；而 SEING 模型要求每个词有两个分布，不仅需要符合上下文的分布，还需要符合内部语素的分布，某种程度上说，这种方法破坏了分布假说，对于中文词表示效果不理想。对比 CWE+P 模型和字词联合训练模型，CWE+P 模型利用构词法的思想认为词的意思部分由字决定，而字词联合训练模型扩展了分布假说，对字的上下文直接进行建模。尽管从本文的实验中看，字词联合训练略胜一筹，但是 CWE+P 模型通过调节参数，或者变化构词方法，仍然很有潜力。

为了更直观地感受这些模型之间的差异，本文选取若干词，以及通过这些模型得到的最近邻。结果列于表 4-3。

从第一个例子"巡逻"中可以看出，三种对汉字建模的模型均可以在 Skip-gram 的基础上，让字面上更相似的词在空间距离中的距离更短。如"巡逻队"、"巡

| 模型 | Skip-gram | CWE+P | SEING | 字词联合训练 |
|------|-----------|-------|-------|-------------|
| 巡逻 | 值勤<br>巡逻队<br>侦察<br>护航<br>执勤 | 巡逻队<br>巡逻艇<br>执勤<br>护航<br>巡逻员 | 巡逻队<br>巡逻艇<br>巡逻舰<br>侦察<br>防空 | 巡逻队<br>值勤<br>巡逻艇<br>侦察<br>蛙人 |
| 星期天 | 周日<br>周末<br>星期四<br>礼拜天<br>星期六 | 星期二<br>星期三<br>星期四<br>星期五<br>星期六 | 星期二<br>周日<br>星期日<br>星期五<br>星期六 | 周日<br>星期六<br>星期二<br>星期日<br>礼拜天 |
| 风光旖旎 | 湖光山色<br>叠翠<br>草泽<br>满陇桂<br>丰茂 | 湖光山色<br>山清水秀<br>风动石<br>茶园<br>风光 | 旖旎<br>风光<br>湖光山色<br>碧波<br>风景点 | 湖光山色<br>葱郁<br>丰茂<br>山清水秀<br>叠翠 |

表 4-3　不同字词模型得到的最近邻对比表

逻艇"这些词，在 Skip-gram 模型中排名比较靠后，而融入了对汉字的建模之后，排名均有提升。第二个例子"星期天"可以看出，CWE+P 和 SEING 模型更多地考虑了字面上的相似性（"星"、"期"二字），最近邻中主要为星期中的其它日子。而字词联合训练更多地保留了语义信息，不仅得到了 Skip-gram 模型原本就能得到的"周日"和"礼拜天"，再配合字对词的平滑，可以发现"星期日"这个字面上更相似的同义词。第三个例子本文选取了低频词"风光旖旎"，所有模型均可以发现近义词"湖光山色"。SEING 模型对字面相似有单独的建模，因此最相似的两个词是"旖旎"和"风光"；CWE+P 模型中字对词义的影响也比较大，因此"风动石"、"风光"也排在了靠前的位置；字词联合训练通过平滑得到的词多为形容风景秀丽的词，相比 Skip-gram 模型得到的"草泽"等词，与原词的意思更匹配。

　　尽管在词义相似度任务中，不同的模型得到的词向量展现出了不同的语义特性，但是在文本分类任务中，这些模型相比 Skip-gram 模型，都能一致地提升性能，具体结果列于表 4-4 中。从这些结果中可以看出，Skip-gram 模型生成的词向量，比随机词向量的性能有显著的提升，进一步地，融入了字信息的三种模

| 模型 | avg | cnn | rcnn |
|------|-----|-----|------|
| 随机词向量 | 80.75 | 93.57 | 94.89 |
| Skip-gram | 85.53 | 94.04 | 95.20 |
| CWE+P | 86.15 | 94.17 | 95.27 |
| SEING | 85.66 | 94.15 | **95.32** |
| 字词联合训练 | **86.23** | **94.22** | 95.30 |

表 4-4 各模型在中文文本分类任务上的表现

型得到的词向量，比 Skip-gram 均有一定程度的提升。这三种文本分类模型中，avg 使用了结构最简单的 Logistic 回归，cnn 使用了相对较为复杂的卷积神经网络，而 rcnn 使用的是本文提出的循环卷积网络。从实验结果中可以看出，在相对简单的模型中（如 avg），使用更好的词向量，可以带来较大幅度的提升；而在相对复杂的模型中（如 rcnn），词向量带来的提升较少。纵观这三种模型，以及五种不同的词向量，可以很明显地看出，对于文本表示，模型结构的影响远远大于初始值（或者特征）的选择。这也是后文深入探索文本表示模型的一个出发点。

根据上述实验及分析，本文认为字词联合训练通过字对词的平滑，将更多的词和上下文建立起了联系。这种方式在保留分布假说的基础上，可以使用更丰富的上下文信息，也得到了更有效的词表示，在词义相似度任务以及文本分类任务中，相比此前的方法均有一定的提升。

### 4.5.4 上下文加入字的影响

在本文提出的字词联合训练模型中，将词引入到字的上下文空间中，前面的实验已经说明，这种方式无论对于字表示还是词表示均有一定的提升。

对应于将词引入到字的上下文空间中，本文也同时尝试了将字引入到词的上下文空间中。具体方法为，通过将上下文各词拆分成字，将上下文词以及对应的字全部作为目标词的上下文。通过这种改造之后，同样使用分词和词义相似度任务进行评价。结果列于表 4-5。表中，分词使用 F 值作为评价指标，ws240 和 ws297 使用皮尔逊相关系数作为指标。

从实验结果中看，加入字作为词的上下文，对分词实验有小幅提升；然而对于词的语义相似度任务，性能有非常严重的下降。为了更直观地说明造成这一现象的原因，这里同样展示若干词的最近邻，列于表 4-6。

| 模型 | 分词 | ws240 | ws297 |
|---|---|---|---|
| Skip-gram 字向量 | 93.48 | - | - |
| Skip-gram 词向量 | - | 43.51 | 52.53 |
| 字词联合训练 | 93.65 | **46.68** | **54.45** |
| 字词联合训练 + 字上下文 | **93.69** | 37.11 | 43.16 |

表 4-5　字词联合训练时，加入字作为词的上下文的实验结果（×100）

| 模型 | Skip-gram | 字词联合训练 | 字词联合训练 + 字上下文 |
|---|---|---|---|
| 江 | 泾<br>江口<br>青弋<br>西江<br>泗 | 浙<br>奉化<br>江阴<br>宁海<br>太仓 | 浙<br>陕<br>泗<br>潭<br>湘 |
| 星期天 | 周日<br>周末<br>星期四<br>礼拜天<br>星期六 | 周日<br>星期六<br>星期二<br>星期日<br>礼拜天 | 周日<br>热天<br>月姬<br>周末<br>周六 |

表 4-6　字词联合训练中加入字作为词的上下文，得到的最近邻对比表

　　由于在词的上下文中加入字之后，在字、词评价任务上有不同的结论，因此这里分别从字和词两个角度进行分析。对于字"江"，加入字作为上下文的模型得到的最近邻，均为含义上比较相关的汉字；而 Skip-gram 模型得到的最近邻主要是一些江的名字；字词联合训练得到的最近邻主要是浙江和江苏的城市名。单独从最近邻的相关性上分析，这三个模型从不同的角度描述了"江"所包含的意思，但是对于分词任务而言，由于字是其中最主要的特征，一份准确刻画不同字之间的关系的字向量，应当对于分词有促进作用。对于词"星期天"，加入了字作为上下文的模型，得到的最近邻里出现了一些语义上不太相关的词，如"热天"。本文认为这是因为字本身并没有明确的语义，一个字在不同的词中很可能有完全不同的意思。因此，在上下文中融入了汉字，反而让上下文分布变得更模糊，从而导致弱化了词的语义。

　　综上所述，本文认为在上下文中加入汉字，可以直接建立起汉字之间的关

系，对于生成的字向量有一定的提升，通过分词任务评价，也可以看出相比单纯的字词联合训练有小幅提升。但是当上下文中混入汉字之后，由于这些字并不直接拥有明确的语义信息，反而干扰了词的上下文分布，对词表示起到了负面作用。因此，在使用字表示时，可以考虑在词中加入字作为上下文，而在使用词表示时，只使用字词联合训练模型则更合适。

## 4.6 本章小结

本章提出字词联合训练方法，在字的上下文中引入了词，使用词的语义空间对字进行建模，获得更好的字表示。相比直接沿用英文词向量的训练方法，字词联合训练在中文分词任务上，性能增益率有 35 个百分点的提升。另一方面，由于在字词联合训练中，增加了对汉字的建模，这些汉字建立了一些词之间的关系，使得词的上下文更丰富，提升了词表示的语义。在中文词义相似度任务中，字词联合训练比现有两种借助汉字提升词表示语义的模型更为有效。在文本分类任务中，使用字词联合训练得到的词向量在用做特征以及神经网络初始值时，相比此前方法也有一定的提升。此外，实验结果还指出，大规模少量噪声的语料相比小规模几乎无噪声的语料依然有较大的优势，因此本文提出的字词联合训练方法，有潜力在更大规模的语料上发挥出自己的优势。

# 第五章  基于循环卷积网络的文档表示及应用

在文本分类、信息检索等实际任务中，不仅需要词级别的语义表示，更需要句子和文档级别的语义表示。本章首先总结了现有神经网络模型对文档的表示方法，然后提出了基于循环卷积网络的文档表示方法。该方法可以在线性时间复杂度下，对不同距离的上下文进行自适应建模，得到有效的文本表示，并在文本分类任务上，相对现有方法有显著的提升。

## 5.1  引言

文本分类在网页检索、信息筛选、情感分析等任务中是一个至关重要的步骤 [2]。文本分类中的关键问题在于文本表示，在传统机器学习方法中，通常以特征表示的形式出现。文本分类中最常用的特征表示方法是词袋子模型。词袋子模型中，最常用的特征是词、二元词组、多元词组（n-gram）以及一些人工抽取的模板特征。在以特征的形式表示文本之后，传统模型往往使用词频、互信息 [19]、pLSA [13]、LDA [38] 等方法筛选出最有效的特征。然而，传统方法在表示文本时，会忽略上下文信息，同时也会丢失词序信息。比如以下例子：

> A sunset stroll along the South Bank affords an array of stunning vantage points.

分析句子中的"Bank"一词，如果孤立地看这个词，我们并不知道这是表示"河岸"还是"银行"，这时就需要通过上下文对词义进行消歧。当纳入"Bank"的前一个词，看二元词组"South Bank"时，可以发现两个单词的首字母都是大写的，这对于不了解伦敦的人来说，很可能会以为是"南方银行"。当看了足够的上下文"stroll along the South Bank"，我们才可以肯定这里说的是南岸，和银行无关。尽管传统特征中诸如多元词组以及更复杂的特征（如树核 [94]）也能捕获词序信息，但是这些特征往往会遇到数据稀疏问题，影响到文本分类的精度。

近年来，预训练词向量以及深度神经网络模型为自然语言处理带来了新的思路。本文的第二章和第三章已经介绍了词向量可以从无标注的文本中自动学习得到语义和语法信息，并且能对各项自然语言处理任务带来性能上显著的提升。在词向量的帮助下，有人提出一些组合语义的方法来表示文本的语义。

Socher 等人在 2011 年 [105, 109]、2013 年 [110] 发表了**递归神经网络**（Recursive Neural Network）的相关工作。该方法被证实在构建句子级语义时较为有效。然而，递归神经网络需要按照一个树形结构来构建句子的语义，其性能依赖于构建文本树的精度。而且，构建这棵树需要至少 $O(n^2)$ 的时间复杂度，其中 $n$ 表示句子的长度。当模型在处理长句子或者文档时，所花费的时间往往是不可接受的。更进一步地，在做文档表示时，两个句子之间的关系不一定能构成树形结构。因此递归神经网络可能不适合构建长句子或者文档的语义。

**循环神经网络**（Recurrent Neural Network）可以在 $O(n)$ 时间内构建文本的语义 [26]。该模型逐词处理整个文档，并把所有上文的语义保存到一个固定大小的隐藏层中。循环神经网络的优势在于它可以更好地捕捉上下文信息，对长距离的上下文信息进行建模。然而，循环神经网络是一个有偏的模型，如对于正向的循环神经网络而言，文本中靠后的词相对靠前的词占据了更主导的地位。由于这一语义偏置的特性，循环神经网络在构建整个文本的语义时，会更多地包含文本后面部分的信息。但是实际上并非所有文本的重点都放在最后，这可能会影响其生成的语义表示的精确度。

为了解决语义偏置的问题，有人提出用**卷积神经网络**（Convolutional Neural Network）来构建文本语义 [18]。卷积神经网络利用最大池化技术能从文本中找出最有用的文本片段，其复杂度也是 $O(n)$。因此卷积神经网络在构建文本语义时有更大的潜力。然而，现有卷积神经网络的模型总是使用比较简单的卷积核，如固定窗口 [18, 46]。在使用这类模型时，如何确定窗口大小是一个关键问题。当窗口太小时，可能导致上下文信息保留不足，难以对词进行精确刻画；而当窗口太大时，会导致参数过多，增加模型优化难度。因此，需要考虑，如何构建模型，才能更好地捕获上下文信息，减少选择窗口大小带来的困难。并以此为基础来更好地完成文本分类的任务。

为了解决上述模型的缺陷，本文提出了**循环卷积网络**（Recurrent Convolutional Neural Network），并将其用到文本分类中。首先，本文使用一个双向循环结构对上下文进行建模。对比基于窗口的上下文建模方法，循环结构的参数更少，在引入较少噪声的前提下，可以捕获尽可能远的上下文信息，从而保留长距离的词序信息。第二，本文使用最大池化技术自动判断对文本分类最重要的特征。结合这两者，循环卷积网络同时拥有了循环神经网络和卷积神经网络的优势，既能很好地刻画上下文信息，又能无偏地描述整个文本的内容，并且其

复杂度仅为 $O(n)$。

本文在四个不同的数据集上，对比了循环卷积网络与此前最好方法的性能。这四个数据集包含了中英文的文本分类任务，分类体系包括学科分类、作者母语分类和情感分类。实验结果表明，本文的方法相对此前的方法有明显的优势。

## 5.2　相关工作

本节首先介绍现有神经网络文档向量表示技术，这类技术均均基于组合语义的思路，从词的语义组合得到文档的语义。5.2.1 小节介绍了组合语义的思想及起源，5.2.2 小节到 5.2.4 小节介绍了三种现有的文档向量表示方法。最后 5.2.5 小节简述了传统文本分类方法。

### 5.2.1　组合语义

在获取句子和文档的语义表示时，很容易想到直接沿用词的分布假说，对文档进行建模。然而，如果采用分布假说直接生成句子或者文档的向量表示，会遇到极大的数据稀疏问题。Koehn 在 2005 年发表的文献 [53] 中曾对 Europarl 语料进行了统计，结果表明，在该语料中，总共有 47,889,787 个词，其中有 304,786 个不同的词，出现至少 10 次的词一共有 58,552 个；而语料中一共有 1,920,209 个句子，其中不同的句子有 1,860,118 个，至少出现 10 次的句子更是只有 597 句。因此，如果将句子看成一个整体，用词向量模型来训练句子的表示，由于绝大多数句子因为只出现过一次，训练的结果将毫无统计意义。另一方面，分布假说是针对词义的假说，这种通过上下文获取语义的方式对句子和文档是否有效，还有待讨论。因此需要寻求新的思路对句子和文档进行建模。

德国数学家弗雷格（Gottlob Frege）在 1892 年就曾提出：一段话的语义由其各组成部分的语义以及它们之间的组合方法所确定 [30]。现有的句子或者文档表示也通常以该思路为基础，通过语义组合的方式获取。常用的组合语义组合函数，如线性加权、矩阵乘法、张量乘法等，在 Hermann 的文献 [36] 中有详细的总结。近年来基于神经网络的语义组合技术为文档表示带来了新的思路。从神经网络的结构上看，主要可以分为三种方式：递归神经网络（5.2.2 小节）、循环神经网络（5.2.3 小节）和卷积神经网络（5.2.4 小节）。

### 5.2.2    递归神经网络

递归神经网络（Recursive Neural Network）的结构如图 5-1，其核心为通过一个树形结构，从词开始逐步合成各短语的语义，最后得到整句话的语义。

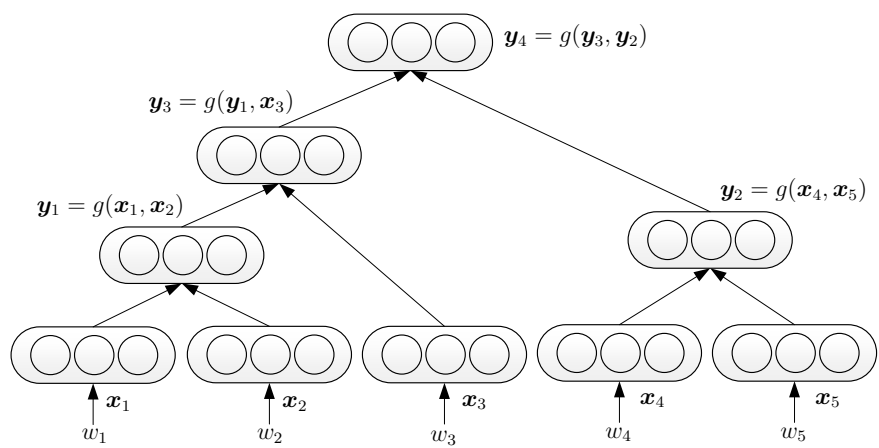

图 5-1    递归神经网络模型结构图

递归神经网络使用的树形结构一般为二叉树，在某些特殊情况下（如依存句法分析树 [107]）也使用多叉树。本文主要从树的构建方式和子节点到父节点的组合函数，这两方面介绍介绍递归神经网络。

树形结构有两种方式生成：一、使用句法分析器构建句法树 [105, 110]；二、使用贪心方法选择重建误差最小的相邻子树，逐层合并 [109]。这两种方法各有优劣，使用句法分析器的方法可以保证生成的树形结构是一棵句法树，树中各个节点均对应句子中的短语，通过网络合并生成的各个节点的语义表示也对应各短语的语义。使用贪心方法构建树形结构则可以通过自动挖掘大量数据中的规律，无监督地完成这一过程，但是树中的各个节点不能保证有实际的句法成分。

子节点到父节点的组合函数 $y = f(a, b)$ 主要有三种：

一、句法组合。这种方式下，子节点的表示为向量 $a, b$，父节点可以通过矩阵运算得到：

$$y = \phi\left(H\left[a; b\right]\right) \tag{5.1}$$

其中 $\phi$ 为非线性的激活函数，权重矩阵 $H$ 可能固定 [108]，也可能根据子树对应的句法结构不同，而选用不同的矩阵 [103]。该方法一般用于句法分析中。

二、矩阵向量法 [78]。在这种表示下，每个节点由两部分表示组成，一个矩阵和一个向量，对于 $A, \boldsymbol{a}$ 子节点和 $B, \boldsymbol{b}$ 子节点，其组合函数为：

$$\boldsymbol{y} = \phi\left(H\left[B\boldsymbol{a}; A\boldsymbol{b}\right]\right) \tag{5.2}$$

$$Y = W_M \begin{bmatrix} A \\ B \end{bmatrix} \tag{5.3}$$

其中 $W_M \in \mathbb{R}^{|\boldsymbol{a}| \times 2|\boldsymbol{a}|}$，保证父节点对应的语义变换矩阵 $Y \in \mathbb{R}^{|\boldsymbol{a}| \times |\boldsymbol{a}|}$，与叶节点的 $A$、$B$ 矩阵维度一致。使用这种方法，每个词均有一个语义变换矩阵，对于否定词等对句法结构另一部分有类似影响的词而言，普通的句法组合方式没法很好地对其建模，而这种矩阵向量表示则可以解决这一问题。Socher 等人将该方法用于关系分类中 [106]。

三、张量组合。张量组合方式使用张量中的每一个矩阵，将子节点组合生成父节点表示中的一维。

$$\boldsymbol{y} = \phi\left([\boldsymbol{a}; \boldsymbol{b}]^{\mathsf{T}} W^{[1:d]} [\boldsymbol{a}; \boldsymbol{b}] + H[\boldsymbol{a}; \boldsymbol{b}]\right) \tag{5.4}$$

其中 $W^{[1:d]}$ 表示张量 $W$ 中的第 1 到 $d$ 个切片矩阵。不同的切片用于生成父节点 $\boldsymbol{y}$ 中不同的维度。该方法是句法组合方法的泛化形式，有更强的语义组合能力，Socher 等人将其用于情感分析任务中 [104]。

递归神经网络在构建文本表示时，其精度依赖于文本树的精度。无论使用哪种构建方式，哪种组合函数，构建文本树均需要至少 $O(n^2)$ 的时间复杂度，其中 $n$ 表示句子的长度。当模型在处理长句子或者文档时，所花费的时间往往是不可接受的。更进一步地，在做文档表示时，两个句子之间的关系不一定能构成树形结构。因此递归神经网络在大量句子级任务中表现出色，但可能不适合构建长句子或者文档级别的语义。

### 5.2.3　循环神经网络

循环神经网络（Recurrent Neural Network）由 Elman 等人在 1990 年首次提出 [26]。该模型的核心是通过循环方式逐个输入文本中的各个词，并维护一个隐藏层，保留所有的上文信息。

循环神经网络是递归神经网络的一个特例，可以认为它对应的是一棵任何

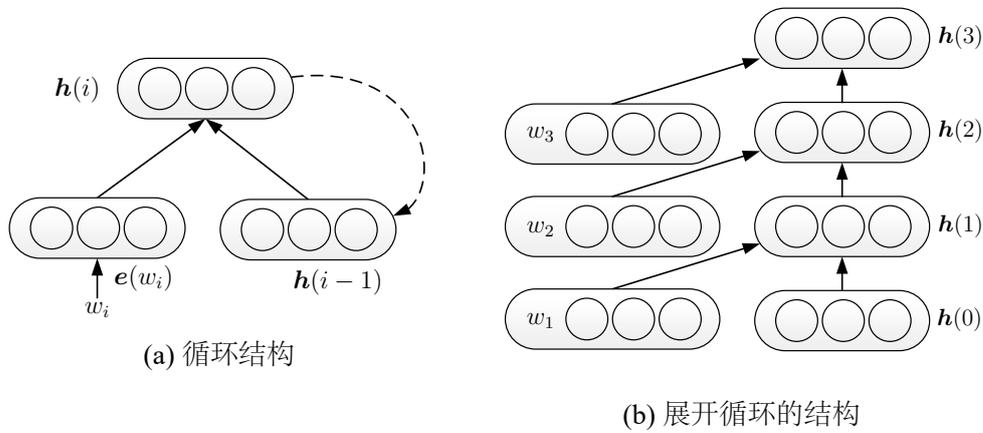

(a) 循环结构

(b) 展开循环的结构

图 5-2　循环神经网络模型结构图

一个非叶结点的右子树均为叶结点的树。这种特殊结构使得循环神经网络具有两个特点：一、由于固定了网络结构，模型只需在 $O(n)$ 时间内即可构建文本的语义。这使得循环神经网络可以更高效地对文本的语义进行建模。二、从网络结构上看，循环神经网络的层数非常深，句子中有几个词，网络就有几层。因此，使用传统方法训练循环神经网络时，会遇到梯度衰减或梯度爆炸的问题，这需要模型使用更特别的方法来实现优化过程 [8, 42]。

循环神经网络对文本语义的构建过程与 2.2.4 小节中介绍的循环神经网络语言模型类似。每个词与代表所有上文的隐藏层组合成新的隐藏层 (结构如图 5-2a，算法如式 5.5)，从文本的第一个词循环计算到最后一个词。当模型输入所有的词之后，最后一个词对应的隐藏层代表了整个文本的语义。

$$\boldsymbol{h}(i) = \phi\left(H\left[\boldsymbol{e}(w_i); \boldsymbol{h}(i-1)\right]\right) \tag{5.5}$$

优化方式上，循环神经网络与其它网络结构也略有差异。在普通的神经网络中，反向传播算法可以利用导数的链式法则直接推算得到。但是在循环神经网络中，由于其隐藏层到下一个隐藏层的权重矩阵 $H$ 是复用的，直接对权重矩阵求导非常困难。循环神经网络最朴素的优化方式为沿时间反向传播技术 (Back-propagation Through Time，BPTT)。该方法首先将网络展开成图 5-2b 的形式，对于每一个标注样本，模型通过普通网络的反向传播技术对隐藏层逐个更新，并反复更新其中的权重矩阵 $H$。由于梯度衰减的问题，使用 BPTT 优化循环神经网络时，只传播固定的层数 (比如五层)。为了解决梯度衰减问题，Hochreiter 和

Schmidhuber 在 1997 年提出了 LSTM（Long Short-Term Memory）模型 [43]。该模型引入了记忆单元，可以保存长距离信息，是循环神经网络的一种常用的优化方案。

无论采用哪种优化方式，循环神经网络的语义都会偏向文本中靠后的词。因此，循环神经网络很少直接用来表示整个文本的语义。但由于其能有效表示上下文信息，因此被广泛用于序列标注任务，如 2.2.4 小节中提到的神经网络语言模型。

### 5.2.4　卷积神经网络

卷积神经网络（Convolutional Neural Network）最早由 Fukushima 在 1980 年提出 [31]，此后，LeCun 等人对其做了重要改进 [61]。

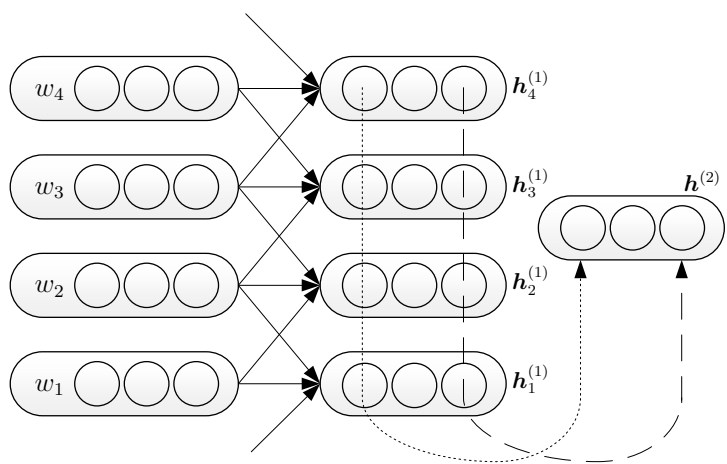

图 5-3　卷积神经网络模型结构图

卷积神经网络的结构如图 5-3，其核心是局部感知和权值共享。在一般的前馈神经网络中，隐藏层的每个节点都与输入层的各个节点有全连接；而在卷积神经网络中，隐藏层的每个节点只与输入层的一个固定大小的区域（$win$ 个词，对应图中 $win = 3$）有连接。从固定区域到隐藏层的这个子网络，对于输入层的所有区域是权值共享的。输入层到隐藏层的公式，形式化为：

$$\boldsymbol{x}_i = [\boldsymbol{e}(w_{i-\lfloor win/2 \rfloor}); \ldots ; \boldsymbol{e}(w_i); \ldots ; \boldsymbol{e}(w_{i+\lfloor win/2 \rfloor})] \tag{5.6}$$

$$\boldsymbol{h}_i^{(1)} = \tanh\left(W\boldsymbol{x}_i + \boldsymbol{b}\right) \tag{5.7}$$

在得到若干个隐藏层之后，卷积神经网络通常会采用池化技术，将不定长度的隐藏层压缩到固定长度的隐藏层中。常用的有均值池化和最大池化 [18]。最大池化的公式为：

$$\boldsymbol{h}^{(2)} = \max_{i=1}^{n} \boldsymbol{h}_i^{(1)} \tag{5.8}$$

卷积神经网络通过其卷积核，可以对文本中的每个部分的局部信息进行建模；通过其池化层，可以从各个局部信息中整合出全文语义，模型的整体复杂度为 $O(n)$。

卷积神经网络应用非常广泛。在自然语言领域，Collobert 等人首次将其用于处理语义角色标注任务，有效提升了系统的性能 [18]。2014 年，Kalchbrenner 等人与 Kim 分别发表了利用卷积神经网络做文本分类的论文 [47, 48]。Zeng 等人提出使用卷积神经网络做关系分类任务，取得了一定的成功 [127]。

### 5.2.5　文本分类

传统文本分类方法主要着眼于三个问题：特征表示、特征筛选和合适机器学习算法的选择。

特征表示方面，最常用的是词袋子特征，复杂一些的有词性标签、名词短语 [66] 以及树核 [94] 等特征。不同的特征从不同的角度对数据进行刻画，通常需要通过各种特征的组合，才能更好地描述文本。然而这些特征都存在数据稀疏问题。

特征选择希望通过删除噪声特征来提高文本分类的性能。最常用的方法是删除停用词（比如"的"），更高级的方法包括利用信息增益、互信息 [19] 等指标来筛选特征。近年来，Ng 等人提出的在优化目标中加入 $L_1$ 正则化 [85]，自动学习出稀疏特征，在文本分类的大规模应用中起到了重要作用。

在选择机器学习算法方面，几乎所有的分类器算法都在文本分类中有应用，如最近邻分类器，决策树分类器。在面对大规模文本分类任务时，高效的线性分类器应用的最为广泛，如 Logistic 回归（LR）、朴素贝叶斯（NB）和支持向量机（SVM）。

## 5.3　模型

本文提出一个深度模型来构建文本的语义，图 5-4 展示了该模型的网络结构。网络的输入是文档 $D$，由词序列 $w_1, w_2 \ldots w_n$ 组成。网络的输出节点数与

分类类别数相同。本文使用 $P(k|D,\theta)$ 来表示文档属于类别 $k$ 的概率，其中 $\theta$ 是网络的参数。

### 5.3.1　词表示

本文将词和其上下文组合在一起，用来表示一个词。上下文可以帮助词进行消歧，从而获得更精确的语义。本文使用一个双向循环结构来捕捉上下文。

定义 $\boldsymbol{c}_l(w_i)$ 为词 $w_i$ 的上文表示，$\boldsymbol{c}_r(w_i)$ 为词 $w_i$ 的下文表示。$\boldsymbol{c}_l(w_i)$ 和 $\boldsymbol{c}_r(w_i)$ 均为稠密实数向量，向量的维度是 $|\boldsymbol{c}|$。$\boldsymbol{c}_l(w_i)$ 的计算公式见式 5.9，其中 $\boldsymbol{e}(w_{i-1})$ 是词 $w_{i-1}$ 的词向量。词向量也是一个低维实数向量，维度为 $|\boldsymbol{e}|$。$\boldsymbol{c}_l(w_{i-1})$ 是词 $w_{i-1}$ 的上文表示。所有文档的第一个词的上文表示均使用同样的参数 $\boldsymbol{c}_l(w_1)$。$W^{(l)}$ 是一个矩阵，用于把表示上文的隐藏层转移到下一个词的上文表示中。$W^{(sl)}$ 也是矩阵，用于将当前词的语义合成到下一个词的上文表示中。$\phi$ 是非线性激活函数。$w_i$ 的下文表示 $\boldsymbol{c}_r(w_i)$ 也有相同的计算方式，具体如式 5.10。同样地，所有文档的最后一个词的下文表示也使用同样的参数 $\boldsymbol{c}_r(w_n)$。

$$c_l(w_i) = \phi(W^{(l)}c_l(w_{i-1}) + W^{(sl)}\boldsymbol{e}(w_{i-1})) \tag{5.9}$$

$$c_r(w_i) = \phi(W^{(r)}c_r(w_{i+1}) + W^{(sr)}\boldsymbol{e}(w_{i+1})) \tag{5.10}$$

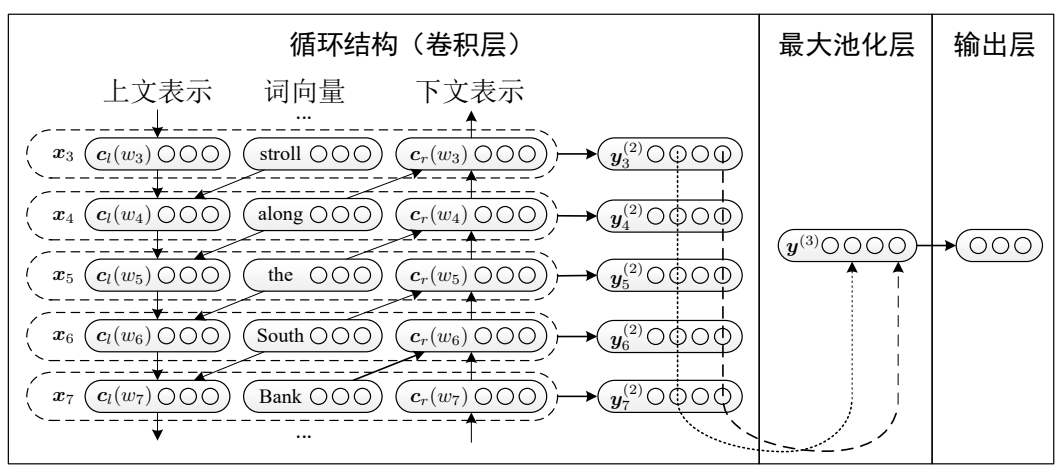

图 5-4　循环卷积网络模型结构图

图 5-4 为循环卷积网络的整体结构。图中展示了例句"A sunset stroll along the South Bank affords an array of stunning vantage points."的中间部分。其中各元

素的下标表示对应各词在句子中的位置，如"Bank"是句子中的第 7 个词，在图中记为 $w_7$。

根据公式 5.9、5.10 所示，上下文向量可以分别捕捉到上文和下文的语义信息。例如，在图 5-4 中，$c_l(w_7)$ 包含了词"Bank"的上文"A sunset stroll along the South"以及之前的所有词所包含的信息。而 $c_r(w_7)$ 则包含了"Bank"的下文"affords an array of stunning vantage points"所包含的信息。

在得到词 $w_i$ 的上下文信息之后，本文利用这些上下文信息对其进行消歧。本文将词 $w_i$ 的表示 $\boldsymbol{x}_i$ 定义为：词 $w_i$ 的上文表示 $c_l(w_i)$、下文表示 $c_r(w_i)$ 以及其词向量 $\boldsymbol{e}(w_i)$ 的拼接。

$$\boldsymbol{x}_i = [\boldsymbol{c}_l(w_i); \boldsymbol{e}(w_i); \boldsymbol{c}_r(w_i)] \tag{5.11}$$

使用循环结构，只需要对文本进行一次正向（从左往右）扫描，就可以获得所有的上文表示 $c_l$；同样地，只需要一次反向（从右往左）扫描，就可以获得所有的下文表示 $c_r$。因此，整个过程的复杂度是 $O(n)$。与此同时，使用这种循环结构捕获到的上下文信息，会比使用窗口表示的上下文有更好的效果。原因是，固定窗口的上下文表示方法只包含了部分上下文信息。

当得到词 $w_i$ 对应的表示 $\boldsymbol{x}_i$ 之后，本文对其作一个线性变换，再通过一个非线性的激活函数，将值送入神经网络的下一层。

$$\boldsymbol{y}_i^{(2)} = \tanh\left(W^{(2)}\boldsymbol{x}_i + \boldsymbol{b}^{(2)}\right) \tag{5.12}$$

其中 $\boldsymbol{y}_i^{(2)}$ 是隐含语义向量，在下一步中，本文会对其进行进一步分析，找出对文本分类最有用的隐含语义因子。

### 5.3.2  文本表示

本文使用卷积神经网络来构建整个文本的语义。从卷积网络的角度来看，上文介绍的循环结构可以看做卷积核。当得到所有的词表示之后，本文在此加入一个最大池化层（max-pooling layer）。

$$\boldsymbol{y}^{(3)} = \max_{i=1}^{n} \boldsymbol{y}_i^{(2)} \tag{5.13}$$

这里，max 操作是逐个元素计算的，也就是说，$\boldsymbol{y}^{(3)}$ 的第 $k$ 维就是各 $\boldsymbol{y}_i^{(2)}$ 向量的第 $k$ 维的最大值。借助最大池化层，可以将不同长度的文本转成固定长度的向量，从而表示整个文本。除了最大池化层，还有均值池化层 [18] 等其它池化技术。本文选用最大池化技术的主要出发点是，对于文本分类而言，最具决定性的词或者短语往往只有几处，而不是均匀散在文本各处。最大池化正好可以找出其中最有判别力的语言片段。同时，最大池化层的复杂度也是 $O(n)$。循环卷积网络中，循环结构和卷积结构是串联的，池化层使用的是循环结构的输出，因此整个模型的复杂度也是 $O(n)$。

模型的最后一部分是输出层，输出层定义如下。

$$\boldsymbol{y}^{(4)} = W^{(4)}\boldsymbol{y}^{(3)} + \boldsymbol{b}^{(4)} \tag{5.14}$$

最后，使用 softmax 函数将输出值转为概率值。

$$P(i|D,\theta) = \frac{\exp\left(\boldsymbol{y}_i^{(4)}\right)}{\sum_{k=1}^{n}\exp\left(\boldsymbol{y}_k^{(4)}\right)} \tag{5.15}$$

### 5.3.3 模型训练

**训练模型参数**

网络中的所有参数 $\theta$ 见表 5-1。

训练目标为最大化以下似然，其中 $\mathbb{D}$ 是训练文档集，$class_D$ 是文档 $D$ 的正确分类：

$$\theta \mapsto \sum_{D \in \mathbb{D}} \log P(class_D|D,\theta) \tag{5.16}$$

本文使用随机梯度下降法 [10] 来优化上述训练目标。每次迭代，随机选取一个样本 $(D, class_D)$，使用下式进行一次梯度迭代。式中，$\alpha$ 是学习速率。

$$\theta \leftarrow \theta + \alpha \frac{\partial \log P(class_D|D,\theta)}{\partial \theta} \tag{5.17}$$

对于循环结构的优化本文采用沿时间反向传播技术（backpropagation through time），该技术在本文第二章第 5.2.3 小节中介绍循环神经网络时已经提到，这

| 参数符号 | 维度 | 含义 |
|---|---|---|
| $e(\mathbb{V})$ | $\lvert e \rvert \times \lvert \mathbb{V} \rvert$ | 所有词向量 |
| $b^{(2)}$ | $H$ | 隐藏层偏置项 |
| $b^{(4)}$ | $O$ | 输出层偏置项 |
| $c_l(w_1)$ | $\lvert c \rvert$ | 初始上文参数 |
| $c_r(w_n)$ | $\lvert c \rvert$ | 初始下文参数 |
| $W^{(2)}$ | $H \times (\lvert e \rvert + 2\lvert c \rvert)$ | 隐藏层转移矩阵 |
| $W^{(4)}$ | $O \times H$ | 输出层转移矩阵 |
| $W^{(l)}$ | $\lvert c \rvert \times \lvert c \rvert$ | 上文语义转移矩阵 |
| $W^{(r)}$ | $\lvert c \rvert \times \lvert c \rvert$ | 下文语义转移矩阵 |
| $W^{(sl)}$ | $\lvert e \rvert \times \lvert c \rvert$ | 当前词融入上文语义转移矩阵 |
| $W^{(sr)}$ | $\lvert e \rvert \times \lvert c \rvert$ | 当前词融入下文语义转移矩阵 |

表 5-1　循环卷积网络参数列表

里不再赘述。

在训练中，本文参考了 Plaut 和 Hinton 的建议 [93]，使用了一个神经网络训练中常用的优化技巧。所有的参数在初始化时均使用均匀分布，其中随机数的最大绝对值为该元素入节点个数的平方根。入节点个数也就是神经网络中上一层的节点个数。对应的学习速率同样也除以入节点个数。

**词向量预训练**

词向量是一种词的分布表示，这种表示更适合作为神经网络的输入。最近的研究 [27, 40] 表明，如果选择一个好的初始值，神经网络可以收敛到更好的局部最优解。本章根据第三章中得到的经验，选取了适合本任务而且效率较高的 Skip-gram 模型，来优化词向量。Skip-gram 模型的介绍可见第二章第 2.2.6 小节。

## 5.4　实验设计

为了证明循环卷积网络的有效性，本文在四个不同种类的数据集上做对比实验。这些数据集包含了中文和英文两种语言，分类体系包括学科分类、文档作者母语分类和情感分类，文本长度包括句子级和篇章级。四个数据集的统计信息见表 5-2。

下文 5.4.1 小节详细描述了实验设置，包括数据预处理、参数设置等；5.4.2 小节描述了所有实验中均用到的对比模型。

| 数据集 | 类别数 | 训练/验证/测试集划分 | 平均文档长度 | 语言 |
|---|---|---|---|---|
| 20Newsgroups | 4 | 7520/836/5563 | 429 | 英文 |
| 复旦文本分类 | 20 | 8823/981/9832 | 2981 | 中文 |
| ACL 论文集 | 5 | 146257/28565/28157 | 25 | 英文 |
| 斯坦福情感树库 | 5 | 8544/1101/2210 | 19 | 英文 |

表 5-2 文本分类数据集概要信息

### 5.4.1 实验设置

对于英文数据集，本文使用 Stanford Tokenizer[1]对语料进行分词。对于中文数据集，使用 ICTCLAS[2]分词。文本中的停用词和特殊符号均当作普通单词保留。这四个数据集均已分好训练集和测试集，其中 ACL 论文集和斯坦福情感树库已经分好了训练、验证、测试集。对于另两个数据集，本文随机选取训练集的 10% 作为验证集，剩下部分作为真实的训练集。

网络超参数的选择一般需要根据数据的不同而有所调节。在实验中，本文参考了 Collobert 等人 [18] 和 Turian 等人 [116] 的工作，只使用了一组最常用的超参数。具体来说，学习速率 $\alpha = 0.01$，隐藏层大小 $H = 100$，词向量维度 $|e| = 50$，上下文向量维度 $|c| = 50$。中英文的词向量使用中英文维基百科进行训练，训练工具包为 word2vec[3]。公式 5.9 和公式 5.10 中的激活函数 $\phi$，本文使用双曲正切函数 tanh。如果使用更适合深层网络的 ReLU 激活函数 [32] 能对系统性能有小幅度的提升，但由于对比方法同样使用 tanh 作为激活函数，本实验中依然选择使用 tanh。

### 5.4.2 对比方法

#### 词袋子、二元词组 + Logistic 回归、支持向量机

Wang 和 Manning 在 2012 年发表的文章 [120] 中展示了若干种实现简便的文本分类基准实验。这些实验中主要使用词和二元词组作为特征，并用常规分类器算法进行分类。本文在实验中选择了词以及二元词组作为特征，特征的权重为各词在文档中的词频。同时，本文选择 Logistic 回归和支持向量机作为分

---

[1] http://nlp.stanford.edu/software/tokenizer.shtml
[2] http://ictclas.nlpir.org
[3] http://code.google.com/p/word2vec

类器，具体使用了 LIBLINEAR 工具包[4]，并使用其默认参数作为分类器的设置。

**平均词向量 + Logistic 回归**

该方法使用文档中各词词向量的加权平均值作为文档的表示，然后使用 Logistic 回归分类器对其进行分类。其中各个词的权重为词在文档中的词频。该方法可能是利用词向量表示文本的一种最简单的方法，通常基于神经网络的分类模型均会选用该方法作对比 [58, 110]。

作为一种文本表示，平均词向量除了用在单语文本分类中，还出现在其它场景下，如 Huang 等人在其模型中，使用平均词向量作为文档的全局表示 [44]，辅助提升词向量的语义。Klementiev 等人也使用平均词向量作为文本表示，并基于此进行跨语言的文本分类 [50]。

**卷积神经网络**

Collobert 等人在 2011 年将卷积神经网络引入自然语言处理中。尽管他们使用卷积神经网络做的是语义角色标注任务，但是该模型可以非常容易地移植到文本分类任务中。只需要将其输出层修改为和循环卷积网络中一样的分类器即可。卷积神经网络使用固定的窗口作为上下文表示，对比循环卷积网络，其卷积核是窗口中若干词的拼接，也就是将循环卷积网络模型中对词 $w_i$ 的表示 $\boldsymbol{x}_i$ 替换为：

$$\boldsymbol{x}_i = [\boldsymbol{e}(w_{i-\lfloor win/2 \rfloor}); \ldots; \boldsymbol{e}(w_i); \ldots; \boldsymbol{e}(w_{i+\lfloor win/2 \rfloor})]$$

与本文同期，也有若干其它工作使用卷积神经网络直接做文本分类任务，大体思想与上述模型类似，如文献 [47, 48, 59]。

**其他对比方法**

会在具体实验中介绍。

## 5.5   实验及分析

本节前四小节详细介绍了各数据集上的实验结果，并对各个实验进行深入分析。最后 5.5.5 小节总结了所有的实验。

---

[4]http://www.csie.ntu.edu.tw/~cjlin/liblinear/

### 5.5.1  20Newsgroups[5]

20Newsgroups 数据集包括了 20 个新闻组的共约 2 万封邮件，是目前使用最广泛的英文文本分类数据集之一。该数据集有若干不同的版本，本文选用了 bydate 版本，因为该版本已经将数据集分成了训练集和测试集，方便和现有工作做对比。

本文按照 Hingmire 等人的工作 [38] 将数据集划分成 comp、politics、rec 和 religion 一共四个大类，各类别分别包括：

- comp：comp.os.ms-windows.misc、comp.graphics、comp.sys.mac.hardware、comp.windows.x、comp.sys.ibm.pc.hardware

- politics：talk.politics.misc、talk.politics.guns、talk.politics.mideast

- rec：rec.autos、rec.motorcycles、rec.sport.baseball、rec.sport.hockey

- religion：talk.religion.misc、alt.atheism、soc.religion.christian

20Newsgroups 数据集是一个类别不均衡的文本分类数据集，即使按照上述方法分成四个大类，各类别样本数仍然大约呈 6:3:5:3 的比例。

该数据集在这种划分下的最佳结果由文献 [38] 所得，文中提出了一种结合了 LDA、EM 算法和朴素贝叶斯分类器的文本分类模型，并命名为 ClassifyLDA-EM。本文按照这项工作的设置，使用宏平均 F1 值（Macro F-measure）评价各模型在 20Newsgroups 数据集上的表现。宏平均 F1 值定义为各分类 F1 值的宏平均数。

表 5-3 展示了各模型在 20Newsgroups 数据集上的表现。其中表格最上面一部分区域表示传统文本分类算法的效果，中间部分为此前该数据集上最好的模型 ClassifyLDA-EM 的表现，下面部分分别为本文实现的卷积神经网络和循环卷积网络的性能。根据实验结果，可以得到下述结论：

一、无论是卷积神经网络还是循环卷积网络，它们都比传统基于分类器的方法要有明显的优势。基于词特征及二元词组特征的传统分类方法，最多只有 93.12% 的性能，相比而言，循环卷积网络达到了 96.49% 的性能，分类误差减

---

| 模型 | 宏平均 F1 |
|------|-----------|
| 词袋子 + Logistic 回归 | 92.81 |
| 二元词组 + Logistic 回归 | 93.12 |
| 词袋子 + 支持向量机 | 92.43 |
| 二元词组 + 支持向量机 | 92.32 |
| 平均词向量 + Logistic 回归 | 89.39 |
| ClassifyLDA-EM [38] | 93.60 |
| 卷积神经网络 | 94.79 |
| 循环卷积网络 | **96.49** |

表 5-3    各模型在 20Newsgroups 数据集上的表现

少了 49%。此前最好的方法 ClassifyLDA-EM 也只达到了 93.6% 的性能，循环卷积网络在其基础上降低了 45% 的分类错误。

二、对比卷积神经网络和循环卷积网络，可以发现循环卷积网络在文本分类性能上有明显的优势。循环卷积网络相对卷积神经网络的分类误差，减少了 33%。对此，本文设计了一个实验，更仔细地对比分析这两个模型。

**对比循环卷积网络和卷积神经网络**

卷积神经网络和循环卷积网络最主要的不同点在于对上下文的表示。卷积网络使用一个固定大小的窗口来表示上下文信息，而循环卷积网络使用循环结构来构建任意距离的上下文信息。卷积网络的性能受到窗口大小的影响，窗口太小会丢失一些长距离的关系；窗口太大会引入数据稀疏问题，而且参数过多会使得模型更难训练。

为了更好地分析卷积神经网络的性能，本文尝试了从 1 到 19 之间的所有奇数大小的窗口，作为卷积网络中上下文信息的数据源。比如说，窗口为 1 时，$x_i =$ $[e(w_i)]$，词的表示就是词向量。当窗口为 3 时，$x_i = [e(w_{i-1}); e(w_i); e(w_{i+1})]$，也就是说，词表示为当前词、前一个词以及下一个词的词向量的组合。

图 5-5 展示了在 20Newsgroups 数据集上，卷积神经网络在不同大小的窗口下，在测试集上取得的评估结果。图中的横坐标表示卷积神经网络选择的不同窗口大小，纵坐标为模型在测试集上的性能，以宏平均 F1 值为指标。可以看到，卷积网络 (蓝线) 和设想的一样，随着窗口的变大，测试效果先变好，再变差。当窗口大小取 11 个词时 (前五个词，后五个词)，模型的性能达到最佳状态。而

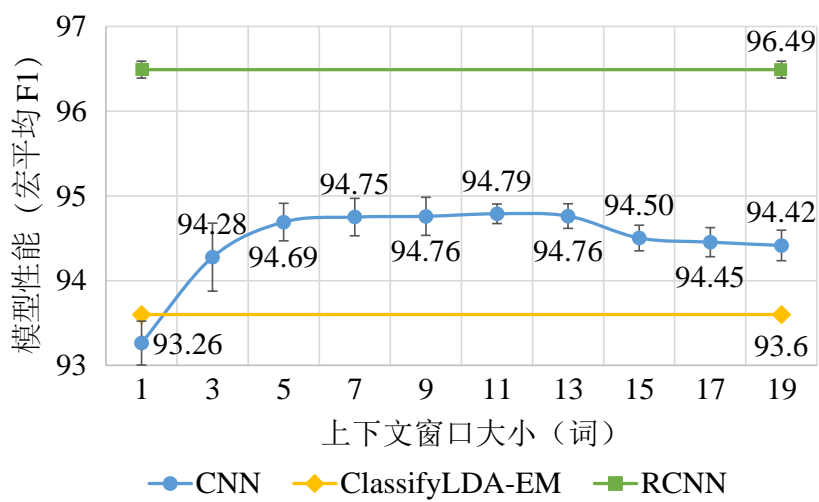

图 5-5 窗口大小对卷积神经网络等模型的性能影响

循环卷积网络（绿线）的性能不依赖于窗口大小，比卷积神经网络选最佳窗口大小时的效果更好。

由于在实验中，本文使用完全相同的参数和实现方法，卷积神经网络与循环卷积网络的仅有差别在于对上下文的表示，因此本文认为循环卷积网络对于卷积神经网络的优势来自于循环卷积网络中的循环结构可以保留更长的上下文信息；同时相比大窗口的卷积神经网络，循环卷积网络也会引入更少的噪声。

### 5.5.2 复旦文本分类[6]

复旦文本分类数据集由复旦大学李荣陆 [134] 提供。该数据集是一个中文文本分类语料，其中包括艺术、教育、能源等一共 20 个类别。复旦文本分类数据集同样也是一个类别分布不均衡的数据集，其中较大的类别，如经济类，有 3200 篇文档；而较小的类别，如文学类，只有 70 篇文档。数据集中训练样本与测试样本大致呈 1:1 的比例。

该数据集上此前的最佳结果由田文波等人的 Labeled-LDA 模型 [133] 所得。

表 5-4 展示了各模型在复旦文本分类数据集上的表现。与 20Newsgroups 的实验结果类似，循环卷积网络和卷积神经网络比其它对比模型有明显的优势。同时，在该数据集上循环卷积网络比卷积神经网络的分类误差少 19%。

| 模型 | 准确率 |
|---|---|
| 词袋子 + Logistic 回归 | 92.08 |
| 二元词组 + Logistic 回归 | 92.97 |
| 词袋子 + 支持向量机 | 93.02 |
| 二元词组 + 支持向量机 | 93.03 |
| 平均词向量 + Logistic 回归 | 86.89 |
| Labeled-LDA [133] | 90.80 |
| 卷积神经网络 | 94.04 |
| 循环卷积网络 | **95.20** |

表 5-4　各模型在复旦文本分类数据集上的表现

通过这个实验，可以发现循环卷积网络对于中文的文本表示依然有效。对于传统基于特征的文本分类方法而言，特征抽取是至关重要的一个步骤。相比而言，英文有较多较好的文本特征抽取工具，而中文的文本特征抽取目前尚无法达到英文的水准，比如英文句法分析工具的性能一般比中文句法分析工具的性能好 10 个百分点。

本文提出的循环卷积网络不依赖于除了分词工具外的自然语言处理工具，直接从词出发，构建文本的语义表示，并且分类效果较传统方法有较大的优势。因此利用循环卷积网络做文本表示以及完成文本分类任务，对于处理缺乏自然语言处理工具的语种，是一种值得考虑的方法。

### 5.5.3　ACL 论文集[7]

ACL 论文集（ACL Anthology Network）[96] 包含了国际计算语言学协会（ACL，The Association for Computational Linguistics）旗下的多个会议和期刊的论文。Post 和 Bergsma 对论文集中从 2001 年到 2009 年的论文进行了作者母语标注。标注过程主要依据作者的国籍，标注了五种最常见的作者母语：英语、日语、德语、汉语和法语 [94]。为了保证数据集的有效性，Post 等人抛弃了模棱两可的文章，只保留有把握的部分，一共对于 8483 篇论文中的 1959 篇进行了标注。

为了充分调研模型在不同长度文本下的性能，本文按照文献 [94] 中的实验设置，对各篇论文中的每个句子进行分类。在这个分类数据集上，目前最佳的

---

[7]http://old-site.clsp.jhu.edu/~sbergsma/Stylo/

结果为 Post 等人 [94] 所得，他们在实验中对比分析了若干种不同的树核（Tree kernel）特征。本文在对比实验中列举了其中最有代表性的两种特征，分别为：一、使用 Berkeley parser [92] 抽取得到的深度为 1 的上下文无关文法（CFG）的个数，作为特征集合；二、由 Charniak 和 Johnson 提出的，对上述特征重排序得到的特征集合（C&J）[14]。

| 模型 | 准确率 |
| --- | --- |
| 词袋子 + Logistic 回归 | 46.67 |
| 二元词组 + Logistic 回归 | 47.00 |
| 词袋子 + 支持向量机 | 45.24 |
| 二元词组 + 支持向量机 | 46.14 |
| 平均词向量 + Logistic 回归 | 41.32 |
| CFG [94] | 39.20 |
| C&J [94] | **49.20** |
| 卷积神经网络 | 47.47 |
| 循环卷积网络 | **49.19** |

表 5-5　各模型在 ACL 论文集数据集上的表现

　　表 5-5 展示了各模型在 ACL 论文集数据集上的表现。与前两个数据集类似，卷积神经网络和循环卷积网络依然有较好的效果。

　　在该数据集上，两个基于树核特征的方法，分别获得了 39.2% 和 49.2% 的性能。由此可见，人工设计特征达到的性能，对特征的依赖程度非常大。对比循环卷积网络和这两个人工精心设计的特征，实验表明本文的方法比直接使用 CFG 特征有显著的优势，并且和 C&J 特征有相似的效果。本文认为循环卷积网络有这样的效果是因为它能捕获长距离的上下文信息。尽管树核的特征也能捕获长距离的句法依赖关系，但是循环卷积网络不需要人工设计特征就能达到相似的效果。对于一些缺乏句法分析工具的语言，循环卷积网络会有较大的优势。

### 5.5.4　斯坦福情感树库[8]

　　斯坦福情感树库（Stanford Sentiment Treebank）由 Socher 等人标注并发布。2005 年，Pang 和 Lee 抓取了 IMDB 网站[9]上的若干条电影评论数据，得到了电

---

[8] http://nlp.stanford.edu/sentiment/

[9] http://www.imdb.com/

影评论数据集 [87]。2013 年，Socher 等人 [110] 在电影评论数据集的基础上，使用 Stanford Parser [49] 对其中的每个句子进行句法分析，并对得到的每个句法子树所对应的短语进行情感倾向打分，每个短语由三位标注者打分，并取其平均值作为标注的短语情感。

该数据集一共包括 11855 条电影评论，和此前最好的工作 [110] 一致，本文也使用五分类（非常正面、正面、中立、负面、非常负面）标签体系。在文献 [110] 中，Socher 等人利用了对每条评论中所有短语的标注信息，训练了递归神经网络。然而在训练循环卷积网络时，只使用了对于整个句子的情感倾向标注，而没有使用其中对于短语和子句的标注。

与本文同期，另有一项基于卷积网络的文本分类工作发表 [48]，下文会针对这些方法进行详细的比较。

| 模型 | 准确率 |
|------|--------|
| 词袋子 + Logistic 回归 | 40.86 |
| 二元词组 + Logistic 回归 | 36.24 |
| 词袋子 + 支持向量机 | 40.70 |
| 二元词组 + 支持向量机 | 36.61 |
| 平均词向量 + Logistic 回归 | 32.70 |
| 递归神经网络 [109] | 43.20 |
| 递归张量网络 [110] | 45.70 |
| 卷积神经网络（Kim）[48] | **48.00** |
| 卷积神经网络 | 46.35 |
| 循环卷积网络 | 47.21 |

表 5-6　各模型在斯坦福情感树库数据集上的表现

表 5-6 展示了各模型在斯坦福情感树库数据集上的表现。与前三个数据集类似，卷积神经网络和循环卷积网络依然有较好的效果。

实验结果中，文献 [48] 与本文实现的卷积神经网络在模型结构上完全一致。这两个同样的模型得到不同的结果主要是因为文献 [48] 用了较多的神经网络优化技巧，主要包括 Dropout[41, 121]、AdaDelta[126] 等先进的归一化方法和优化方法。这些最新提出的优化方法对所有的神经网络模型都有效果。但是由于本文的主要对比模型递归神经网络 [109, 110] 并没有使用这些先进的优化方法，为了更明确地对比模型结构带来的性能提升，本文实现的模型中也没有使用文献

[48] 所用的优化方法。

在这个实验中，可以发现卷积神经网络或者循环卷积网络在句子级分类任务上，相对递归神经网络仍然有较大的提升。基于卷积的网络在表示文本语义上相对此前基于递归结构的神经网络模型具有更大的优势。本文认为这是因为卷积神经网络可以更好地捕获上下文信息，并从中选择出最重要的特征；而递归网络需要按照树形结构构建文本语义，这依赖于所构建的树的精度。另一方面，递归网络的时间复杂度是 $O(n^2)$，而循环卷积网络只需要 $O(n)$ 的时间复杂度。在实践中，递归张量网络按照文献 [110] 中的报告，需要 3 到 5 小时的训练时间。而循环卷积网络同样在斯坦福情感树库数据集上用单线程训练，只需要若干分钟。

**关键短语**

| | 正面情感 | 负面情感 |
|---|---|---|
| 循环卷积网络 | well *worth* the<br>**a *wonderful*** **movie**<br>even *stinging* at<br>and *invigorating* film<br>and *ingenious* entertainment<br>and *enjoy* .<br>'s *sweetest* movie | A *dreadful* live-action<br>Extremely *boring* .<br>Extremely *dumb* .<br>an *awfully* derivative<br>'s *weaker* than<br>incredibly *dull* .<br>**very *bad* sign** |
| 递归张量网络 | an amazing performance<br>most visually stunning<br>wonderful all-ages triumph<br>**a wonderful movie** | for worst movie<br>A lousy movie<br>a complete failure<br>**very bad sign** |

表 5-7 循环卷积网络与递归张量网络抽取的正负情感关键短语

为了直观地观察循环卷积网络与递归张量网络在文档构建时的表现，本文在此列举了这两个模型抽取得到的若干"关键短语"。在循环卷积网络中，关键短语是指模型在最大池化层提取出来次数最多的短语片段。由于在循环卷积网络中，每个词实际上是带有整个文档的语义的，因此本文只列举了中心词附近的多词短语。在递归张量网络中 [110]，关键的短语为情感最强烈的短语。

表 5-7 列举了两个模型在训练后，从测试集中抽取的关键短语，这里仅列举三元短语。其中递归张量网络抽取得到的关键短语直接引自文献 [110]。

从表中可以发现，循环卷积网络与递归张量网络会同时将一些短语判断为关键短语，如"a wonderful movie"、"very bad sign"。而对于大多数关键短语而言，这两个模型抽取的结果是不同的。递归张量网络由于使用了句法分析器，对句法树中的每一个节点进行标注，因此得到的关键短语均为句法树上的某个子树。而循环卷积网络不依赖于句法分析器，得到的短语并不是普通句法意义上的短语，而是一个中心词与其上下文的组合。

根据循环卷积网络的特点，在分析其抽取得到的关键短语时，应该更多地关注其中心词，并且以上下文作为参考，确保其意义正确。从表中观察，对于正面情感最重要的词是"worth"、"sweetest"、"wonderful"之类，而对于负面情感最重要的词是"awfully"、"bad"、"boring"等，这与人的直观感受一致。因此本文认为循环卷积网络不仅能有效地对文本进行建模，而且在建模过程中的步骤也是有意义的。

### 5.5.5　实验总结

根据上述四个数据集上的结果，本文主要得出以下结论：

- 对比神经网络方法（卷积、递归、循环卷积）和传统文本分类算法（词袋子 +LR），实验结果表明神经网络方法在四个数据集上均有优势。这说明神经网络可以有效地构建文本语义。相比传统词袋子的方法，神经网络模型可以更有效地对上下文进行建模，保留词序信息，同时减少了数据稀疏问题。

- 在斯坦福情感树库数据集上，本文对比了卷积网络、循环卷积网络和递归网络，并发现基于卷积的网络（卷积网络与循环卷积网络）有更好的分类性能。本文认为卷积网络在表示文本语义上相对此前的递归神经网络模型具有更大的优势。基于卷积的网络相对递归网络的主要优势在于：一、递归神经网络需要按照树形结构构建文本语义，这依赖于所构建的树的精度；而卷积网络直接从纯文本中对上下文建模，并从中选择出最重要的特征，对上下文的建模更有效，同时也不依赖于其它的自然语言处理工具，避免了误差传递。二、卷积网络可以在线性时间内得到文本的表示，而递归网络需要句子长度的平方级时间，对于长句子而言，卷积网络会有更高的效率。

- 在除了 ACL 论文集以外的数据集中，循环卷积网络都达到了目前最好的成绩。在 ACL 论文集数据集中，循环卷积网络也取得了和此前最好几乎一样的成绩。在 20Newsgroups 和复旦文本分类数据集中，本文分别将错误率降低了 33% 和 19%。这说明循环卷积网络能有效地表示文本，并且能在文本分类任务中体现出其优势。

- 在 ACL 数据集上，本文对比循环卷积网络和人工精心设计的特征。实验表明循环卷积网络比直接使用 CFG 特征有显著的优势，并且和 C&J 特征有相似的效果。本文认为循环卷积网络有这样的效果是因为它能捕获长距离的上下文信息。尽管树核的特征也能捕获长距离的句法依赖关系，但是循环卷积网络不需要人工设计特征就能达到相似的效果。对于一些缺乏句法分析工具的语言，循环卷积网络会有较大的优势。

- 对比卷积网络和循环卷积网络，本文发现循环卷积网络在每个数据集上都比基于窗口的卷积网络的效果要好。这说明循环结构相比窗口模式，可以更有效地对上下文进行建模。

纵观四个数据集上的效果，本文认为利用循环卷积网络做文本表示，以及以此为基础进行文本分类，是一种有效的方法。该方法不依赖于现有的自然语言处理工具，对文本的语言、分类体系不敏感，并且相比人工精心设计的特征也有非常大的竞争力。

## 5.6　本章小结

本章提出使用循环卷积网络做文本表示，并用其完成了文本分类任务。该模型通过循环结构捕获上下文信息，并通过卷积网络构建整个文本的表示。相比现有的递归神经网络，循环卷积网络不依赖于现有的自然语言处理工具，并且在更小的时间复杂度内对文本进行建模。相比现有的卷积神经网络，循环卷积网络可以捕获长距离的依赖关系，对上下文更有效地建模。实验结果表明循环卷积网络在多个数据集上均取得了最优的成绩。

# 第六章　总结与展望

文本表示是自然语言处理的基础工作，具有重要的理论意义和广阔的应用前景。本文对基于神经网络的文本表示技术中，最重要的两个问题，词和文档的表示，进行深入分析，比较现有方法的优劣，并提出自己的方法。

本文的主要工作和贡献总结如下：

在基于神经网络的词向量表示方法中，本文对现有的词向量表示技术进行了系统的理论对比及实验分析。理论方面，本文阐述了各种现有模型的联系，从上下文的表示、上下文与目标词之间的关系两方面对模型进行了分类整理，并证明了其中最重要的两个模型 Skip-gram 与 GloVe 之间的关系。实验方面，本文从模型、语料和训练参数三个角度分析训练词向量的关键技术。本文选取了三大类一共八个指标对词向量进行评价，这三大类指标涵盖了现有的词向量用法。通过实验的比较，本文发现，训练词向量首先需要选择一个与任务相匹配语料，选择领域内的语料训练词向量，能对系统性能有巨大的提升。对于确定类型的语料，语料规模越大，词向量的性能越好。其次，训练词向量需要选择一个合适的模型，当语料较小时，应当选用模型结构最简单的 Skip-gram 模型，而当语料较大时，选用 CBOW 或者本文提出的 Order 模型会有更好的效果。最后，在训练词向量模型时，需要利用任务验证集或类似任务，找到词向量训练的最佳迭代次数。

在中文表示技术中，本文提出了基于字词联合训练的中文表示方法。现有的中文表示技术往往沿用了英文的思路，直接从词的层面对文本表示进行构建。本文根据中文的特点，提出了基于字词联合训练的表示技术。该方法在字表示的空间中融入了上下文的词，利用词的语义空间，更好地对汉字建模；同时在词表示中借助字的平滑作用，更好地对词进行建模。对于字表示，本文在分词任务上对比了字词联合训练与单独训练的字表示，实验表明字词联合训练得到的字向量，在中文分词任务上有显著的优势；对于词表示，本文采用词义相似度任务以及文本分类任务对比字词联合训练与单独训练的词表示，实验结果同样展示了字词联合训练的有效性。

在基于神经网络的文档表示技术中，本文分析了现有的文档表示技术：基于循环网络的表示技术、基于递归网络的表示技术和基于卷积网络的表示技术。

并针对现有的三种表示技术的不足，提出了基于卷积循环网络的文档表示技术。该方法克服了此前递归网络的复杂度过高的问题，循环网络的语义偏置问题，以及卷积网络窗口较难选择的问题。本文在文本分类任务上对新提出的表示技术进行了对比分析，实验表明基于循环卷积网络的文本表示技术比现有的表示技术在文本分类任务上，有更好的性能。

在机器学习领域有一个公认的观点是，模型选用的特征决定了机器学习算法所能达到的上界；而具体模型的选择，则决定了对上界的逼近程度。因此，特征的表示在机器学习中是至关重要的一个步骤。在基于神经网络的词向量表示技术中，尽管不同的模型有着不同的性能，但这些模型均基于分布假说，语义由其上下文决定。这些模型所能表达出的语义，受到语料中各词上下文分布的约束。

最近一两年，已经有人尝试跳出分布假说的框架，使用更广泛的信息，对语义进行建模。Weston 等人设计了一个模型，从知识库中获取语义表示，并用于提升关系抽取任务的性能 [123]。Wang 等人指出，利用知识库中的知识，可以进行词义消歧，进一步提升词向量的性能 [122]。

未来的工作需要考虑，如何利用海量的多源异构的数据，从中挖掘出有用的信息，更好地对数据和知识进行表示。

## 附录 A  Skip-gram 模型与"词-词"矩阵分解模型关系的证明

基于"词-词"矩阵分解的模型与神经网络词向量模型中最简单的 Skip-gram 模型有同样的最优解。下面给出简要证明。

首先列出第二章中 Skip-gram 模型的整体优化目标,公式 2.22:

$$\sum_{(w_i,c)\in\mathbb{D}}\sum_{w_j\in c}\log P(w_i|w_j) \tag{2.22}$$

该式的两层求和分别为遍历语料中的每一个词,以及遍历每个词的上下文。而每次参与求和的元素 $\log P(w_i|w_j)$,只与 $(w_i, w_j)$ 这一组词对有关。词对中的两个词均来自词表,因此对于一个固定的词表来说,不同类型的求和项最多只有 $|\mathbb{V}|^2$ 个。因此,原式通过调整求和顺序,合并相同的 $(w_i, w_j)$ 词对,等价于:

$$\sum_{v_i\in\mathbb{V}}\sum_{w_j\in\mathbb{V}}x_{ij}\log P(v_i|v_j) \tag{A.1}$$

其中,$\mathbb{V}$ 为词表,$v_i$ 和 $v_j$ 均为词表中的某个词,$x_{ij}$ 表示词对 $(v_i, v_j)$ 在公式 2.22 中,各个求和项里的出现次数。

根据第二章中 Skip-gram 模型的条件概率公式 2.23:

$$P(v_i|v_j) = \frac{\exp\left(\boldsymbol{e}'(v_i)^{\mathsf{T}}\boldsymbol{e}(v_j)\right)}{\sum_{v_k\in\mathbb{V}}\exp\left(\boldsymbol{e}'(v_k)^{\mathsf{T}}\boldsymbol{e}(v_j)\right)} \tag{2.23}$$

将其代入公式 A.1,得到:

$$\sum_{v_i\in\mathbb{V}}\sum_{v_j\in\mathbb{V}}x_{ij}\log\frac{\exp\left(\boldsymbol{e}'(v_i)^{\mathsf{T}}\boldsymbol{e}(v_j)\right)}{\sum_{v_k\in\mathbb{V}}\exp\left(\boldsymbol{e}'(v_k)^{\mathsf{T}}\boldsymbol{e}(v_j)\right)} \tag{A.2}$$

为了方便记述,这里将公式 A.2 中的符号做一定的简化。令 $\boldsymbol{p}_k = \boldsymbol{e}'(w_k)$,$\boldsymbol{q}_k =$

$\boldsymbol{e}(w_k)$，则公式 A.2 简化为：

$$\sum_{i=1}^{|\mathbb{V}|} \sum_{j=1}^{|\mathbb{V}|} x_{ij} \log \frac{\exp\left(\boldsymbol{p}_i^\mathsf{T} \boldsymbol{q}_j\right)}{\sum_{k=1}^{|\mathbb{V}|} \exp\left(\boldsymbol{p}_k^\mathsf{T} \boldsymbol{q}_j\right)} \tag{A.3}$$

假设向量 $\boldsymbol{p}$ 和 $\boldsymbol{q}$ 的维度足够大，则 $\boldsymbol{p}_i^\mathsf{T} \boldsymbol{q}_j$ 可以拟合任意矩阵。记：

$$\begin{aligned} a_{ij} &= \boldsymbol{p}_i^\mathsf{T} \boldsymbol{q}_j \\ b_{ij} &= \frac{\exp(a_{ij})}{\sum_{k=1}^{|\mathbb{V}|} \exp(a_{kj})} \end{aligned} \tag{A.4}$$

则根据 softmax 函数的性质，对于某个固定的 $j$ 存在约束：

$$\sum_{k=1}^{|\mathbb{V}|} b_{kj} = 1 \tag{A.5}$$

将公式 A.4 代入公式 A.3 的训练目标中，并且交换两个求和的顺序，可得，Skip-gram 的目标为最大化：

$$\sum_{j=1}^{|\mathbb{V}|} \sum_{i=1}^{|\mathbb{V}|} x_{ij} \log(b_{ij}) \tag{A.6}$$

由于存在约束 A.5，在对 $x_{ij}$ 做归一化之后，公式 A.6 第二个求和符号部分 $\sum_{i=1}^{|\mathbb{V}|} x_{ij} \log(b_{ij})$ 可以看做交叉熵，根据交叉熵的性质，公式 A.6 的最优解为：

$$b_{ij} = \frac{x_{ij}}{\sum_{k=1}^{|\mathbb{V}|} x_{kj}} \tag{A.7}$$

根据公式 A.4 中的定义，可得：

$$\frac{\exp(a_{ij})}{\sum_{k=1}^{|\mathbb{V}|} \exp(a_{kj})} = \frac{x_{ij}}{\sum_{k=1}^{|\mathbb{V}|} x_{kj}} \tag{A.8}$$

其中，如果 $a_{ij} = \log(x_{ij}) + t_j$，其中 $t_j$ 为只与 $j$ 有关的变量，则公式 A.8 成立。

　　因此，当且仅当 $\boldsymbol{p}_i^\mathsf{T} \boldsymbol{q}_j = \log(x_{ij})$ 可以满足时，最大化公式 2.22（求解 Skip-gram 模型）与最小化以下二式，具有相同的最优解：

$$\sum_{j=1}^{|\mathbb{V}|} \sum_{i=1}^{|\mathbb{V}|} \left( \log(x_{ij}) - \boldsymbol{p}_i^{\mathrm{T}} \boldsymbol{q}_j \right)^2 \tag{A.9}$$

$$\sum_{j=1}^{|\mathbb{V}|} \sum_{i=1}^{|\mathbb{V}|} \left( \log \frac{x_{ij}}{\sum_{k=1}^{|\mathbb{V}|} x_{kj}} - \boldsymbol{p}_i^{\mathrm{T}} \boldsymbol{q}_j \right)^2 \tag{A.10}$$

由于上述各式中 $x_{ij}$ 的定义与 GloVe 中 $x_{ij}$ 的计算方法相同，因此在最优解可以取到的情况下，Skip-gram 模型与"词-词"矩阵分解的模型具有相同的最优解。其中公式 A.9 是对 log 共现次数进行分解，与 GloVe 的优化目标（公式 2.1）一致；而公式 A.10 是对平移后的 PMI（Pointwise Mutual Information）共现矩阵进行分解。

# 参考文献

[1] 汉语信息处理词汇 *01 部分: 基本术语（GB12200.1-90）6*. 中国标准出版社, 1991.

[2] Charu C Aggarwal and ChengXiang Zhai. A survey of text classification algorithms. In *Mining text data*, pages 163–222. Springer, 2012.

[3] Marco Baroni, Georgiana Dinu, and Germán Kruszewski. Don't count, predict! a systematic comparison of context-counting vs. context-predicting semantic vectors. In *Proceedings of the 52nd Annual Meeting of the Association for Computational Linguistics (ACL)*, volume 1, pages 238–247, 2014.

[4] Robert M. Bell and Yehuda Koren. Lessons from the netflix prize challenge. *SIGKDD Explor. Newsl.*, 9:75–79, 2007.

[5] Yoshua Bengio, Aaron Courville, and Pascal Vincent. Representation learning: A review and new perspectives. *IEEE TPAMI*, 35(8):1798–1828, 2013.

[6] Yoshua Bengio, Réjean Ducharme, and Pascal Vincent. A neural probabilistic language model. In *Advances in Neural Information Processing Systems*, pages 932–938, 2001.

[7] Yoshua Bengio, Réjean Ducharme, Pascal Vincent, and Christian Jauvin. A Neural Probabilistic Language Model. *The Journal of Machine Learning Research*, 3:1137–1155, 2003.

[8] Yoshua Bengio, Patrice Simard, and Paolo Frasconi. Learning long-term dependencies with gradient descent is difficult. *Neural Networks, IEEE Transactions on*, 5(2):157–166, 1994.

[9] Antoine Bordes, Xavier Glorot, Jason Weston, and Yoshua Bengio. Joint learning of words and meaning representations for open-text semantic parsing. In

*International Conference on Artificial Intelligence and Statistics*, pages 127–135, 2012.

[10] Léon Bottou. Stochastic gradient learning in neural networks. In *Proceedings of Neuro-Nımes*, volume 91, 1991.

[11] John S Bridle. Probabilistic interpretation of feedforward classification network outputs, with relationships to statistical pattern recognition. In *Neurocomputing: Algorithms, Architectures and Applications*, pages 227–236. Springer, 1990.

[12] Peter F Brown, Peter V Desouza, Robert L Mercer, Vincent J Della Pietra, and Jenifer C Lai. Class-based n-gram models of natural language. *Computational linguistics*, 18(4):467–479, 1992.

[13] Lijuan Cai and Thomas Hofmann. Text categorization by boosting automatically extracted concepts. In *SIGIR*, pages 182–189, 2003.

[14] Eugene Charniak and Mark Johnson. Coarse-to-fine n-best parsing and maxent discriminative reranking. In *ACL*, pages 173–180, 2005.

[15] Xinxiong Chen, Lei Xu, Zhiyuan Liu, Maosong Sun, and Huanbo Luan. Joint learning of character and word embeddings. In *International Joint Conference on Artificial Intelligence*, 2015.

[16] William W Cohen, Robert E Schapire, and Yoram Singer. Learning to order things. *Journal of Artificial Intelligence Research*, 10:243–270, 1998.

[17] Ronan Collobert and Jason Weston. A unified architecture for natural language processing: Deep neural networks with multitask learning. *International Conference on Machine Learning*, 2008.

[18] Ronan Collobert, Jason Weston, Léon Bottou, Michael Karlen, Koray Kavukcuoglu, and Pavel Kuksa. Natural language processing (almost) from scratch. *The Journal of Machine Learning Research*, 12:2493–2537, 2011.

[19] Thomas M Cover and Joy A Thomas. *Elements of information theory*. John Wiley & Sons, 2012.

[20] Ido Dagan, Lillian Lee, and Fernando CN Pereira. Similarity-based models of word cooccurrence probabilities. *Machine Learning*, 34(1-3):43–69, 1999.

[21] Ido Dagan, Shaul Marcus, and Shaul Markovitch. Contextual word similarity and estimation from sparse data. In *Proceedings of the 31st annual meeting on Association for Computational Linguistics*, pages 164–171, 1993.

[22] Paramveer S. Dhillon, Dean P. Foster, and Lyle H. Ungar. Multi-view learning of word embeddings via cca. In *Advances in Neural Information Processing Systems*, pages 199–207, 2011.

[23] Paramveer S. Dhillon, Dean P. Foster, and Lyle H. Ungar. Eigenwords: Spectral word embeddings. *The Journal of Machine Learning Research*, 16, 2015.

[24] Elizabeth D Dolan and Jorge J Moré. Benchmarking optimization software with performance profiles. *Mathematical programming*, 91(2):201–213, 2002.

[25] John Duchi, Elad Hazan, and Yoram Singer. Adaptive subgradient methods for online learning and stochastic optimization. *The Journal of Machine Learning Research*, 12:2121–2159, 2011.

[26] Jeffrey L Elman. Finding structure in time. *Cognitive science*, 14(2):179–211, 1990.

[27] Dumitru Erhan, Yoshua Bengio, Aaron Courville, Pierre-Antoine Manzagol, Pascal Vincent, and Samy Bengio. Why does unsupervised pre-training help deep learning? *The Journal of Machine Learning Research*, 11:625–660, 2010.

[28] Lev Finkelstein, Evgeniy Gabrilovich, Yossi Matias, Ehud Rivlin, Zach Solan, Gadi Wolfman, and Eytan Ruppin. Placing search in context: The concept revisited. *ACM Transactions on Information Systems*, 20(1):116–131, 2002.

[29] John R Firth. A synopsis of linguistic theory, 1930-1955. *Studies in Linguistic Analysis*, 1957.

[30] Gottlob Frege. Über sinn und bedeutung. *Funktion - Begriff - Bedeutung*, 1892.

[31] Kunihiko Fukushima. Neocognitron: A self-organizing neural network model for a mechanism of pattern recognition unaffected by shift in position. *Biological cybernetics*, 36(4):193–202, 1980.

[32] Xavier Glorot, Antoine Bordes, and Yoshua Bengio. Deep sparse rectifier networks. In *Proceedings of the 14th International Conference on Artificial Intelligence and Statistics. JMLR W&CP Volume*, volume 15, pages 315–323, 2011.

[33] Irving J Good. The population frequencies of species and the estimation of population parameters. *Biometrika*, 40(3-4):237–264, 1953.

[34] Michael U Gutmann and Aapo Hyvärinen. Noise-contrastive estimation of unnormalized statistical models, with applications to natural image statistics. *The Journal of Machine Learning Research*, 13(1):307–361, 2012.

[35] Zellig S Harris. Distributional structure. *Word*, 1954.

[36] Karl Moritz Hermann. *Distributed Representations for Compositional Semantics*. PhD thesis, University of Oxford, 2014.

[37] Felix Hill, KyungHyun Cho, Sebastien Jean, Coline Devin, and Yoshua Bengio. Not all neural embeddings are born equal. *arXiv preprint arXiv:1410.0718*, 2014.

[38] Swapnil Hingmire, Sandeep Chougule, Girish K Palshikar, and Sutanu Chakraborti. Document classification by topic labeling. In *SIGIR*, pages 877–880, 2013.

[39] Geoffrey E Hinton. Learning distributed representations of concepts. In *Proceedings of the eighth annual conference of the cognitive science society*, volume 1, page 12, 1986.

[40] Geoffrey E Hinton and Ruslan R Salakhutdinov. Reducing the dimensionality of data with neural networks. *Science*, 313(5786):504–507, 2006.

[41] Geoffrey E Hinton, Nitish Srivastava, Alex Krizhevsky, Ilya Sutskever, and Ruslan R Salakhutdinov. Improving neural networks by preventing co-adaptation of feature detectors. *arXiv preprint arXiv:1207.0580*, 2012.

[42] Sepp Hochreiter. The vanishing gradient problem during learning recurrent neural nets and problem solutions. *International Journal of Uncertainty, Fuzziness and Knowledge-Based Systems*, 6(02):107–116, 1998.

[43] Sepp Hochreiter and Jürgen Schmidhuber. Long short-term memory. *Neural computation*, 9(8):1735–1780, 1997.

[44] Eric H Huang, Richard Socher, Christopher D Manning, and Andrew Y Ng. Improving word representations via global context and multiple word prototypes. In *ACL*, pages 873–882, 2012.

[45] Michael N Jones and Douglas JK Mewhort. Representing word meaning and order information in a composite holographic lexicon. *Psychological review*, 114(1):1, 2007.

[46] Nal Kalchbrenner and Phil Blunsom. Recurrent convolutional neural networks for discourse compositionality. In *Workshop on CVSC*, pages 119–126, 2013.

[47] Nal Kalchbrenner, Edward Grefenstette, and Phil Blunsom. A convolutional neural network for modelling sentences. In *Proceedings of the 52nd Annual Meeting of the Association for Computational Linguistics*, pages 655–665, 2014.

[48] Yoon Kim. Convolutional neural networks for sentence classification. In *Proceedings of the 2014 Conference on Empirical Methods in Natural Language Processing (EMNLP)*, pages 1746–1751, 2014.

[49] Dan Klein and Christopher D. Manning. Accurate unlexicalized parsing. In *ACL*, pages 423–430, 2003.

[50] Alexandre Klementiev, Ivan Titov, and Binod Bhattarai. Inducing crosslingual distributed representations of words. In *Coling*, pages 1459–1474, 2012.

[51] Reinhard Kneser and Hermann Ney.  Improved backing-off for m-gram lan-guage modeling.  In *Acoustics, Speech, and Signal Processing, 1995. ICASSP-95., 1995 International Conference on*, volume 1, pages 181–184, 1995.

[52] Youngjoong Ko.  A study of term weighting schemes using class information for text classification.  In *SIGIR*, pages 1029–1030, 2012.

[53] Philipp Koehn.  Europarl: A parallel corpus for statistical machine translation. In *MT summit*, volume 5, pages 79–86, 2005.

[54] Thomas K Landauer.  On the computational basis of learning and cognition: Arguments from lsa. *Psychology of learning and motivation*, 41:43–84, 2002.

[55] Thomas K Landauer and Susan T Dumais. A solution to plato's problem: The latent semantic analysis theory of acquisition, induction, and representation of knowledge. *Psychological review*, 104(2):211, 1997.

[56] Thomas K Landauer, Peter W Foltz, and Darrell Laham.  An introduction to latent semantic analysis. *Discourse processes*, 25(2-3):259–284, 1998.

[57] Gabriella Lapesa, Stefan Evert, and Sabine Schulte im Walde. Contrasting syn-tagmatic and paradigmatic relations: Insights from distributional semantic mod-els. In *Proceedings of the Third Joint Conference on Lexical and Computational Semantics (SEM 2014)*, pages 160–170, 2014.

[58] Quoc V Le and Tomas Mikolov.  Distributed representations of sentences and documents.  In *ICML*, 2014.

[59] Rémi Lebret and Ronan Collobert.  Word embeddings through hellinger pca. *EACL 2014*, page 482, 2014.

[60] Rémi Lebret and Ronan Collobert. Word embeddings through hellinger pca. In *Proceedings of the 14th Conference of the European Chapter of the Association for Computational Linguistics*, pages 482–490, 2014.

[61] Yann LeCun, Léon Bottou, Yoshua Bengio, and Patrick Haffner. Gradient-based learning applied to document recognition. *Proceedings of the IEEE*, 86(11):2278–2324, 1998.

[62] Daniel D Lee and H Sebastian Seung. Algorithms for non-negative matrix factorization. In *Advances in neural information processing systems*, pages 556–562, 2001.

[63] Gregory W Lesher, Bryan J Moulton, D Jeffery Higginbotham, et al. Effects of ngram order and training text size on word prediction. In *Proceedings of the RESNA'99 Annual Conference*, pages 52–54, 1999.

[64] Omer Levy and Yoav Goldberg. Neural word embedding as implicit matrix factorization. In *Advances in Neural Information Processing Systems*, pages 2177–2185, 2014.

[65] Omer Levy, Yoav Goldberg, and Ido Dagan. Improving distributional similarity with lessons learned from word embeddings. *Transactions of the Association for Computational Linguistics*, 2015.

[66] David D Lewis. An evaluation of phrasal and clustered representations on a text categorization task. In *SIGIR*, pages 37–50, 1992.

[67] Yitan Li, Linli Xu, Fei Tian, Liang Jiang, Xiaowei Zhong, and Enhong Chen. Word embedding revisited: A new representation learning and explicit matrix factorization perspective. In *International Joint Conference on Artificial Intelligence*, 2015.

[68] Chih-Jen Lin. Projected gradient methods for nonnegative matrix factorization. *Neural computation*, 19(10):2756–2779, 2007.

[69] Dekang Lin and Xiaoyun Wu. Phrase clustering for discriminative learning. In *Proceedings of the Joint Conference of the 47th Annual Meeting of the ACL and the 4th International Joint Conference on Natural Language Processing*, pages 1030–1038, 2009.

[70] Kevin Lund, Curt Burgess, and Ruth Ann Atchley. Semantic and associative priming in high-dimensional semantic space. In *Proceedings of the 17th annual conference of the Cognitive Science Society*, volume 17, pages 660–665, 1995.

[71] Andrew L Maas, Raymond E Daly, Peter T Pham, Dan Huang, Andrew Y Ng, and Christopher Potts. Learning word vectors for sentiment analysis. In *Proceedings of the 49th Annual Meeting of the Association for Computational Linguistics (ACL)*, pages 142–150, 2011.

[72] Tomáš Mikolov. *Statistical language models based on neural networks*. PhD thesis, Brno University of Technology, 2012.

[73] Tomas Mikolov, Kai Chen, Greg Corrado, and Jeffrey Dean. Efficient estimation of word representations in vector space. *International Conference on Learning Representations Workshop Track*, 2013.

[74] Tomas Mikolov, Martin Karafiát, Lukas Burget, Jan Cernockỳ, and Sanjeev Khudanpur. Recurrent neural network based language model. In *INTERSPEECH 2010, 11th Annual Conference of the International Speech Communication Association*, pages 1045–1048, 2010.

[75] Tomas Mikolov, Ilya Sutskever, Kai Chen, Greg S Corrado, and Jeff Dean. Distributed representations of words and phrases and their compositionality. In *Advances in neural information processing systems*, pages 3111–3119, 2013.

[76] Tomas Mikolov, Wen-tau Yih, and Geoffrey Zweig. Linguistic regularities in continuous space word representations. In *NAACL-HLT*, pages 746–751, 2013.

[77] Dmitrijs Milajevs, Dimitri Kartsaklis, Mehrnoosh Sadrzadeh, and Matthew Purver. Evaluating neural word representations in tensor-based compositional settings. In *Proceedings of the 2014 Conference on Empirical Methods in Natural Language Processing (EMNLP)*, pages 708–719, 2014.

[78] Jeff Mitchell and Mirella Lapata. Composition in distributional models of semantics. *Cognitive science*, 34(8):1388–1429, 2010.

[79] Andriy Mnih and Geoffrey Hinton. Three new graphical models for statistical language modelling. In *Proceedings of the 24th international conference on Machine learning*, pages 641–648, 2007.

[80] Andriy Mnih and Geoffrey E Hinton. A scalable hierarchical distributed language model. In *Advances in neural information processing systems*, pages 1081–1088, 2008.

[81] Andriy Mnih and Koray Kavukcuoglu. Learning word embeddings efficiently with noise-contrastive estimation. In *Advances in Neural Information Processing Systems*, pages 2265–2273, 2013.

[82] Frederic Morin and Yoshua Bengio. Hierarchical probabilistic neural network language model. In *Proceedings of the international workshop on artificial intelligence and statistics*, pages 246–252, 2005.

[83] Frederic Morin and Yoshua Bengio. Hierarchical probabilistic neural network language model. In *AISTATS*, pages 246–252, 2005.

[84] Tetsuji Nakagawa, Kentaro Inui, and Sadao Kurohashi. Dependency tree-based sentiment classification using crfs with hidden variables. In *Human Language Technologies: The 2010 Annual Conference of the North American Chapter of the Association for Computational Linguistics*, pages 786–794, 2010.

[85] Andrew Y Ng. Feature selection, l1 vs. l2 regularization, and rotational invariance. In *ICML*, page 78, 2004.

[86] Bo Pang and Lillian Lee. A sentimental education: Sentiment analysis using subjectivity summarization based on minimum cuts. In *Proceedings of the 42nd annual meeting on Association for Computational Linguistics*, page 271, 2004.

[87] Bo Pang and Lillian Lee. Seeing stars: Exploiting class relationships for sentiment categorization with respect to rating scales. In *ACL*, pages 115–124, 2005.

[88] Arkadiusz Paterek. Improving regularized singular value decomposition for collaborative filtering. In *Proc. KDD Cup Workshop at SIGKDD'07, 13th ACM Int. Conf. on Knowledge Discovery and Data Mining*, pages 39–42, 2007.

[89] Fuchun Peng, Fangfang Feng, and Andrew McCallum. Chinese segmentation and new word detection using conditional random fields. In *Proceedings of the 20th international conference on Computational Linguistics*, page 562, 2004.

[90] Jeffrey Pennington, Richard Socher, and Christopher D Manning. GloVe : Global Vectors for Word Representation. In *Proceedings of the Empiricial Methods in Natural Language Processing*, 2014.

[91] Fernando Pereira, Naftali Tishby, and Lillian Lee. Distributional clustering of english words. In *Proceedings of the 31st annual meeting on Association for Computational Linguistics*, pages 183–190, 1993.

[92] Slav Petrov, Leon Barrett, Romain Thibaux, and Dan Klein. Learning accurate, compact, and interpretable tree annotation. In *Coling-ACL*, pages 433–440, 2006.

[93] David C Plaut and Geoffrey E Hinton. Learning sets of filters using backpropagation. *Computer Speech & Language*, 2(1):35–61, 1987.

[94] Matt Post and Shane Bergsma. Explicit and implicit syntactic features for text classification. In *ACL*, pages 866–872, 2013.

[95] Lutz Prechelt. Early stopping-but when? *Neural Networks: Tricks of the trade*, pages 55–69, 1998.

[96] Dragomir R Radev, Pradeep Muthukrishnan, and Vahed Qazvinian. The acl anthology network corpus. In *NLPIR4DL*, pages 54–61, 2009.

[97] Lev Ratinov and Dan Roth. Design challenges and misconceptions in named entity recognition. In *Proceedings of the Thirteenth Conference on Computational Natural Language Learning*, pages 147–155, 2009.

[98] Magnus Sahlgren. *The Word-Space Model*. PhD thesis, Gothenburg University, 2006.

[99] Gerard Salton and Christopher Buckley. Term-weighting approaches in automatic text retrieval. *Information processing & management*, 24(5):513–523, 1988.

[100] Hinrich Schütze. Context space. In *AAAI fall symposium on probabilistic approaches to natural language*, pages 113–120, 1992.

[101] Hinrich Schütze. Dimensions of meaning. In *Supercomputing'92., Proceedings*, pages 787–796, 1992.

[102] Richard Socher. *Recursive Deep Learning for Natural Language Processing and Computer Vision*. PhD thesis, Stanford University, 2014.

[103] Richard Socher, John Bauer, Christopher D Manning, and Andrew Y Ng. Parsing with compositional vector grammars. In *ACL*, pages 455–465, 2013.

[104] Richard Socher, Danqi Chen, Christopher D Manning, and Andrew Ng. Reasoning with neural tensor networks for knowledge base completion. In *Advances in Neural Information Processing Systems*, pages 926–934, 2013.

[105] Richard Socher, Eric H Huang, Jeffrey Pennington, Andrew Y Ng, and Christopher D Manning. Dynamic pooling and unfolding recursive autoencoders for paraphrase detection. In *NIPS*, volume 24, pages 801–809, 2011.

[106] Richard Socher, Brody Huval, Christopher D Manning, and Andrew Y Ng. Semantic compositionality through recursive matrix-vector spaces. In *EMNLP-CoNLL*, pages 1201–1211, 2012.

[107] Richard Socher, Andrej Karpathy, Quoc V Le, Christopher D Manning, and Andrew Y Ng. Grounded compositional semantics for finding and describing images with sentences. *Transactions of the Association for Computational Linguistics*, 2:207–218, 2014.

[108] Richard Socher, Cliff C Lin, Chris Manning, and Andrew Y Ng. Parsing natural scenes and natural language with recursive neural networks. In *Proceedings of the 28th international conference on machine learning (ICML-11)*, pages 129–136, 2011.

[109] Richard Socher, Jeffrey Pennington, Eric H Huang, Andrew Y Ng, and Christopher D Manning. Semi-supervised recursive autoencoders for predicting sentiment distributions. In *Proceedings of the Conference on Empirical Methods in Natural Language Processing (EMNLP11)*, pages 151–161, 2011.

[110] Richard Socher, Alex Perelygin, Jean Wu, Jason Chuang, Christopher D. Manning, Andrew Ng, and Christopher Potts. Recursive deep models for semantic compositionality over a sentiment treebank. In *Proceedings of the 2013 Conference on Empirical Methods in Natural Language Processing (EMNLP13)*, pages 1631–1642, 2013.

[111] Pontus Stenetorp, Hubert Soyer, Sampo Pyysalo, Sophia Ananiadou, and Takashi Chikayama. Size (and domain) matters: Evaluating semantic word space representations for biomedical text. In *Proceedings of the 5th International Symposium on Semantic Mining in Biomedicine*, 2012.

[112] Fei Sun, Jiafeng Guo, Yanyan Lan, Jun Xu, and Xueqi Cheng. Inside out: Two jointly predictive models for word representations and phrase representations. In *Proceedings of the 30th AAAI conference*, 2016.

[113] Ilya Sutskever, James Martens, and Geoffrey E Hinton. Generating text with recurrent neural networks. In *Proceedings of the 28th International Conference on Machine Learning (ICML-11)*, pages 1017–1024, 2011.

[114] Buzhou Tang, Xuan Wang, and Xiaohong Wang. Chinese word segmentation based on large margin methods. *International Journal of Asian Language Processing*, 19(2):55–68, 2009.

[115] Kristina Toutanova, Dan Klein, Christopher D Manning, and Yoram Singer. Feature-rich part-of-speech tagging with a cyclic dependency network. In *Proceedings of the 2003 Conference of the North American Chapter of the Association for Computational Linguistics on Human Language Technology-Volume 1*, pages 173–180, 2003.

[116] Joseph Turian, Lev Ratinov, and Yoshua Bengio. Word representations: a simple and general method for semi-supervised learning. In *Proceedings of the 48th*

*annual meeting of the association for computational linguistics (ACL)*, pages 384–394, 2010.

[117] Peter D Turney. Similarity of semantic relations. *Computational Linguistics*, 32(3):379–416, 2006.

[118] Peter D Turney and Patrick Pantel. From frequency to meaning: Vector space models of semantics. *Journal of artificial intelligence research*, 37:141–188, 2010.

[119] Kun Wang, Chengqing Zong, and Keh-Yih Su. A character-based joint model for chinese word segmentation. In *Proceedings of the 23rd International Conference on Computational Linguistics*, pages 1173–1181, 2010.

[120] Sida Wang and Christopher D Manning. Baselines and bigrams: Simple, good sentiment and topic classification. In *ACL*, pages 90–94, 2012.

[121] Sida I. Wang and Christopher D. Manning. Fast dropout training. In *In Proceedings of the ICML*, 2013.

[122] Zhen Wang, Jianwen Zhang, Jianlin Feng, and Zheng Chen. Knowledge graph and text jointly embedding. In *Proceedings of the 2014 Conference on Empirical Methods in Natural Language Processing (EMNLP)*, pages 1591–1601, 2014.

[123] Jason Weston, Antoine Bordes, Oksana Yakhnenko, and Nicolas Usunier. Connecting language and knowledge bases with embedding models for relation extraction. In *Proceedings of the 2013 Conference on Empirical Methods in Natural Language Processing (EMNLP)*, pages 1366–1371, 2013.

[124] Wei Xu and Alex Rudnicky. Can artificial neural networks learn language models? In *Sixth International Conference on Spoken Language Processing*, 2000.

[125] Nianwen Xue et al. Chinese word segmentation as character tagging. *Computational Linguistics and Chinese Language Processing*, 8(1):29–48, 2003.

[126] Matthew D Zeiler. Adadelta: an adaptive learning rate method. *arXiv preprint arXiv:1212.5701*, 2012.

[127] Daojian Zeng, Kang Liu, Siwei Lai, Guangyou Zhou, and Jun Zhao. Relation classification via convolutional deep neural network. In *Proceedings of COL-ING 2014, the 25th International Conference on Computational Linguistics: Technical Papers*, pages 2335–2344, 2014.

[128] Yin Zhang, Rong Jin, and Zhi-Hua Zhou. Understanding bag-of-words model: a statistical framework. *International Journal of Machine Learning and Cybernetics*, 1(1-4):43–52, 2010.

[129] Hai Zhao, Chang-Ning Huang, Mu Li, and Bao-Liang Lu. Effective tag set selection in chinese word segmentation via conditional random field modeling. In *Proceedings of PACLIC*, volume 20, pages 87–94, 2006.

[130] Hai Zhao and Chunyu Kit. Integrating unsupervised and supervised word segmentation: The role of goodness measures. *Information Sciences*, 181(1):163–183, 2011.

[131] Xiaoqing Zheng, Hanyang Chen, and Tianyu Xu. Deep learning for Chinese word segmentation and POS tagging. In *Proceedings of the 2013 Conference on Empirical Methods in Natural Language Processing*, pages 647–657, 2013.

[132] 刘群, 张华平, 俞鸿魁, 程学旗. 基于层叠隐马模型的汉语词法分析. 计算机研究与发展, 41(8):1421–1429, 2004.

[133] 李文波, 孙乐, 张大鲲. 基于 Labeled-LDA 模型的文本分类新算法. 计算机学报, 31(4):620–627, 2008.

[134] 李荣陆. 文本分类及其相关技术研究. PhD thesis, 上海: 复旦大学计算机与信息技术系, 2005.

[135] 黄昌宁, 赵海. 中文分词十年回顾. 中文信息学报, 21(3):8–19, 2007.

# 发表文章目录

[1] **Siwei Lai**, Liheng Xu, Kang Liu, and Jun Zhao. Recurrent Convolutional Neural Networks for Text Classification. In *Proceedings of the 29th AAAI Conference on Artificial Intelligence (AAAI'15)*, pages 2267–2273. 2015.

[2] 来斯惟, 徐立恒, 陈玉博, 刘康, 赵军. 基于表示学习的中文分词算法探索. 中文信息学报. 27(5):8–14. 2013.

[3] **Siwei Lai**, Yang Liu, Huxiang Gu, Liheng Xu, Kang Liu, Shiming Xiang, Jun Zhao, Rui Diao, Liang Xiang, Hang Li, and Dong Wang. Hybrid Recommendation Models for Binary User Preference Prediction Problem. *The Journal of Machine Learning Research Workshop and Conference Proceedings (JMLR W&CP)*. 18:137–151. 2012.

[4] Daojian Zeng, Kang Liu, **Siwei Lai**, Guangyou Zhou, and Jun Zhao. Relation Classification via Convolutional Deep Neural Network. In *Proceedings of the 25th International Conference on Computational Linguistics (COLING'14)*, pages 2335–2344. 2014. (Best Papers Award).

[5] Liheng Xu, Kang Liu, **Siwei Lai** and Jun Zhao. Product Feature Mining: Semantic Clues versus Syntactic Constituents. In *Proceedings of the 52nd Annual Meeting of the Association for Computational Linguistics (ACL'14)*, pages 336–346. 2014.

[6] **Siwei Lai**, Kang Liu, Liheng Xu, Jun Zhao. How to Generate a Good Word Embedding? arXiv:1507.05523. (Under review).

# 简　历

**基本情况**

来斯惟，男，1987 年 3 月出生，中国科学院自动化研究所在读博士研究生。

**教育状况**

2006 年 9 月至 2010 年 7 月，北京化工大学，获得工学学士学位，
　　专业：计算机科学与技术。

2010 年 9 月至 2015 年 12 月，中国科学院自动化研究所，硕博连读研究生，
　　专业：模式识别与智能系统。

**研究兴趣**

自然语言处理，推荐系统

**联系方式**

通讯地址：北京市中关村东路 95 号

邮编：100190

E-mail: swlai@nlpr.ia.ac.cn

# 致　谢